\documentclass[preprint,12pt]{elsarticle}
\usepackage{amssymb}
\usepackage{amsmath}

\usepackage{booktabs}
\usepackage{mathrsfs}

\usepackage{hyperref}
\usepackage{algorithmic}
\usepackage[ruled]{algorithm2e}
\SetNlSty{\scriptsize}{}{:}

\SetAlFnt{\scriptsize}                   
\SetAlCapFnt{\small}                
\SetNlSty{\scriptsize}{}{:}            

\usepackage{multirow}
\usepackage{bbding}

\usepackage{numprint}
\npdecimalsign{.}
\nprounddigits{3}

\usepackage[english]{babel}
\usepackage{amsthm}
\usepackage{array}
\usepackage{enumitem}
\newtheorem{definition}{Definition}
\newtheorem{theorem}{Theorem}[section]

\usepackage{xcolor}
\usepackage{stfloats}
\usepackage{subfigure}
\usepackage{graphicx}
\usepackage{subcaption}
\usepackage{amsmath}
\usepackage{diagbox}
\usepackage{caption}

\usepackage[numbers]{natbib}
\usepackage{xr}

\SetKwComment{Comment}{/* }{ */}
\theoremstyle{remark}
\newtheorem{remark}{Remark}

\usepackage{silence}
\theoremstyle{definition}
\usepackage{xr}

\usepackage{hyperref}

\usepackage{silence}

\begin{document}


\begin{frontmatter}



\title{Rebalancing with Calibrated Sub-classes (RCS): A Statistical Fusion-based Framework for Robust Imbalanced Classification across Modalities}




\author[label1]{Priyobrata~Mondal}
\author[label1]{Faizanuddin~Ansari}
\author[label1]{Swagatam~Das }

\affiliation[label1]{organization={Electronics and Communication Sciences Unit},
            addressline={Indian Statistical Institute}, 
            city={Kolkata},
            postcode={ 700108}, 
            state={West Bengal},
            country={India}}

\begin{abstract}
   \textcolor{black}{Class imbalance, where certain classes have insufficient data, poses a critical challenge for robust classification, often biasing models toward majority classes. Distribution calibration offers a promising avenue to address this by estimating more accurate class distributions. In this work, we propose Rebalancing with Calibrated Sub-classes (RCS) — a novel distribution calibration framework for robust imbalanced classification. RCS aims to fuse statistical information from the majority and intermediate class distributions via a weighted mixture of Gaussian components to estimate minority class parameters more accurately. An encoder-decoder network is trained to preserve structural relationships in imbalanced datasets and prevent feature disentanglement. Post-training, encoder-extracted feature vectors are leveraged to generate synthetic samples guided by the calibrated distributions. This fusion-based calibration effectively mitigates overgeneralization by incorporating neighborhood distribution information rather than relying solely on majority-class statistics. Extensive experiments on diverse image, text, and tabular datasets demonstrate that RCS consistently outperforms several baseline and state-of-the-art methods, highlighting its effectiveness and broad applicability in addressing real-world imbalanced classification challenges.
\color{black}RCS code is available at \url{https://anonymous.4open.science/r/RCS-CF76}.\color{black}}
\end{abstract}



\begin{keyword}
Class Imbalance
\sep  Distribution Calibration
\sep Synthetic Sample Generation


\end{keyword}
\end{frontmatter}
\WarningFilter*{hyperref}{Suppressing link with}
\WarningFilter*{hyperref}{Token not}
\WarningFilter{latex}{Citation}
\WarningFilter{latex}{Overfull}
\WarningFilter{latex}{Underfull}

\externaldocument{Supplementary}

\section{Introduction}
In recent years, deep learning algorithms have demonstrated remarkable success in various domains, ranging from computer vision 
and natural language processing 
and healthcare~\cite{SHAH2026103503}.
The ability of deep neural networks to learn complex patterns and representations from vast amounts of data has propelled their widespread adoption. However, despite their efficacy, deep learning models often grapple with the challenge of imbalanced datasets, where certain classes remain underrepresented in the training set, leading to suboptimal performance, biased predictions, and diminished generalization capabilities~\cite{SANTOS2023228}. The issue of imbalanced data distribution is pervasive in real-world scenarios, where certain classes may be rare or have limited instances compared to others. Traditional deep learning algorithms are inclined to prioritize the majority class, resulting in a biased learning process that adversely affects the model's ability to classify minority classes accurately. Such an imbalance poses a significant hurdle, particularly in applications where the consequences of misclassification are severe, \textcolor{black}{such as in medical diagnosis
and fraud detection.}

The objective of oversampling techniques is to address the imbalance in dataset classes by comprehending the distribution of minority classes and enhancing them. Traditional methods, such as the Synthetic Minority Oversampling Technique (SMOTE)~\cite{chawla2002smote} introduced by Chawla et al. in 2002, and different variants of SMOTE 
, generate synthetic data for the minority class through interpolations among the nearest neighbors of a specific point from that class. However, these approaches are less effective when dealing with high-dimensional data, such as images, due to challenges associated with the `curse of dimensionality.' 
To address this, deep generative models like Generative Adversarial Networks (GANs) \cite{goodfellow2014generative} and Variational Auto-Encoders (VAEs) \cite{kingma2013auto} have been employed. These models, utilized in approaches such as GAMO~\cite{mullick2019generative}, DGC~\cite{wang2020deep}, DeepSMOTE~\cite{dablain2021deepsmote}, and DGCMM~\cite{wang2022deep}, generate additional samples for the minority class to mitigate class imbalance in high-dimensional data. Despite their significant advancements over classical methods, there exists a potential drawback of diminished diversity in the generated minority samples, particularly in scenarios with limited data. This limitation is further exacerbated by factors such as limited representation within the training data, leading to biased or incomplete sampling of the minority class. Additionally, challenges such as overfitting, complexity of data distribution, and the design of loss functions can contribute to the reduced diversity observed in the generated samples.

The consequence of such limitations becomes clear when considering the generalizability of classifier training. Balanced representation of each class is crucial, particularly for imbalanced data. In such scenarios, the majority class often dominates, potentially leading to biased models that neglect the nuances of the underrepresented class. Oversampling techniques have emerged as a solution, aiming to preserve valuable information and remain independent of domain-specific knowledge. However, their effectiveness hinges on two fundamental principles:\\
\textbf{\textit{Authenticity}}: Newly generated samples must genuinely belong to the underrepresented class without introducing misleading artifacts. This ensures the classifier learns from accurate representations of the minority class, preventing misinterpretations and biases.\\
 \textbf{\textit{Pattern Coverage}}: Synthesized samples should capture the diverse range of patterns exhibited within the underrepresented class. This enhances the classifier's generalizability, accurately identifying unseen rare instances.

 Unfortunately, conventional oversampling approaches often suffer from shortcomings in these crucial aspects. Traditional methods can inadvertently introduce spurious data or fail to capture the full spectrum of patterns within the minority class. This can lead to models that struggle with generalizability and exhibit poor performance on unseen data. To address these limitations, we propose a novel disentangled oversampling method that adheres to the fundamental principles of authenticity and pattern coverage. 
 
 Our proposal first attempts to create disentangled representations from the already available imbalanced data if the feature vectors are not disentangled. Our goal is to extract feature vectors from the given dataset in a manner that ensures that there is no overlap in the feature space, provided there is no overlap in the data space. Below, we formally present the concept of disentangled representation.
\begin{definition} Let $X$ be the set of data points, $\mathscr{L}$ be the space of latent representations, and $\mathscr{E}:X\mapsto \mathscr{L}$ be a function mapping data points to feature space. Let $U_1, U_2,\cdots, U_K \subset X$ represent elements from different classes. Without loss of generality, suppose $U_i,U_j \subset X$ denote datapoints from the $i^{th}$ and $j^{th}$ classes respectively, and $\mathscr{E}(U_i) = \tilde{U}_i$, $\mathscr{E}(U_j) = \tilde{U}_j$. Let $q_1,q_2,\cdots,q_K$ be the class conditional distributions for the latent representations $\tilde{U}_1,\tilde{U}_2,\cdots,\tilde{U}_K$. Then $\{z\sim q_i : z = \mathscr{E}(x) \text{ and } x\in U_i\}\cap \{z\sim q_j : z = \mathscr{E}(x) \text{ and } x\in U_j\} = \phi$.
\end{definition}
This kind of disentangled representation, where the features of different classes are well separated and do not overlap, helps to fulfill the authenticity principle mentioned earlier.
\color{black} 
 
 Furthermore, these disentangled representations of the feature vectors are utilized to generate new samples for classes with insufficient data to address pattern coverage. For these classes, the data distribution is estimated using the statistical characteristics of the majority classes, aiming to approximate the original distribution for the classes with insufficient data. This method, termed distribution calibration~\cite{yang2021free, vigneswaran2021feature}, endeavors to capture the distribution of the minority class as if the dataset were balanced, thereby producing new samples representing the minority class patterns.

\begin{figure}
\scriptsize
  \centering
  \subfigure[\label{fig:1a}]{\includegraphics[width=0.218\textwidth]{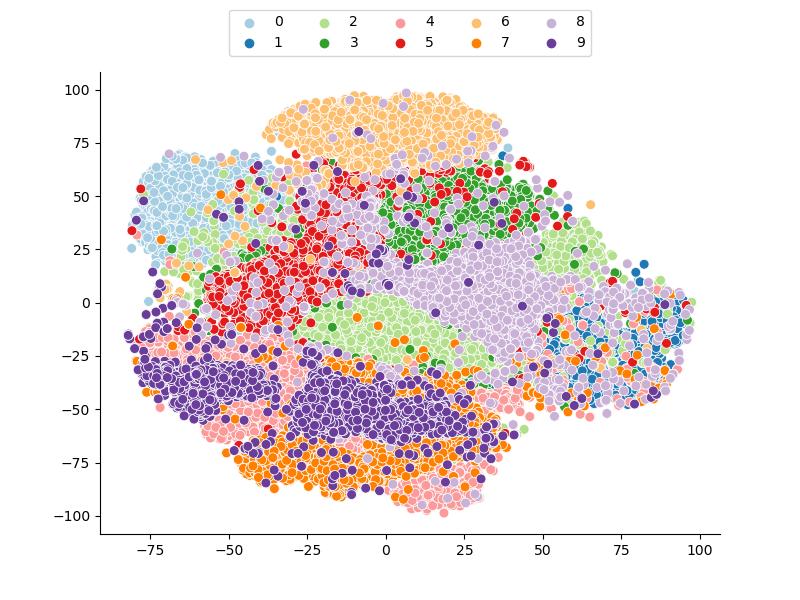}}\hspace{-0.4cm}
    \subfigure[\label{fig:1b}]{\includegraphics[width=0.218\textwidth]{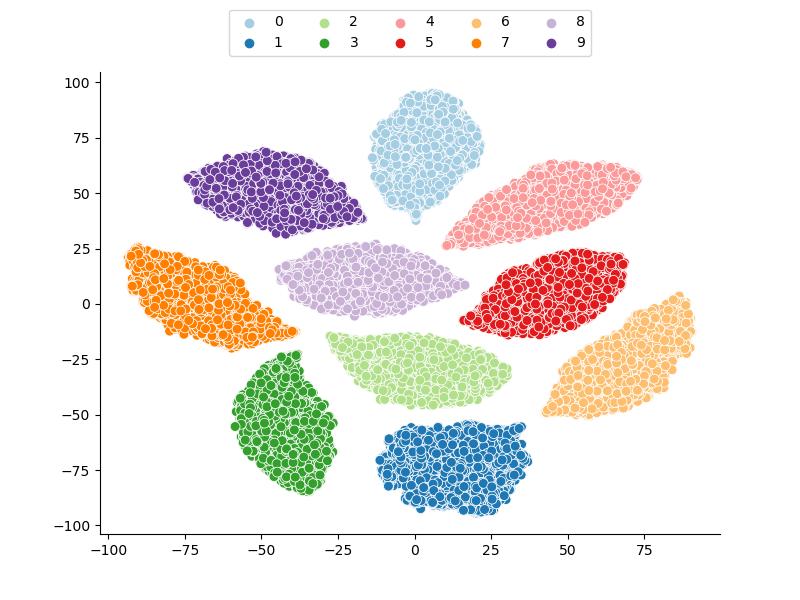}}\hspace{-0.4cm}
  \subfigure[\label{fig:1c}]{\includegraphics[width=0.218\textwidth]{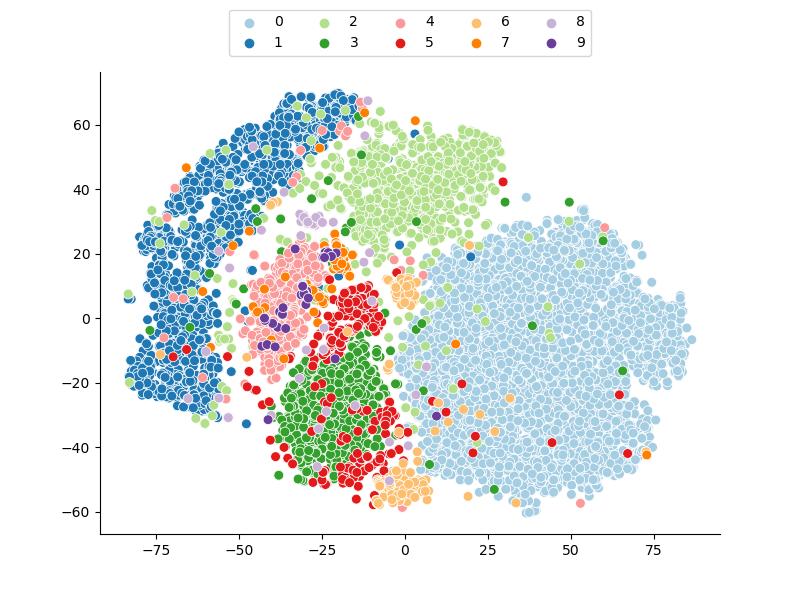}}\hspace{-0.4cm}
  \subfigure[\label{fig:1d}]{\includegraphics[width=0.218\textwidth]{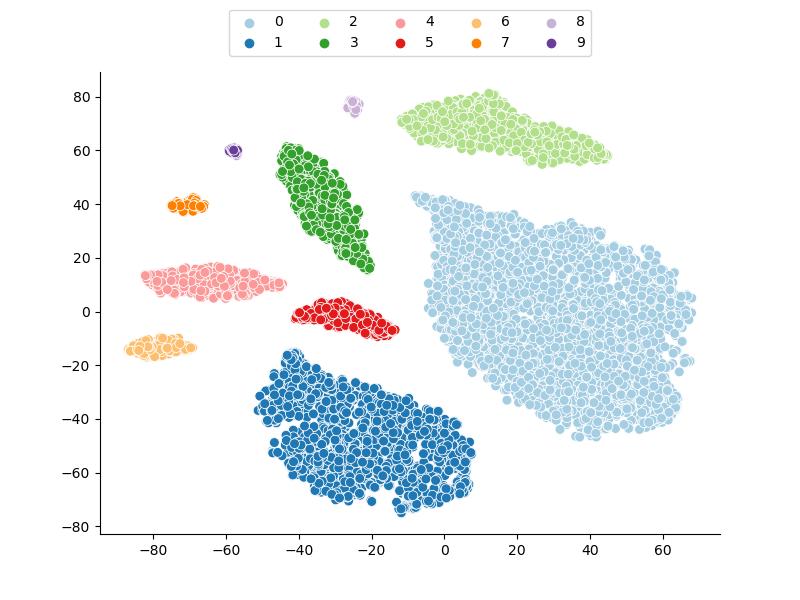}}\hspace{-0.4cm}
  \subfigure[\label{fig:1e}]{\includegraphics[width=0.218\textwidth]{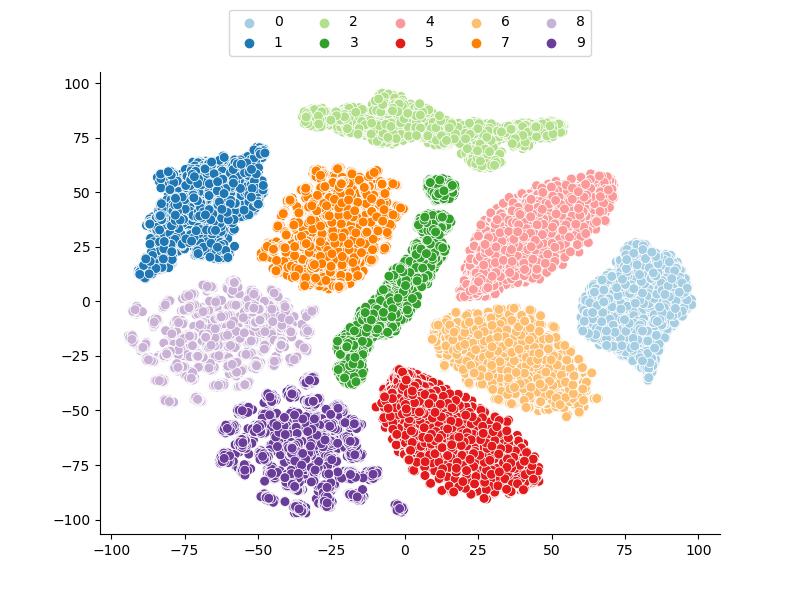}}\hspace{-0.4cm}
  \vspace{-0.2cm}
  \caption{Figures depict the t-SNE plots of feature vectors in the latent space for the MNIST dataset. (a) Shows the balanced feature vectors when the vanilla autoencoder is trained using the balanced data. (b) Displays the balanced feature vectors when our autoencoder is trained using the balanced data. (c) Illustrates the imbalanced feature vectors when trained using the vanilla autoencoder on the imbalanced dataset. (d) Represents the imbalanced feature vectors from our autoencoder network. (e) Visualizes the oversampling from the latent feature vectors obtained from our trained autoencoder network.}
  \label{fig:1}
\end{figure}

Our methodology adopts a systematic approach composed of two key stages:\\
 \textbf{\textit{Stage 1: Disentanglement}}: We leverage a disentanglement technique to extract latent representations of the data. This process ensures that the latent space captures the underlying factors of variation within each class, minimizing interference between different classes. Furthermore, our approach incorporates a sophisticated regularization strategy within the Autoencoder training framework. The regularization strategy consists of the combined effect of the Mean-Square Error loss (regularizes the reconstruction capability of the autoencoder), Classifier loss (regularizes the latent space such that the representations are discriminative with respect to the class labels ), and the Supervised Contrastive loss (regularizes the latent space such that latent space vectors from the same class are close to each other and the latent vectors from different classes are far away from each other). This strategy plays a crucial role in ensuring that the learned latent representations effectively capture the underlying structures and variations in the data, particularly emphasizing the preservation of distinctions between different classes. By leveraging this regularization mechanism, our method aims to maintain a balanced and discriminative representation space, thereby enhancing the integrity and fidelity of the learned feature vectors across diverse data distributions (Our autoencoder (Figure~\ref{fig:1b}, \ref{fig:1d}) shows better class separation compared to the vanilla autoencoder (Figure~\ref{fig:1a}, ~\ref{fig:1c}) for both balanced and imbalanced data.).

 \textbf{\textit{Stage 2: Synthesis}}: 
 Our novel data-calibration method is introduced at this stage and is premised on assuming that feature vectors and samples from each category conform to Gaussian distributions. Leveraging the statistical characteristics of the majority classes alongside the disentangled representations, we generate new latent vectors that aim to capture the potential distribution of the underrepresented class if the dataset were balanced. This synthesis ensures that the generated samples not only exhibit authenticity but also encompass the entire range of patterns within the minority class (In Figure~\ref{fig:1e}, we observe our data synthesis mechanism from the imbalanced latent vectors (represented in Figure~\ref{fig:1d}). Figure~\ref{fig:1b} shows the t-SNE when the balanced data is trained using our Autoencoder. The goal of our data synthesis is to achieve this scenario.).

\color{black}
The major contributions of our research are as follows:
    \textbf{(1)} We introduce \textbf{a novel calibration method to estimate the distribution} of minority class samples (Section~\ref{proposed_calib}). By leveraging local characteristics from sufficiently large classes, this method enables more accurate modeling and generation of underrepresented classes, thereby improving overall model performance.
    \textbf{(2)} Unlike traditional approaches focusing mainly on the majority class, our method also incorporates intermediate classes (Section~\ref{data_gen}). Accounting for their characteristics yields more faithful and inclusive distribution estimation, resulting in robust dataset modeling.
    \textbf{(3)}  We propose a regularization strategy that captures class-specific traits while enlarging inter-class margins. This fosters a more discriminative feature space, supporting improved, diverse, and representative sample generation.
    \textbf{(4) }We validate our contributions on\textbf{ multiple modalities} such as \textbf{image, text, and tabular datasets,} benchmarking against state-of-the-art methods. Results consistently show superior accuracy and generalization, demonstrating both the effectiveness and practical relevance of our approach.
\color{black}


 The remainder of this paper is organized as follows: Section 2 reviews existing class imbalance techniques; Section 3 introduces our two-stage oversampling method; Section 4 reports results across datasets; Section 5 presents an ablation study; and Section 6 concludes with findings and future directions.
\section{Related Work}
\label{sec:related_work}
Approaches to class imbalance are broadly categorized into data-level, algorithm-level, and hybrid methods \cite{das2022supervised}. Data-level techniques, particularly oversampling, are a standard remedy. Classical oversampling algorithms, such as RWO~\cite{ZHANG201499}, SMOTE \cite{chawla2002smote} and its variants (e.g., Borderline-SMOTE \cite{han2005borderline}, ADASYN \cite{Adasyn_2008}), generate synthetic minority samples via interpolation. While effective for low-dimensional data, their reliance on distance metrics makes them less suitable for high-dimensional domains like images.
For complex data, deep generative models offer a powerful alternative. Models such as Variational Autoencoders (VAEs) \cite{kingma2013auto} and Generative Adversarial Networks (GANs) \cite{goodfellow2014generative} can synthesize high-fidelity minority samples. Conditional GANs \cite{gauthier2014conditional} and BAGAN \cite{mariani2018bagan} generate class-specific samples, while integrated frameworks like GAMO \cite{mullick2019generative} jointly optimize data generation and classifier training in an adversarial game. A significant advancement is DeepSMOTE \cite{dablain2021deepsmote}, which performs oversampling in a trained autoencoder's latent space, bridging the gap between feature representation and sample generation. Other notable works include DGC \cite{wang2020deep} and DGCMM \cite{wang2022deep}, which use deep generative processes with Bayesian inference to model complex class distributions.

Our work builds upon these deep generative oversampling strategies. However, unlike methods that generate samples using convex combinations of all minority instances (e.g., \cite{mullick2019generative, pmlr-v206-mondal23a}), or those that do not fully account for the underlying data manifold, our approach proposes a more efficient and distribution-aware generation process. A detailed discussion of related works is provided in the Supplementary section \ref{supp:related_work}.

\textbf{What is Distribution Calibration?}  Distribution Calibration, in the context of few-shot learning introduced by Yang et al. in ~\cite{yang2021free}, refers to a strategy for calibrating the distribution of classes with few training examples by transferring statistics from classes with sufficient examples. These calibrated statistics are then used to generate feature vectors corresponding to each element in the minority class by sampling from the calibrated Gaussian distributions. This allows for generating more diverse and accurate feature distributions, which can improve the performance of few-shot learning models.
Vigneswaran et al. ~\cite{vigneswaran2021feature} proposed TailCalibX, which follows a similar strategy of estimating the parameters of the Gaussian Distribution corresponding to each sample in the minority class based on the nearest neighbor approach and samples an equal number of samples from a particular class for Imbalanced Datasets. Another calibration approach, as proposed in \cite{10191975}, to tackle imbalance, considers the concept of the fractional norm to calculate K-nearest neighbor and adaptive weighing to calculate a calibrated distribution for the minority class. 
\textcolor{black}{The existing calibration method's shortcomings in handling class imbalance lie in the fact that the method uses $k$-nearest classes for generating the samples in the minority class. By considering the nearest class feature vectors, the minority class elements are sampled so that they learn the nearest class in a better way than the furthest class, which in turn may affect the inherent characteristics of the underlying minority class. Hence, to tackle these shortcomings, we present a method that does not focus on the overall characteristics of the nearby class but considers the local characteristics of those classes.}

\section{Proposed Methodology}

\subsection{Preliminaries}
\label{prelims}

In Table~\ref{tab:notations} we described the notations we have used in this manuscript.
\begin{table}[!ht]
\caption{Table of Notations}
\label{tab:notations}
\resizebox{\textwidth}{!}{%
\begin{tabular}{cl}
\hline
\textbf{Notations}                                                           & \textbf{Descriptions}                                                                                                                                               \\ \hline
\textcolor{black}{$X$}                                                                & \textcolor{black}{Set of all Images, Tabular or Textual data}                                                                                                                                                   \\
$\mathcal{Y}$                                                                & Set of all Labels                                                                                                                                                   \\
$\mathscr{L}$                                        & Latent space feature vectors                                                                                                                                        \\
$x$                                                                          & An image data                                                                                                                                                       \\
$y$                                                                          & Label corresponding to the image data $x$                                                                                                                           \\
$\mathbb{R}$                                         & Set of real numbers                                                                                                                                                        \\
$\mathscr{M}$                                        & Majority Class                                                                                                                                                      \\
$\Hat{\mathscr{I}}$                                        & Set of Intermediate Classes                                                                                                                                                  \\
$\mathscr{I}$                                        & Intermediate Class $\mathscr{I} \in \Hat{\mathscr{I}}$                                                                                                                                                  \\
$\Hat{\mathfrak{m}}$                                   & Set of Minority Classes                                                                                                                                                      \\
$\mathfrak{m}$                                   & Minority Class $\mathfrak{m} \in \Hat{\mathfrak{m}}$                                                                                                                                                     \\
$d$                                                                          & dimension of the feature space                                                                                                                             \\
$n$                                                                          & Number of samples in the training set                                                                                                                               \\
$|.|$                                            & The cardinality of a set
                    \\
$\mathcal{L}_{AC}$                                   & Class Aware Loss                                                                                                                                                    \\
$\mathcal{L}_{CG}$                                   & Classifier Guidance Loss                                                                                                                                            \\
$\mathcal{L}_{CS}$                                   & Supervised Contrastive loss                                                                                                                                         \\
$(\mu^{\prime}_i, \Sigma^{\prime}_i)$                 & Mean and variance of the majority class corresponding to the $i^{th}$ GMM component      
\\
$(\tilde{\mu}_i,\tilde{\Sigma}_i)$                   & Mean and variance of the intermediate class ($\mathscr{I}$) corresponding to the $i^{th}$ GMM component      
\\
$(\bar{\mu}_i,\bar{\Sigma}_i)$                       & Estimated Mean and variance corresponding to the $i^{th}$ element in the minority class           
\\
$\mu_{\mathscr{M}}$             & Collection of the mean of each GMM component in the Majority class
\\
$\Sigma_{\mathscr{M}}$        & Collection of the variance of each GMM component in the Majority class
\\
$\mu_{\Hat{\mathscr{I}}}$       & Collection of the mean of each GMM component in all the Intermediate classes                                                                               
\\
$\Sigma_{\Hat{\mathscr{I}}}$      & Collection of variance of each GMM component in all the Intermediate classes
\\
$\mu_{\mathscr{I}}$       & Collection of the mean of each GMM component in an Intermediate class  $\mathscr{I}$, where $\mathscr{I}\in\Hat{\mathscr{I}}$                                                                              
\\
$\Sigma_{\mathscr{I}}$      & Collection of variance of each GMM component in an Intermediate class $\mathscr{I}$, where $\mathscr{I}\in\Hat{\mathscr{I}}$   
\\
\textcolor{black}{$(\hat{\mu}_i, \hat{\Sigma}_i)$}                       & \textcolor{black}{Mean and variance of the $i^{th}$ GMM component from the collection of the mean and variance of the Intermediate and Majority classes represented by $\hat{\mu}$ and $\hat{\Sigma}$ respectively.}                                                                  \\
$K$                                                                          & Number of class                                                                                                                                                     \\
$k$                                                  & Number of neighbours                                                                                                                                                \\
$\eta$                                                  & Imbalance tuner hyperparameter                                                                                                                                                \\
$\zeta$                                                  & the threshold cardinality which determines the classes that belong to the collection intermediate classes, $\mathscr{I}$ and the collections of minority classes $\mathfrak{m}$                                                                                                                                                \\
$N_i$                                                                        & Number of elements in the $i^{th}$ class                                                                                                                             \\
$\xi_i$                                                                      & Number of components in the $i^{th}$ class                                                                                                                           \\
$S_i$                                                & Number of elements in the $i^{th}$ component                                                                                                                         \\
$p_i$                                                                        & Probability of a feature vector belonging to the $i^{th}$ class                                                                                                                     \\
$\hat{x}$                                                                    & Representation of the reconstructed image $x$                                                                                                                                      \\
$l$                                                                          & a feature vector in the latent space $\mathscr{L}$                                                                                                                                 \\
$L_i$                                                   & Feature vectors of the $i^{th}$ class from $\mathscr{L}$
                                    \\
$\tilde{L}$                                                        & Synthetic feature vectors sampled from a given normal distribution
                                    \\
$\hat{L}$                                                                    & Set of oversampled feature vectors                                                                                                                                         \\
$\hat{Y}$                                                                    & Labels of the oversampled feature vectors                                                                                                                           \\
$w_j$                                                                        & Weight associated to mean $(\hat{\mu}_j)$ or variance $(\hat{\Sigma}_j)$ for estimating the mean $(\bar{\mu}_i)$ and sigma $(\bar{\Sigma}_i)$ in the minority class \\
$W$                                                                         & Collection of the weights $w_i$
\\
$n_i$                                                                    & The number of samples that have to be generated in the $i^{th}$ class                                                                                                \\
$\hat{n}_i$                                                                  & The number of samples that have to be generated corresponding to each element in the $i^{th}$ class                                                                  \\
$t$                                                                          & Temperature hyperparameter for training the autoencoder                                                                                                                                     \\
$\mathcal{Q}_i$                                                              & Samples in the training set which have labels $y_i$                                                                                                                 \\
$Z(i)$                                                                       & Samples in the training set excluding the $i^{th}$ sample                                                                                                           \\
$G$                                                                          & Group                                                                                                                                                               \\
$G_i$                                                                        & $i^{th}$ subgroup formed by decomposition of group $G$                                                                                                             \\
$*$                                                                          & Identity element in the group $G$                                                                                                                                   \\
$\mathcal{T}$                                                                & Linear transformation $\mathcal{T}: X \rightarrow \mathscr{L}$                                                                                                                               \\
$f$, $\tilde{f}$                                                             & Equivariant map                                                                                                                                                     \\
$*_{g}$, $*_{sg}$                                                            & Group action                                                                                                                                                        \\
$\rho$                                                                       & Map from the set of images to the set of labels                                                                                                                            \\ \hline
\end{tabular}
}
\end{table}
\subsubsection{Autoencoder for feature representation}
The autoencoder network consists of an encoder network (\(\mathscr{E}\)) and a decoder network (\(\mathscr{D}\)). Images from the image space (\(X\)) are fed into the encoder network (\(\mathscr{E}\)) to obtain latent representations in the latent space (\(\mathscr{L}\)). The decoder network (\(\mathscr{D}\)) takes these latent representations from the latent space (\(\mathscr{L}\)) as input to reconstruct the images in the reconstructed image space (\(\hat{X}\)). The objective is to make the original images \(X\) as similar as possible to the reconstructed images \(\hat{X}\), such that \(X \approx \hat{X}\). 

\subsubsection{Gaussian Mixture Model (GMM) for a Class}
We can cluster the data points belonging to a particular class in a dataset as a mixture of Gaussian distributions for a parametric data distribution. Let the latent vectors corresponding to the $c^{th}$ class be $L_c = \{l_{c1}, l_{c2},\cdots, l_{cN_{c}}\}$. We perform GMM for class $c$ on the latent feature vectors $L_c$ with the number of components as $\xi_c$. Then the probability density for the sample $l_{cj}$ can be represented as:
\begin{equation}
    f(l_{cj},\Theta) = \sum_{i=1}^{\xi_c}\alpha_if_i(l_{cj},\theta_i),
\end{equation}
where $\alpha_i$ represents the mixture coefficients, the parameter space $\Theta = (\theta_1,\theta_2,\cdots,\theta_{\xi})$, the Gaussian density function corresponding to the parameter $\theta_i$ is $f_i$. 

Suppose we define the parameters corresponding to each component as $\theta_i = (\mu_i,\Sigma_i)$. Then GMM cluster, $f_i(l_{cj},\theta_i)$ is represented by: 
\begin{equation}
    f_i(l_{cj},\theta_i) = \frac{1}{\sqrt{2\pi}|\Sigma_i|}e^{-\frac{1}{2}(l_{cj}-\mu_i)^T\Sigma_i^{-1}(l_{cj}-\mu_i)}.
\end{equation}

We define the number of elements in each component as $S_i$, where $S_i$ is the $i^{th}$ element in the array $S$ (the set containing the number of elements in each mixture component). For our experiments, we considered the number of elements in each component, $S_i$, to be greater than the number of elements in the class having the fewest elements (i.e., $N_K$). Hence, for the $c^{th}$ class the number of components $\xi_c = \frac{N_c}{N_K}$.

\subsection{Problem Definition}
Consider the training dataset $X = {(x_i,y_i)}_{i=1}^{n}$ for a $K$-multi-class imbalance classification problem, where $n$ is the number of samples, $x_i$ is the $i^{th}$ image data, and $y_i$ is the corresponding label. $x_i \in \mathbb{R}^{C\times h\times w}$ and $y_i \in \mathcal{Y}$, where $\mathcal{Y}=\{1,2,\cdots,c,\cdots, K\}$ set of labels and $C, h, w$ are the number of channels, height, and the width of the images respectively. Let $X_{c}$ denote the set of all data points in the class $c$, and the cardinality of the set $X_{c}$, denoted by $N_c = |X_{c}|$. \textcolor{black}{The classes are labelled such that $N_1\geq \cdots \geq \cdots \geq \cdots \geq N_K$.}
Given $X$, in our first stage, our objective is to learn a latent space $\mathscr{L}\in \mathbb{R}^{d}$ with a corresponding distribution $q(l)$. The set of latent vectors corresponding to the datasets $X$ is represented by 
$\mathscr{L}=\{ l_{1},\cdots, l_{i},\cdots, l_{n}\}$.  The set of latent vectors corresponding to the particular class $c$ is represented by $L_{c}$, and the corresponding mean and variance of latent vectors in class $c$ are represented as $\mu_{c}$ and $\Sigma_{c}$.
\textcolor{black}{For the Image datasets, the Autoencoder network that we have utilized can be formally represented as $\mathscr{E}$ (encoder), $\mathscr{D}$ (decoder), and $\mathscr{H}$ (latent vector classifier). 
Using the learned latent representation, we aim to oversample the imbalanced feature vectors in the latent space using our proposed synthetic data oversampling algorithm. The reconstructed vectors obtained from the decoder are represented as $\hat{X}$, and the reconstructed vectors corresponding to the class $c$ are represented by $\hat{X}_c$. 
We employ a discriminative classifier network, denoted as $\mathscr{F}$, for comparison with the state-of-the-art methods. }

\textcolor{black}{  For the tabular data, we use the input data as a feature vector instead of obtaining the latent vectors from the trained autoencoder. Thus, the tabular data $X = (x_i,y_i)_{i=1}^n$ is same as $(l_i,y_i)_{i=1}^n$, where $l_i$ is the latent vector or the feature vector.}


\textcolor{black}{The sequence of the subsequent sections is detailed as follows,
subsection ~\ref{proposed_over_method} outlines our data calibration method, and Subsection~\ref{sparcity_autoencoder} expresses the details of training the Autoencoder.}
\subsection{RCS Oversampling Method}
\label{proposed_over_method}
Our proposed oversampling algorithm will oversample/generate new latent vectors of the minority classes in the latent space. The algorithm tries to estimate the statistics related to the minority classes using the statistics of the majority class and the intermediate classes. This methodology 
will help meet our second principal requirement - pattern coverage. 
\textcolor{black}{When we use the feature vectors (for tabular data) and the latent vectors (for image data), both represented by $\mathscr{L}$, we refer to them collectively as feature vectors for simplicity and to avoid confusion. In this section, the term feature vectors will be used to denote $\mathscr{L}$, and all definitions and operations applied to feature vectors will equally apply to the latent vectors used for image data. We assume that the feature vectors corresponding to different classes are well separated to facilitate effective oversampling.}

In simpler terms, this approach takes information about the common patterns in the majority data and uses that knowledge to create new, valid data points for the minority class. This helps address the issue of having limited data for the minority class while ensuring that the generated data reflects the actual characteristics of that class. Here's an analogy: Imagine you have a collection of pictures of different colored dogs. There are many pictures of brown dogs (the majority class), but only a few pictures of black dogs (the minority class). This algorithm would analyse the feature vectors of the brown dogs (fur texture, body shape) and use that information to create new, realistic pictures of black dogs, filling the gaps in your data collection.
Using statistics from classes with significantly more samples is beneficial because similar classes typically exhibit comparable means and variances in their feature representations \cite{yang2021free}. As a result, the mean and variance of a Gaussian distribution can be applied to similar classes \cite{salakhutdinov2012one}. Furthermore, a sufficient number of samples improves the accuracy of these statistical estimates for any given class. 

Our method also uses the statistics of classes with sufficient samples to estimate the statistics of minority classes. Instead of using statistics from the entire class, we utilise the statistics from the components or subclasses of the class with sufficient samples to better capture the local patterns. Our class division consists of three categories: the Majority class, the Intermediate class, and the Minority class. Intermediate classes typically comprise those classes with fewer samples than the majority class. The lower bound for the number of samples in the intermediate classes is determined by the hyperparameter $\zeta$. This ensures that the classes included in the set of intermediate classes possess sufficient information to cover a diverse range of patterns within the class itself. We achieve this by leveraging the statistics of components or subclasses within those classes and using these statistics to generate new samples for these classes. 
The remaining classes are considered minority classes, and their distribution is calibrated by considering the statistics of components from nearby majority and intermediate classes. The subsequent sections discuss the description of class categorisation and the handling of different class details.


\begin{figure}[!ht]
\scriptsize
  \centering
    \includegraphics[width=0.9\textwidth]{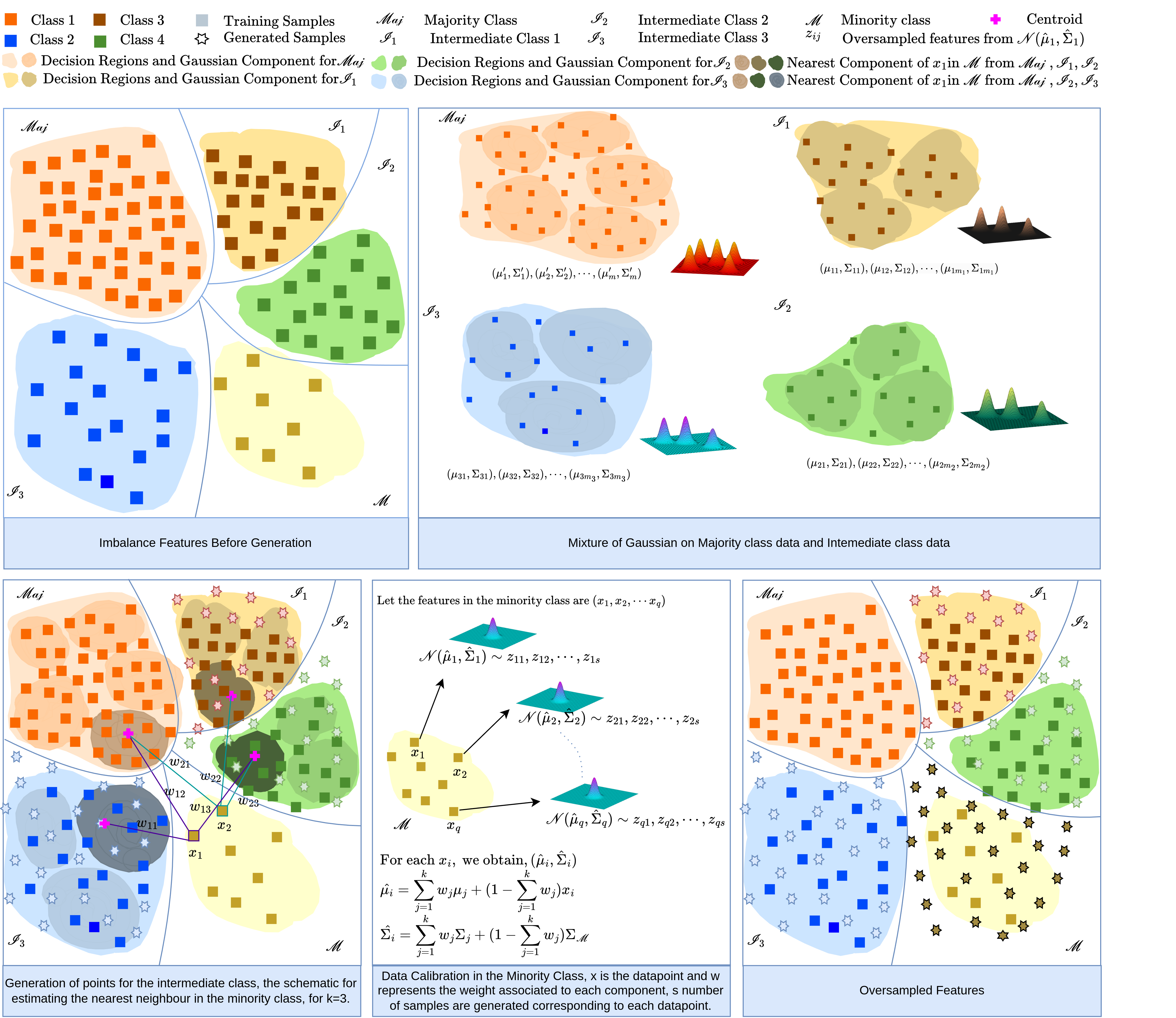}
  \caption{Illustrations of our Data Calibration method, the first figure (from top left) represents the imbalanced feature vectors, the second figure represents the Gaussian Mixture components of the Majority class as well as the Intermediate classes, the third figure represents the oversampling in the latent space, and the visualization of the components participating in estimating the mean and variance corresponding to each point of the minority class from which synthetic feature vectors are sampled from, in the fourth figure corresponding to the estimated mean and variance we sample synthetic feature vectors from this Gaussian distribution and in the fifth figure we display the oversampled feature vectors. }
  \label{fig:calib_motivation}
\end{figure}
\subsubsection{Class-level Categorization for Synthetic Data Generation}
\label{data_gen}

 We start with the imbalanced feature vectors ($\mathscr{L}$) for the synthetic data generation. Based on the sample sizes of different classes in $\mathscr{L}$, we conglomerate the majority, intermediate, and minority classes. The term $\eta$ refers to the imbalance tuner. The imbalance tuner hyperparameter is used to compute the threshold cardinality by defining $\zeta$ as $\frac{N_{1}}{\eta} $, where $N_{1}$ is the number of elements in the majority class. The threshold cardinality $\zeta$ decides whether a particular class would be considered an intermediate or minority class.
 The classes whose sample size is greater than $\zeta$ and less than $N_1$ (i.e., the cardinality of the majority class as defined in our problem definition) are coined as the Intermediate class, and the classes whose sample size is less than $\zeta$ are considered as the Minority class. \textcolor{black}{Thus, based on the cardinality, classes are divided into three groups, namely, the majority class abbreviated as $\mathscr{M}$ and the Intermediate classes  $\Hat{\mathscr{I}}$ and the minority classes as $\Hat{\mathfrak{m}}$. The three groups are handled in the following way:} 
 

\begin{enumerate}[label=(\alph*), itemindent=0pt, topsep=0pt,partopsep=0pt,parsep=0pt,itemsep=0pt]
\item \textit{Handling the Majority Class: }
We perform the GMM clustering for the majority class ($\mathscr{M} $) to group the class into several mixture components and find the statistics mean($\mu_i$) and covariance ($\Sigma_i$) for them.

\item \textit{Handling the Intermediate Classes: }
Consider an Intermediate class $\mathscr{I}$ from the set of all the intermediate classes $\Hat{\mathscr{I}}$.
For each intermediate class $\mathscr{I}$, we perform the GMM clustering on the feature space. We obtain the parameters mean($\mu_i$) and covariance ($\Sigma_i$) from each component in this feature space. 
Without loss of generality, let us assume that the corresponding intermediate class is the $i^{th}$ class. We first compute the number of samples that have to be generated in the $i^{th}$ class as $n_i = N_{1} - N_{i}$. Corresponding to each component in the $i^{th}$ class, we generate $\hat{n}_i$ many samples, where $\hat{n}_i = n_i/\xi_i$, $\xi_i$ being the total number of components in the $i^{th}$ class.
In other words we generate $\hat{n}_i$ synthetic samples from the Gaussian distribution $\mathscr{N}(\mu_{ij},\Sigma_{ij})$, where $\mu_{ij},\Sigma_{ij}$ are the mean and covariance of the $j^{th}$ component in the $i^{th}$ class. Algorithm~\ref{alg:example1} provides a pseudocode of Generating Synthetic Samples in each Intermediate class $\mathscr{I}$.

\begin{algorithm}[!ht]
\scriptsize
   \caption{Pseudocode of Generating Samples in Intermediate class ($\zeta \leq \#Samples < N_{1}$)}
   \label{alg:example1}
   {\bfseries Input:} $\mathscr{L}$ is set of feature vectors, $Y$ set of labels corresponding to $\mathscr{L}$, $N_i$ is the number of elements in the $i^{th}$ class and the class having the maximum number of elements is $N_1$, hyperparameter $\eta$.\\
   {\bfseries Initialize:} $\Hat{L}, \Hat{Y}, \Hat{\mu}, \Hat{\Sigma}, S$ as empty sets, $\xi_i$ as the number of Gaussian Mixture components corresponding to the $i^{th}$ class with feature vectors $L_i$.\\
   {\bfseries Output: }The set of oversampled feature vectors $\Hat{L}$ and the corresponding labels $\Hat{Y}$ for the intermediate classes.\\
   \hrulefill \\
   {\bfseries Compute:}
\begin{algorithmic}[1]
   \scriptsize


   \STATE $\zeta = \frac{N_{1}}{\eta}$\\
 \hspace{6.5cm} \color{brown} \tcc{For loop running over all the classes}\ \color{black}
 \vspace{-0.4cm}
   \FOR{$i\text{ \bfseries from }1$ {\bfseries to} $K$} 
   \STATE $L_i, Y_i$  $\leftarrow$  obtains the feature vectors and labels of the $i^{th}$ class from $\mathscr{L},Y$.
   \STATE $n_i = N_{1} - N_i$  
   \IF{$N_i >= \zeta$ \textbf{and} $N_i < N_1$}
   \STATE $\tilde{\mu}_i$, $\tilde{\Sigma}_i,\xi_i, S_i = GaussianMixture(L_i,Y_i)$ \hspace{-0.2cm} \color{brown} \Comment*[h]{Returns the Gaussian Mixture Components}\  \color{black}
    \STATE $\hat{n}_i = n_i/\xi_i$ \\
    \hspace{-0.8cm} \color{brown}\tcc{For loop generating synthetic sampled from $\mathscr{N}(\tilde{\mu}_i, \tilde{\Sigma}_i)$ in the Intermediate Class $\mathscr{I}$}\ \color{black}
    \vspace{-0.4cm}
       \FOR{$j\text{ \bfseries from }1$ {\bfseries to} $\xi_i$}
            \STATE $\tilde{L}_j \leftarrow $ $\hat{n}_i$ synthetic samples, sampled from the $j^{th}$ Gaussian component with mean $\tilde{\mu}_{ij}$  and sigma $\tilde{\Sigma}_{ij}$.
            \STATE $\tilde{Y}_j \leftarrow$ array of size $\hat{n}_i$ containing the $i^{th}$ class index.
            \STATE $\hat{L} \leftarrow \tilde{L}_i, \hat{Y} \leftarrow \tilde{Y}_i$
        \ENDFOR
   






    \STATE $\Hat{\mu} \leftarrow \tilde{\mu}_i, \Hat{\Sigma} \leftarrow \tilde{\Sigma}_i$. 

    \STATE $S \leftarrow S_i$
    
    \ENDIF

    \ENDFOR\\
    \STATE {\bfseries return} $\hat{L}, \hat{Y}$
\end{algorithmic}
\end{algorithm}
\begin{remark}
        The Algorithm $\ref{alg:example1}$ depicts the generation of synthetic samples. In this algorithm, the samples of the intermediate class are generated using each of the Gaussian Mixture components of the Intermediate Class. Gaussian mixture function returns the number of Gaussian components $\xi_i$, the number of samples in each component $S_i$, and   $\tilde{\mu}_i = \{\tilde{\mu}_{i1},\tilde{\mu}_{i2},\cdots, \tilde{\mu}_{i\xi_i}\}$, $ \tilde{\Sigma}_i = \{\tilde{\Sigma}_{i1}, \tilde{\Sigma}_{i2},\cdots, \tilde{\Sigma}_{i\xi_i}\}$ represents the collection of the mean and variance corresponding to each component in the $i^{th}$ class, $\hat{\mu}$ stores all the mixture components associated to each Intermediate class $(\mathscr{I} \in \Hat{\mathscr{I}})$, and the Majority class $\mathscr{M}$. $\hat{L}, \hat{Y}$ represents the oversampled feature vectors and labels. This algorithm returns $\hat{L} \text{ and } \hat{Y}$ after the oversampling has been performed.
\end{remark}
\item \textit{Handling the Minority Classes: }
If the number of samples in a particular class is less than $\zeta$, we consider those classes as minority classes. From the set of the minority classes $\Hat{\mathfrak{m}}$ we consider $\mathfrak{m}\in\Hat{\mathfrak{m}}$.
Without loss of generality, let us assume that the corresponding minority class $\mathfrak{m}$ is the $i^{th}$ class. 
We first compute the number of samples that have to be generated in the $i^{th}$ class as $n_i = N_{1} - N_{i}$. Corresponding to each datapoint in the $i^{th}$ class, we generate $\hat{n}_i$ many samples, where $\hat{n}_i = n_i/N_i$, $N_i$ being the total number of samples in the $i^{th}$ class. For generating data in the minority classes, we use the data calibration technique discussed in the subsection ~\ref{proposed_calib}. 
\end{enumerate}
Figure \ref{fig:calib_motivation} illustrates the handling of various classes, including the oversampling techniques applied to intermediate and minority classes.

\subsubsection{Distribution Calibration Method for Oversampling Minority Instances}
\label{proposed_calib}

Suppose $l_i$'s are the latent vectors (or feature vectors) of the samples $x_i$'s of the minority class $\mathfrak{m} \in \hat{\mathfrak{m}} $. Assume \textcolor{black}{$(\mu^{\prime}_1,\Sigma^{\prime}_1), (\mu^{\prime}_2,\Sigma^{\prime}_2),\cdots,(\mu^{\prime}_{\xi_s},\Sigma^{\prime}_{\xi_s})$} be the statistics corresponding to the majority class $\mathscr{M}$, and $(\tilde{\mu}_{i1},\tilde{\Sigma}_{i1}), (\tilde{\mu}_{i2},\tilde{\Sigma_{i2})},\cdots,$ $(\tilde{\mu}_{\xi_i},\tilde{\Sigma}_{\xi_i})$ are statistics corresponding to the intermediate classes in $\Hat{\mathscr{I}}$. We conglomerate both the parameters of the majority class as well as intermediate classes to obtain $(\hat{\mu}, \hat{\Sigma}) = (\mu^{\prime}_1,\Sigma^{\prime}_1), (\mu^{\prime}_2,\Sigma^{\prime}_2),\cdots,(\mu^{\prime}_s,\Sigma^{\prime}_{\xi_s}) \bigcup (\tilde{\mu}_{i1},\tilde{\Sigma}_{i1}),$ $ (\tilde{\mu}_{i2},\tilde{\Sigma_{i2})},\cdots,(\tilde{\mu}_{\xi_i},\tilde{\Sigma}_{\xi_i}) = (\hat{\mu}_{1},\hat{\Sigma}_{1}), (\hat{\mu}_{2},\hat{\Sigma}_{2}),\cdots \cdots$. 
Among the means $\hat{\mu}$, we consider the $k$-neighbors corresponding to the datapoint $l_i$. Let $(\hat{\mu}_1,\hat{\Sigma}_1), (\hat{\mu}_2,\hat{\Sigma}_2),$ $\cdots,(\hat{\mu}_k,\hat{\Sigma}_k)$ be statistics corresponding to the k neighbours to the data-point $l_i$ among the means $\hat{\mu}_1,\hat{\mu}_2,\cdots,\hat{\mu}_k$. Then for each such $l_i$ we calibrate the mean ($\bar{\mu}_i$) as:
\begin{equation}
    \bar{\mu}_i = \sum_{j=1}^{k}w_j\hat{\mu}_j+(1 - \sum_{j=1}^{k}w_j)l_i,
\end{equation}
where,
$w_i$'s are the weights associated to each $\mu_i$, and the weight associated to $l_i$ is given by ($1 - \sum_{j=1}^{k}w_j$).\\

The variance ($\bar{\Sigma}_i$) associated with the minority class sample  ($l_i \in \mathfrak{m}$) is given by
\begin{equation}
\bar{\Sigma}_i = \sum_{j=1}^{k}w_j\hat{\Sigma}_j+(1 - \sum_{j=1}^{k}w_j)\Sigma_{\mathfrak{m}},
\end{equation}

where,
$w_i$'s are the weights associated to each $\Hat{\Sigma_i}$, and the weight associated to $l_i$ is given by ($1 - \sum_{j=1}^{k}w_j$).\\

We define the weight $w_i$ corresponding to the ${i^{th}}$ component $(\Hat{\mu}_i,\Hat{\Sigma}_i)$  as:
\begin{equation}
    w_i = \frac{1}{S_{i}}.
\end{equation}
Theorem~\ref{thm} gives the theoretical consistency and shows that the estimated mean $\bar{\mu}_j$ is well-defined and states that the estimated mean corresponding to each feature in the minority class $\mathfrak{m}$ lies in the convex hull regions of the $\hat{\mu}_i's$ obtained from the Gaussian Mixture Models of the Majority Class $\mathscr{M}$ and each of the Intermediate Class $\mathscr{I}\in \Hat{\mathscr{I}}$. The estimated mean $\bar{\mu}_j$ lies closer to $l_j$. This guarantees that the mean of the estimated distribution not only retains the inherent characteristics of the data but also acquires local characteristics of the nearby classes whose components lie closer to $l_j$.
\begin{theorem}
    \label{thm}
    Let $l_1,l_2,\cdots,l_n$ be the samples in the minority class $\mathfrak{m}$. Let $\bar{\mu}_j$ be the calibrating mean associated with each $l_j$, having the form $\bar{\mu}_j = l_j + \Sigma_{i=1}^{k}(\hat{\mu}_i - l_j)w_i$. Then $\bar{\mu}_j$ resides in the convex hull region of the $\hat{\mu}_i$'s and $l_j$. Moreover, the estimated mean $\bar{\mu}_j$ lies closer to $l_j$.
\end{theorem}

\begin{proof}
   The proof of Theorem \ref{thm} is provided in the Supplementary Section \eqref{app_thm1}. 
\end{proof}
Now, the question arises whether the mean of the samples generated in the minority class is a good estimate of the mean of the original distribution of the minority class. Theorem~\ref{thm1} states that the sample mean of the elements sampled from the distribution $\mathscr{N}(\bar{\mu}_j,\bar{\Sigma}_j)$ is the unbiased estimator of the population mean $\bar{\mu}_j$.
\begin{theorem}
    \label{thm1}
    Let $l^{\prime}_1,l^{\prime}_2,\cdots,l^{\prime}_k$ be the samples generated from $\mathscr{N}(\bar{\mu}_j,\bar{\Sigma}_j)$. Then, the sample mean $l^{\prime}$ of the generated points $ l^{\prime}_i$ is the unbiased estimator of the population mean $\bar{\mu}_j$.
\end{theorem}
\begin{proof}
   The proof of Theorem \ref{thm1} is provided in the Supplementary section \ref{app_thm2}. 
\end{proof}

\begin{algorithm}[!ht]
    \scriptsize
   \caption{Pseudocode of Generating Samples in Minority Class($\#Samples < \zeta$)}
   \label{alg:example2}
    {\bfseries Input: } Imbalanced feature vectors and labels as $\mathscr{L}$ and $Y$ respectively.  $\Hat{\mu}$, $\Hat{\Sigma}$, and $S$ as the mean, covariance, and the sample size corresponding to various components.\\
    {\bfseries Initialize: } $\mu_{\mathfrak{m}}, \Sigma_{\mathfrak{m}}$ as empty. \\
   {\bfseries Output: } The Oversampled feature vector $\hat{L}$ and the corresponding labels as $\hat{Y}$. \\
   \hrulefill \\
   {\bfseries Compute:}
   
\begin{algorithmic}[1]
   \scriptsize
\STATE \textcolor{black}{idx $\leftarrow \phi$ }\\
\STATE \textcolor{black}{flag $\leftarrow$ 0}\\
 \hspace{6.2cm} \color{brown} \tcc{For loop running over all the classes.}\ \color{black}
 \vspace{-0.4cm}
    \FOR{$i\textbf{ from }1$ {\bfseries to} $K$}
  
    \STATE $L_i, Y_i$  $\leftarrow$  obtains the feature vectors and labels of the $i^{th}$ class from $\mathscr{L}$, $Y$.
    
    \STATE $n_i = N_{1} - N_i$

    \IF{$N_i < \zeta$}
        \IF{$flag == 0$} 

            \FOR{$j\text{ from }1$ {\bfseries to} $|S|$}
            \STATE \textcolor{black}{$W[j] \leftarrow 1/S_j$}  \color{brown} \hspace{0.1cm}\Comment*[h]{The weights are assigned to each component in this loop.}\ \color{black}
            \ENDFOR
            \STATE flag = 1
        \ENDIF
        \STATE $\hat{n}_i = n_i/N_i$
        \STATE $\mu_{\mathfrak{m}}, \Sigma_{\mathfrak{m}} \leftarrow \text{ Gaussian Distribution parameters corresponding to } L_i$\\
\hspace{2.3cm} \color{brown} \tcc{Loop running over all the elements in the Minority Class.}\ \color{black}
\vspace{-0.4cm}
        \FOR{$l$ {in} $L_i$}
            \STATE Initialize $\Hat{\mu}_l$,$\Hat{\Sigma}_l$
            
            \STATE $idx$ $\leftarrow$ the index of the $k$ nearest neighbors in $\Hat{\mu}$
            \STATE Initialize $\bar{\mu}_l$ as $0$ and $\bar{\Sigma}_l$ as $0$\\
            \color{brown} \hspace{6.1cm}\Comment*[h]{Loop for calculating the Mean.}\ \color{black}
            \FOR{$j$ {in} $idx$}
                \STATE $\bar{\mu}_l = \bar{\mu}_l + W[j]\times \Hat{\mu}[j]$
            \ENDFOR
            \STATE $w = 1 - \sum_{i=1}^{|S|} W$ 
            \STATE $\bar{\mu}_l = \bar{\mu}_l + w\times l$
            \\
            \color{brown} \hspace{5.2cm}\Comment*[h]{Loop for calculating the Covariance.}\ \color{black}
            \vspace{-0.4cm}
            \FOR{$j$ {in} $idx$}
                \STATE $\bar{\Sigma}_l = \bar{\Sigma}_l + W[j]\times \Hat{\Sigma}[j]$
                
            \ENDFOR
            \STATE $\bar{\Sigma}_l = \bar{\Sigma}_l + w\times \Sigma_{\mathfrak{m}}$\\
            \hspace{0.6cm} \color{brown} \tcc{Loop for generating samples from the estimated Gaussian parameters.}\ \color{black}
\vspace{-0.4cm}
            \FOR{$j$ {from} 1 {to} $\hat{n}_i$}
                \STATE $\tilde{L}_{gen} \leftarrow $ Randomly drawn sample from a normal distribution with mean and covariance as $\bar{\mu}_l$ and $\bar{\Sigma}_l$ respectively.
            \ENDFOR
            \STATE $\tilde{Y}_{gen} \leftarrow$ array of size $\hat{n}_i$ containing the $i^{th}$ class index.
        \ENDFOR
        \STATE $\Hat{L} \leftarrow \tilde{L}_{gen}, \Hat{Y} \leftarrow \tilde{Y}_{gen}$

    \ENDIF
   
   \ENDFOR\\
\STATE {\bfseries return} $\hat{L}, \hat{Y}$
\end{algorithmic}
\end{algorithm}
\begin{remark}
    In Algorithm~\ref{alg:example2}, the samples in the minority class are generated using the statistics of the mean and variance of all the components of the majority classes and the Intermediate Classes. For each datapoint in the minority class, the statistics mean is estimated by considering the mean of the neighboring components and the corresponding datapoint of the particular class using the associated weights. Also, for each datapoint, the statistics covariance is calibrated by considering the covariance of the neighboring component and the covariance of this minority class. 
    Without loss of generality, we assume the case for the sample generation in the $c^{th}$ class. 
    The collection $W$ stores $w_i$, where the weights $w_i$ associated with the means and covariance are the inverse of the sample size of the particular component $(\Hat{\mu}_i,\Hat{\Sigma}_i)$ and the weight for the last term (data point of the minority class for the mean calculation and the covariance for the covariance calculation of the minority class) is 1 minus the sum of the rest of the weights. This results in the new mean ($\bar{\mu}_i$) and new covariance ($\bar{\Sigma}_i$) associated with each data point. The number of elements to be generated in the $c^{th}$ class is $n_c$, and the number of elements to be generated corresponding to each element in the $c^{th}$ class is $\hat{n}_c$. Using this new mean and covariance, a set of elements of cardinality $n$ are randomly sampled. $\tilde{L}_{gen}$, $\tilde{Y}_{gen}$ represent the newly generated points and their associated labels corresponding to a particular minority class. The algorithms return the oversampled minority class $\hat{L}$ and its corresponding labels $\hat{Y}$.
\end{remark}
\subsection{Sparsity-Imposing Autoencoders with Class-Preserving Latents}
\textcolor{black}{When dealing with tabular data, we apply our oversampling algorithm directly to the available features to generate synthetic samples for classes with insufficient data. However, challenges arise when we deal with image data as the dataset is complex and has high dimensionality, and the classes need to be better separated. To overcome this, we use an Autoencoder network, which extracts meaningful latent representation from the data and learns the latent space.}
\label{sparcity_autoencoder}
Traditional autoencoders excel in learning latent representations from abundant data, but their performance can suffer when faced with imbalanced datasets. This framework addresses this limitation by incorporating novel constraints and a latent classifier network, leading to improved representation quality and generalization on imbalanced data. 

\textcolor{black}{To overcome the above challenges, we train our autoencoder so that the classes in the latent space are well separated. 
Consider our Image space ($X$) and the corresponding latent space as ($\mathscr{L}$) as a vector space, and let the linear transformation from the Image space to the Latent space be $\mathcal{T}: X \rightarrow \mathscr{L}$. Higgins et al. ~\cite{higgins2018towards} stated the conditions that need to be satisfied to consider a linear transformation as a Group (the definition of which is stated below). A vector representation $(X)$ is disentangled if the decomposition of the vector representation $X=X_1\times X_2\cdots X_K$ into several subspaces is such that each subspace is transformed into a unique symmetric transformation $(\mathcal{T}(X_i))$.}

Let $G$ be a group with the binary operation $*$ and identity element $e$. Let $X$ be any set. We define the action $(\gamma : G\times X \mapsto X)$ of the Group $G$ on $X$ as
\begin{enumerate}
    \item $\gamma(e,x) = x$
    \item $\gamma(a,\gamma(b,x)) = \gamma(a*b,x)$
\end{enumerate}

Suppose the latent space $\mathscr{L} \in \mathbb{R}^d$ is a vector space over $\mathbb{R}$, and a symmetric group decomposes into independent subgroups. Higgins et al. ~\cite{higgins2018towards} proposed that a vector has a disentangled representation if, for a specific decomposition of a symmetric group, the vector decomposition to independent subspaces follows that each subspace is affected by the action of a single group and the remaining subgroups remain unaffected.
 \begin{definition}
    Consider the Group Action, $G\times X \rightarrow X$, and the direct product of the Group $G$ as the decomposition, $G = G_1\times G_2\times \cdots \times G_m$. Let the action of the full group be $*_g$, and the action of each subgroup is $*_{sg}$ respectively. Then, the action is disentangled~\cite{cha2022disentangling} if the decomposition $X = X_1\times X_2\times \cdots \times X_m$, and the actions for each subgroup $*_{sg}: G\times X_{i}\rightarrow X_{i}, i\in {1,2,\cdots,n}$ satisfies:
    \begin{equation}
\begin{aligned}
 (g_1,g_2,\cdots, g_m)*_g(x_1,x_2,\cdots, x_m) 
 &= (g_1*_{sg}x_1,g_2*_{sg}x_2,\cdots, g_m*_{sg}x_m) \\
 &\forall \; g_i \in G_i, \; \forall \; l_i \in L_i
\end{aligned}
\end{equation}
\end{definition}

\begin{definition}
\label{equivariant}
    Let X and Y be a group action of $(G,*)$ on X and Y, respectively. If X and Y are both G-sets for the same group G, then a function $f: X\rightarrow Y$ is said to be equivariant if~\cite{cha2022disentangling}.
    \begin{equation}
        f(g*x) = g*f(x) \hspace{2cm} \forall \hspace{0.1cm} g\in G \hspace{0.2cm}\& \hspace{0.2cm} \forall \hspace{0.1cm} x\in X
    \end{equation}
\end{definition}

\begin{theorem}
\label{thm2}
    Let $f: X\rightarrow \mathscr{L}$ be an equivariant map from the images to the latent feature space, and let $\Tilde{f}: \mathscr{L}\rightarrow Y$ be an equivariant map from the latent feature space to the Reconstructed Image space. Let X, Z, and Y be the G-sets. Then there exists a map $\rho: X\rightarrow Y$, $\rho = \Tilde{f}of$ such that $\rho$ satisfies:\\
    \begin{equation}
    \label{equ12}
        \rho(g*x) = g*\rho(x) \hspace{2cm}  \forall \hspace{0.1cm}g\in G\hspace{0.2cm} \& \hspace{0.2cm}\forall\hspace{0.1cm} x\in X.
    \end{equation}
    
\end{theorem}
\begin{proof}

The proof of Theorem \ref{thm1} is provided in the Supplementary section \ref{app_thm2}. 
\end{proof}

\begin{remark}
From Theorem~\ref{thm2}, we can infer that the goal of disentanglement is finding an equivariant map $\mathscr{\rho}$ which satisfies Equation~\ref{equ12}. Moreover, Autoencoder aims to learn the equivariant map introduced in Definition~\ref{equivariant}.
\end{remark}
To produce a disentangled latent vector representation, as defined above, and class-preserving in the latent space, we impose certain constraints while learning the Autoencoder network. These constraints help us obtain well-separated clusters in the latent space. The constraints we will be imposing are as follows:
\begin{enumerate}[label=(\alph*), itemindent=0pt, topsep=0pt,partopsep=0pt,parsep=0pt,itemsep=0pt]

    \item \textit{Autoencoder Consistency Loss:} 
    Autoencoder Consistency Loss (\(\mathcal{L}_{AC}\)) is essentially the reconstruction loss. However, since we use latents that are explicitly classwise separated in the latent space, the loss emphasizes reconstructing images to be more similar to those within the same class. This further reduces the chance of generating noisy images, thereby justifying the term Autoencoder Consistency Loss. 

    \item \textit{Explicit Between-Class Separation:} We introduce a second constraint term to prevent classes from overlapping in the latent space. This term utilizes the negative dot product between latent vectors from different classes, effectively pushing them apart and promoting discriminative representations. This is achieved by introducing the loss function $\mathcal{L}_{CS}$.
    \item  \textit{MLP-based Latent Classification Guidance:} An MLP classifier operates on the latent representations, providing feedback to the autoencoder. This feedback loop guides the autoencoder to learn feature vectors relevant to classification, further enhancing class separation and representation quality, which we achieve by introducing the loss function $\mathcal{L}_{CG}.$
\end{enumerate} 
Combining these components, the overall loss function aims to achieve three objectives simultaneously:
     (i) Faithful reconstruction of the original data (measured using a reconstruction loss function, i.e., Mean Squared Error Loss).
     (ii) Separation of data points from different classes in the latent space (enforced by the Explicit Between-Class Separation Loss).
     (iii) Providing supervised guidance, encouraging the encoder to learn a more discriminative and structured latent space (enforced by MLP-based Latent Classification Guidance).

Following the above discussion on training the autoencoder, we used a loss function consisting of three components: the Autoencoder Consistency Loss $(\mathcal{L}_{AC})$, which computes the square of the difference between the ground truth element and the corresponding element in the predicted output vector from the decoder defined as:
\begin{equation}
    \mathcal{L}_{AC} = \frac{1}{n}\sum_{i=1}^n(x_i - \hat{x}_i)^2
\end{equation}
; a latent classifier loss $(\mathcal{L}_{CG})$, which preserves the class labels defined as:
\begin{equation}
    \mathcal{L}_{CG} = -\left(\sum_{i=1}^K[y_i\log(p_i) + (1-y_i)\log(1-p_i)]\right)
    \label{CG}
\end{equation}
; and the supervised contrastive loss $(\mathcal{L}_{CS})$, based on Khosla et al. (2020) \cite{khosla2020supervised}, which trains the autoencoder to align feature vectors of the same class in the latent space, defined as:
\begin{equation}
    \mathcal{L}_{CS} = \sum_{i=1}^{n}\log{(\frac{-1}{|Q(i)|})\sum_{j\in Q(i)}\frac{exp(l_i.l_j/t)}{\sum_{z \in Z(i)}exp(l_i.l_z/t)}},
    \label{CS}
\end{equation}
where, n is the total number of samples, $Q(i) = \{j:1<j\leq n\hspace{0.1cm} \&\hspace{0.1cm} y_j = y_i \}$, $Z(i) = \{z:1<z\leq n \hspace{0.1cm}\&\hspace{0.1cm} z\neq i\}$ and $t$ is the Temperature hyperparameter. \\
Then the loss for our autoencoder $(\mathcal{L})$ becomes, 
\begin{equation}
    \mathcal{L} = \mathcal{L}_{AC} + \mathcal{L}_{CG} + \mathcal{L}_{CS}.
    \label{combined_loss}
\end{equation}

Traditional autoencoders excel in learning latent representations from abundant data, but their performance can suffer when faced with imbalanced datasets. This framework for our autoencoder aims to address this limitation by incorporating novel constraints and a dynamic classifier network, leading to improved representation quality and generalization on imbalanced data.

\begin{figure}
\scriptsize
  \centering
  \includegraphics[width=\textwidth]{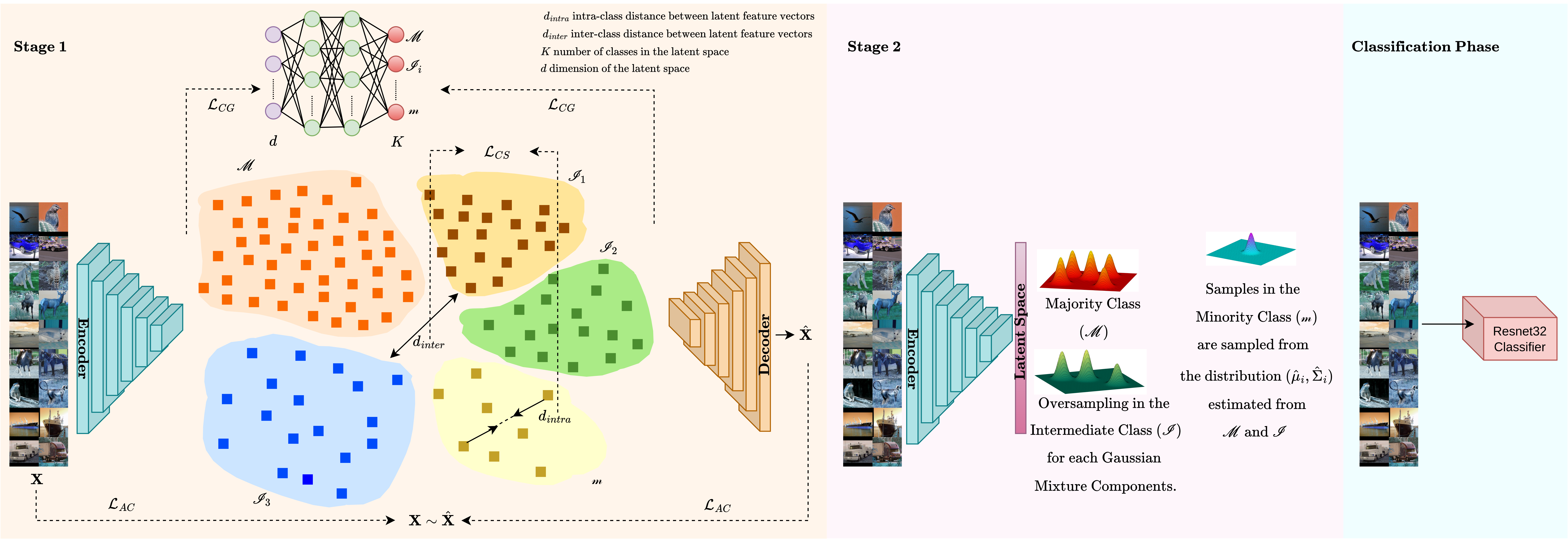}
  \caption{Illustrations of our method, Stage 1 depicts the autoencoder training, and Stage 2 portrays the data calibration in the latent space, followed by a classification phase where we employ ResNet32 architecture for Classification.}
  \label{fig:motivation}
\end{figure}

The raw images are compressed to a lower-dimensional feature vector in the encoder part of the autoencoder. The latent feature vectors are then passed through the classifier network so that our model can make accurate predictions on the latent space, optimizing the parameters by minimizing the cross-entropy loss ($\mathcal{L}_{CG}$). Additionally, our model learns the feature vectors contrastively using the loss function $\mathcal{L}_{CS}$. The decoder utilizes the lower-dimensional feature vector to reconstruct the image. We calculate the mean square error loss $\mathcal{L}_{AC}$ using the reconstructed and the original image. By minimizing the $\mathcal{L}_{CG}$, $\mathcal{L}_{CS}$, and $\mathcal{L}_{AC}$ losses, we effectively train our autoencoder network. This training approach helps the autoencoder learn latent representations that produce disentangled latent representations for different classes. Such latent representations for different classes do not mix with each other, making them useful for oversampling purposes. Overall framework of autoencoder training shown in the Figure \ref{fig:motivation} (Stage 1).
\subsection{Putting it all together}
After training the Autoencoder, the raw dataset is passed through the Encoder network to obtain the latent feature vectors. These latent feature vectors are then categorized into the Majority class, the Intermediate class, and the Minority class. For the Majority class, we compute the Gaussian Mixture components and their corresponding parameters $(\mu_{\mathscr{M}}, \Sigma_{\mathscr{M}})$. For each Intermediate class $\mathscr{I} \in \Hat{\mathscr{I}}$, we derive the Gaussian Mixture components using the distribution parameterized by $(\mu_{\mathscr{I}}, \Sigma_{\mathscr{I}})$, from which the required number of synthetic samples are drawn to balance the Intermediate class.

For each sample in the Minority class $\mathfrak{m} \in \Hat{\mathfrak{m}}$, we estimate the Gaussian parameters based on a combination of the Gaussian components from the Intermediate classes $(\mu_{\Hat{\mathscr{I}}}, \Sigma_{\Hat{\mathscr{I}}})$ and those from the Majority class $(\mu_{\mathscr{M}}, \Sigma_{\mathscr{M}})$. Synthetic samples for the Minority class are then generated using these estimated Gaussian parameters.
We sample synthetic data points in the latent (or feature) space using the estimated parameters. For image data, we pass the oversampled latent (or feature) vectors through the Decoder Network to retrieve the synthetic images. We use the oversampled feature vectors for tabular data to train the classification network. Figure \ref{fig:motivation} shows the overview of the proposed method.
\section{Experiment and Discussions}
\subsection{Dataset Overview}
\begin{table*}[]
\centering
\caption{Summary of the Image Datasets}
\label{tab:summary_image}
\resizebox{0.7\textwidth}{!}{%
\begin{tabular}{|p{2cm}|p{1.6cm}|p{1.5cm}|p{1cm}|p{5cm}|p{5cm}|} 
\toprule
\textbf{Dataset}       & \textbf{Shape}                                  & \textbf{Classes} & \multicolumn{1}{c|}{\textbf{IR}} & \textbf{Training Set (For each Class)}     & \textbf{Testing Set (For each class)}                       \\ \midrule
\textbf{MNIST}         & 28$\times$28$\times$1 & 10               & 100                               & 4000, 2000, 1000, 750, 500, 350, 200, 100, 60, 40  & 1000, 1000, 1000, 1000, 1000, 1000, 1000, 1000, 1000, 1000      \\
\textbf{FMNIST} & 28$\times$28$\times$1 & 10               & 100                               & 4000, 2000, 1000, 750, 500, 350, 200, 100, 60, 40  & 1000, 1000, 1000, 1000, 1000, 1000, 1000, 1000, 1000, 1000      \\
\textbf{SVHN}          & 32$\times$32$\times$3 & 10               & 56.25                            & 4500, 2000, 1000, 800, 600, 500, 400, 250, 150, 80 & 1744, 5099, 4149, 2882, 2523, 2384, 1977, 2019, 1660, 1595 \\
\textbf{CIFAR-10}      & 32$\times$32$\times$3 & 10               & 56.25                            & 4500, 2000, 1000, 800, 600, 500, 400, 250, 150, 80 & 1000, 1000, 1000, 1000, 1000, 1000, 1000, 1000, 1000, 1002      \\
\textbf{STL10}         & 32$\times$32$\times$3  & 10               & 12.5                              & 500, 400, 350, 300, 250, 200, 150, 100, 80, 40                  & 100, 100, 100, 100, 100, 100, 100, 100, 100, 100       \\
\textbf{GTSRB}         & 32$\times$32$\times$3   & 43               & 10.71                              & 2250, 2220, 2160, $\cdots$, 210                  & 750, 450, $\cdots$, 90, 60            \\ 
\textbf{SUN397}         & 32$\times$32$\times$3 \& 64$\times$64$\times$3 & 50               & 11.58                              & 938, 892, 708, 675, 563, $\cdots$, 85, 82, 81, 81                  & 219, 208, 165, 158, 132, $\cdots$, 19, 19, 19, 19        \\ \bottomrule
\end{tabular}%
}
\end{table*}
\begin{table*}[!ht]
\centering
\caption{Summary of the Tabular Datasets}
\label{tab:summary_tabular}
\resizebox{0.7\textwidth}{!}{%
\begin{tabular}{|p{3cm}|p{1.6cm}|p{1.5cm}|p{1cm}|p{5cm}|p{5cm}|} 
\toprule
\textbf{Dataset}       & \textbf{Features}                                  & \textbf{Classes} & \multicolumn{1}{c|}{\textbf{IR}} & \textbf{Training Set (For each Class for all Splits)}     & \textbf{Testing Set (For each Class for all Splits)}                       \\ \midrule
\textbf{Optical\_digits}         & 64 & 2               & 9.13                              & 4052, 444  & 1014, 110      \\
\textbf{Webpage} & 300 & 2               & 34.44                       & 27039, 785  & 6760, 196      \\
\textbf{Isolet}          & 617 & 2               & 11.99                           & 5757, 480 & 1440, 120 \\
\textbf{Contraceptive}      & 8 & 3               & 1.89                          & 502, 408, 266 & 126, 102, 67      \\
\textbf{Dermatology}         & 34 & 6               & 2.32                             & 88, 57, 47, 38                  & 22, 12, 14, 10         \\ 
\textbf{Wine}         & 13 & 3               & 1.43                            & 50, 42, 35                  & 13, 11, 8       \\ \bottomrule
\end{tabular}%
}
\end{table*}

\color{black}

We evaluate on seven benchmark datasets. Following \cite{mullick2019generative}, we use MNIST, FMNIST, CIFAR-10, 
SVHN, and SUN397, maintaining the same number of classes, imbalance ratios, and image sizes: MNIST and FMNIST (10 classes, $28\times28$) with imbalanced training sets of 4000--40 samples per class; CIFAR-10 and SVHN (10 classes, $32\times32$) with 4500--80 samples per class; and SUN397, where 50 classes were selected and images scaled to $32\times32$ and $64\times64$ with an imbalance ratio (IR) of 14.21. In addition, we include GTSRB~\cite{stallkamp2011gtsrb} (43 classes, scaled to $32\times32$, IR = 10.71) and STL-10~\cite{coates2011stl10} (3 channels, $32\times32$, IR = 12.5). Dataset details are provided in Table~\ref{tab:summary_image}.

For the tabular dataset, we considered datasets from the UCI Machine Learning repository: Optical$\_$digits, Webpage, Isolet, Contraceptive, Dermatology, Thyroid, and Wine. Optical$\_$digits, Webpage, and Isolet are datasets with binary classes, whereas Contraceptive, Dermatology, and Wine have multi-class data with 3, 6, and 3 classes, respectively. Detailed descriptions of these datasets are provided in Table~\ref{tab:summary_tabular}.
\textcolor{black}{For textual datasets, we used Spam or Ham, Spam or Not Spam, Emotion, and Hate Speech datasets with detailed descriptions in Supplementary Section~\ref{textual_dataset}.}

\subsection{Training Overview}
\label{train_over}
As described above we used the Encoder-Decoder architecture. The Encoder network consisted of two Convolution Layers, two Batch Normalization layers, and a fully connected layer to obtain the Latent vector with dimensions 300 for the single-channel image and 600 for the multi-channel image. The Decoder network consisted of the deconvolution layers and the batch normalization layers. The Latent Classifier Network consisted of a Multi-layered Perceptron with 4 hidden layers.
We used the Adam optimizer with a learning rate of $1e-4$. The Autoencoder was trained over 200 epochs using the loss function $\mathcal{L}$.

For the sample generation process, we generated the samples in two phases. In the first phase, we generated the samples by traversing through each class, and the number of samples in each class was greater than $\zeta$ were oversampled using Algorithm \ref{alg:example1}.
In the second phase, we similarly traversed through each class, and if the number of samples in each class was less than $\zeta$, we oversampled those classes using Algorithm \ref{alg:example2}.
After including the synthetic data points, we checked the performance of the oversampled data using the ResNet32 network. \textcolor{black}{For image datasets, the optimal hyperparameter values are reported in  Table~\ref{tab:ablation_tuning} (Supplementary).} 

For the Tabular data, the network was trained with Cross-entropy loss with the Adam optimizer, with a learning rate of 0.0001 and a number of epochs of 100. 
For classification of the oversampled feature vectors, we used a Multi-layered Perceptron, trained over 100 epochs with Cross-entropy loss with Adam optimizer, with a learning rate of $1e-3$. \textcolor{black}{For tabular datasets, the optimal hyperparameters are reported in Table~\ref{tab:parameters_tab} (Supplementary).}


\subsection{Evaluation Metrics Used}
\label{eval}
To evaluate the performance of imbalance classification, four well-known metrics were used: Balanced Accuracy (BACC~\footnote{\tiny\url{https://scikit-learn.org/stable/modules/generated/sklearn.metrics.balanced_accuracy_score.html}}): average recall across all classes; Matthews Correlation Coefficient (MCC~\footnote{\tiny\url{https://scikit-learn.org/stable/modules/generated/sklearn.metrics.matthews_corrcoef.html}}): a balanced measure considering true positives, true negatives, false positives, and false negatives; F1-Score~\footnote{\tiny\url{https://scikit-learn.org/stable/modules/generated/sklearn.metrics.f1_score.html}}: combined precision and recall into a single balanced measure; and Geometric Mean (Gmean~\footnote{\tiny\url{https://imbalanced-learn.org/stable/references/generated/imblearn.metrics.geometric_mean_score.html}}): geometric average of class-specific performance metrics. For all these metrics, higher values indicated better performance.

\subsection{Methods Used for Comparison}
For the comparison of our method on the Image datasets, we used the following algorithms. We considered the Baseline method (ResNet32 trained using the cross-entropy loss function without applying any imbalance handling techniques), Resampling methods such as Random Oversampling(ROS),\\
SMOTE~\cite{chawla2002smote}, Generative based methods such as Conditional DCGAN (cDCGAN)~\cite{gauthier2014conditional}, Generative Adversarial Minority Oversampling (GAMO)~\cite{mullick2019generative},  DeepSMOTE~\cite{dablain2021deepsmote}, Deep Generative Mixture Model for Robust Imbalance Classification (DGCMM)~\cite{wang2022deep} and Calibration based method TailCalibX~\cite{vigneswaran2021feature}. In the TailCalibX method, oversampling was performed in the latent feature space, followed by decoding the synthesized samples back to the input space using a decoder. For the SMOTE oversampling algorithm, features are first extracted using the ResNet32 classifier network, after which the SMOTE algorithm is applied to the extracted feature vectors.

For tabular and textual datasets, the peer methods used for comparison included Baseline(BL), SMOTE(SM)~\cite{chawla2002smote}, ADASYN (ADA), BorderlineSMOTE (blSM)~\cite{han2005borderline}, and SVMSMOTE (svmSM)~\cite{SVM2012}, ClusterSMOTE (clSM)~\cite{ClusterSMOTE_2006}, CURE-SMOTE (cuSM)~\cite{ma2017cure}, DBSMOTE (dbSM)~\cite{2012dbsmote}, Kmeans SMOTE (kmSM)~\cite{zhao2021kmeans}, MeanShiftSMOTE (msSM)~\cite{ashour2022adaptive}, MWMOTE   (mwSM) \cite{MWMOTE2014}, SOICJ~\cite{sanchez2013SOICJ}, CE-SMOTE (ceSM)~\cite{CESMOTE2010}. 

\subsection{Results and Discussions}

In this section, we outline the evaluation procedure and report results on image and tabular datasets, with textual dataset details and results provided in Supplementary Section~\ref{textual_data_discussion} and Section~\ref{textual_result_discussion}, respectively.

\subsubsection{Evaluation Procedure}
\label{test}
When dealing with the image datasets, we repeated the entire training process in Subsection~\ref{train_over} for five different seeds, where the training set is imbalanced, and the test set is balanced.
We obtained the mean and the standard deviation on receiving the results corresponding to these seeds. For the tabular and text datasets we fixed the seed value  and used stratified 5-fold cross-validation to train the models, compute the metric values, and report the mean and standard deviation across all splits.
\subsubsection{Discussion on Image Dataset}
\paragraph{\textbf{Discussion on MNIST and FMNIST}}
\label{dismnist}
As shown in Table \ref{tab:mnist_fmnist}, our method consistently outperforms all other methods across all evaluation metrics. On the MNIST dataset, our method achieves a $1.9\%$ improvement in Balanced Accuracy compared to DGCMM, the best-performing state-of-the-art method. Additionally, the MCC, F1-Score, and G-mean improve by $1.9\%$, $1.9\%$, and $1\%$, respectively, over DGCMM. For the FMNIST dataset, our method outperforms GAMO—the strongest baseline—by $0.4\%$ in Balanced Accuracy, while achieving gains of $0.4\%$, $0.5\%$, and $0.2\%$ in MCC, F1-Score, and G-mean, respectively. Compared to MNIST, FMNIST presents more complex patterns, as clothing items exhibit higher intra-class variation and are less distinct from the background. Among the baselines, TailCalibX performs poorly on both datasets due to its inability to capture class distributions in single-channel images. In contrast, our method effectively addresses this limitation by modeling the local distribution within each class.
Since the MNIST dataset is simple, it helps demonstrate model interpretability and explainability techniques.
Compared to MNIST, FMNIST presents more complex patterns, as clothing items exhibit higher intra-class variation and are less distinct from the background. 
\begin{table}
\caption{Results for MNIST and FMNIST}
\label{tab:mnist_fmnist}
\resizebox{\textwidth}{!}{%
\begin{tabular}{cccccclcccc}
\toprule
 & \multicolumn{4}{c}{\textbf{\hspace{4cm}MNIST}} &  & \multicolumn{4}{c}{\textbf{\hspace{4cm}FMNIST}} \\ \cmidrule(l){3-6} \cmidrule(l){8-11} 
\multirow{-2}{*}{\textbf{Methods}}& \multirow{-2}{*}{\textbf{Venue}} & \textbf{BACC} & \textbf{MCC} & \textbf{F1-Score} & \textbf{Gmean} &  & \textbf{BACC} & \textbf{MCC} & \textbf{F1-Score} & \textbf{Gmean} \\ \cmidrule(l){1-2} \cmidrule(l){3-6} \cmidrule(l){8-11}
\textbf{Baseline} & & 0.923$\pm$0.008 & 0.914$\pm$0.009 & 0.921$\pm$0.008 & 0.956$\pm$0.005 &  & 0.769$\pm$0.004 & 0.747$\pm$0.004 & 0.757$\pm$0.006 & 0.865$\pm$0.002 \\
\textbf{ROS} & & 0.89$\pm$0.003 & 0.889$\pm$0.026 & 0.89$\pm$0.008 & 0.895$\pm$0.027 &  & 0.804$\pm$0.003 & 0.783$\pm$0.004 & 0.802$\pm$0.003 & 0.792$\pm$0.003 \\
\textbf{SMOTE} & [JAIR'2002] & 0.932$\pm$0.009 & 0.925$\pm$0.01 & 0.931$\pm$0.009 & 0.962$\pm$0.005 &  & 0.782$\pm$0.007 & 0.761$\pm$0.007 & 0.77$\pm$0.01 & 0.873$\pm$0.004 \\
\textbf{cDCGAN} & [ICLR'2016] & 0.943$\pm$0.004 & 0.937$\pm$0.005 & 0.942$\pm$0.005 & 0.968$\pm$0.003 &  & 0.788$\pm$0.007 & 0.769$\pm$0.009 & 0.779$\pm$0.005 & 0.877$\pm$0.004 \\
\textbf{GAMO} & [ICCV'2019] & 0.873$\pm$0.008 & 0.861$\pm$0.008 & 0.869$\pm$0.009 & 0.928$\pm$0.005 &  & 0.841$\pm$0.011 & 0.825$\pm$0.012 & 0.835$\pm$0.011 & 0.909$\pm$0.006 \\
\textbf{TailCalibX} & [ICLR'2021] & 0.807$\pm$0.016	& 0.788$\pm$0.018	& 0.795$\pm$0.015	& 0.889$\pm$0.009 & 	& 0.69$\pm$0.019	& 0.66$\pm$0.021	& 0.672$\pm$0.018	& 0.816$\pm$0.012 \\
\textbf{DeepSMOTE} & [TNNLS'2022] & 0.943$\pm$0.005 & 0.937$\pm$0.005 & 0.942$\pm$0.005 & 0.968$\pm$0.003 &  & 0.802$\pm$0.011 & 0.783$\pm$0.012 & 0.793$\pm$0.01 & 0.885$\pm$0.006 \\
\textbf{DGCMM} & [TPAMI'2023] & 0.945$\pm$0.002 & 0.939$\pm$0.002 & 0.943$\pm$0.002 & 0.969$\pm$0.001 &  & 0.819$\pm$0.003 & 0.803$\pm$0.003 & 0.812$\pm$0.003 & 0.896$\pm$0.002 \\ \cmidrule(l){1-2} \cmidrule(l){3-6} \cmidrule(l){8-11}
\textbf{RCS} & & \textbf{0.962$\pm$0.002} & \textbf{0.958$\pm$0.002} & \textbf{0.962$\pm$0.002} & \textbf{0.979$\pm$0.001} &  & \textbf{0.845$\pm$0.004} & \textbf{0.829$\pm$0.004} & \textbf{0.84$\pm$0.006} & \textbf{0.911$\pm$0.002} \\ \bottomrule
\end{tabular}
}
\end{table}
\paragraph{\textbf{Discussion on CIFAR10, SVHN}} 
\label{discifar10}
Table~\ref{tab:cifar10_svhn_resnet} presents the performance metrics of various state-of-the-art methods and our method on these datasets.
Despite having three channels, CIFAR-10 remains challenging due to the small size of the images, which makes it harder for models to capture fine details.
Due to the overlap between various classes, any models trained on this dataset fail to distinguish more minor differences.
The SVHN dataset requires careful preprocessing, which includes normalization, cropping, and the challenges with overlapping digits. Our Autoencoder network was trained to overcome this preprocessing overhead so that the classes are well separated in the latent space.
Conditional DCGAN performed worse than our method because of the nature of images in the SVHN dataset.
For CIFAR10, the balanced accuracy of our algorithm improved by 4\% compared to DGCMM, which is the best among the competitor methods; also the MCC and Gmean increased by $3.8 \%, 3.1 \%$ respectively to the DGCMM method and the F1-Score metric for our method has increased by $5.6 \%$. For the SVHN dataset, we improved by 1.2\% over DGCMM, which is the best-performing state-of-the-art method. The MCC, F1-Score, and Gmean metrics for RCS on SVHN increased by $ 1.1 \%, 1.1 \%,$ and $0.7 \%$ respectively than the DGCMM method.

\begin{table}
\caption{Results for CIFAR10 and SVHN}
\label{tab:cifar10_svhn_resnet}
\resizebox{\textwidth}{!}{%
\begin{tabular}{cccccclcccc}
\toprule
& \multicolumn{4}{c}{\textbf{\hspace{4cm}CIFAR10}} &  & \multicolumn{4}{c}{\textbf{\hspace{4cm}SVHN}} \\ \cmidrule(l){3-6} \cmidrule(l){8-11} 
\multirow{-2}{*}{\textbf{Methods}}& \multirow{-2}{*}{\textbf{Venue}} & \textbf{BACC} & \textbf{MCC} & \textbf{F1-Score} & \textbf{Gmean} &  & \textbf{BACC} & \textbf{MCC} & \textbf{F1-Score} & \textbf{Gmean} \\ \cmidrule(l){1-2} \cmidrule(l){3-6} \cmidrule(l){8-11}
\textbf{Baseline} & & 0.377$\pm$0.006 & 0.316$\pm$0.007 & 0.334$\pm$0.015 & 0.592$\pm$0.005 &  & 0.624$\pm$0.016 & 0.588$\pm$0.018 & 0.608$\pm$0.015 & 0.773$\pm$0.011 \\
\textbf{ROS} & & 0.235$\pm$0.009 & 0.155$\pm$0.009 & 0.203$\pm$0.02 & 0.463$\pm$0.009 &  & 0.543$\pm$0.027 & 0.504$\pm$0.029 & 0.507$\pm$0.027 & 0.718$\pm$0.019 \\
\textbf{SMOTE} & [JAIR'2002] & 0.371$\pm$0.013 & 0.309$\pm$0.013 & 0.332$\pm$0.028 & 0.587$\pm$0.011 &  & 0.617$\pm$0.017 & 0.58$\pm$0.018 & 0.601$\pm$0.019 & 0.769$\pm$0.011 \\
\textbf{cDCGAN} & [ICLR'2016] & 0.395$\pm$0.021 & 0.337$\pm$0.024 & 0.358$\pm$0.018 & 0.607$\pm$0.017 &  & 0.701$\pm$0.039 & 0.672$\pm$0.043 & 0.686$\pm$0.045 & 0.823$\pm$0.025 \\
\textbf{GAMO} & [ICCV'2019] & 0.43$\pm$0.012 & 0.375$\pm$0.014 & 0.402$\pm$0.012 & 0.635$\pm$0.009 &  & 0.719$\pm$0.016 & 0.691$\pm$0.017 & 0.714$\pm$0.016 & 0.835$\pm$0.01 \\
\textbf{TailCalibX} & [ICLR'2021] & 0.355$\pm$0.015	& 0.294$\pm$0.018	& 0.302$\pm$0.022	& 0.574$\pm$0.013 & 	& 0.528$\pm$0.021	& 0.481$\pm$0.024	& 0.501$\pm$0.021	& 0.707$\pm$0.015 \\
\textbf{DeepSMOTE} & [TNNLS'2022] & 0.43$\pm$0.007 & 0.373$\pm$0.008 & 0.411$\pm$0.012 & 0.635$\pm$0.006 &  & 0.691$\pm$0.033 & 0.662$\pm$0.035 & 0.678$\pm$0.039 & 0.817$\pm$0.021 \\
\textbf{DGCMM} & [TPAMI'2023] & 0.432$\pm$0.004 & 0.38$\pm$0.004 & 0.397$\pm$0.007 & 0.636$\pm$0.003 &  & 0.764$\pm$0.009 & 0.742$\pm$0.01 & 0.762$\pm$0.009 & 0.863$\pm$0.006 \\ \cmidrule(l){1-2} \cmidrule(l){3-6} \cmidrule(l){8-11}
\textbf{RCS} & & \textbf{0.472$\pm$0.011} & \textbf{0.418$\pm$0.012} & \textbf{0.467$\pm$0.011} & \textbf{0.667$\pm$0.008} &  & \textbf{0.776$\pm$0.011} & \textbf{0.753$\pm$0.012} & \textbf{0.773$\pm$0.012} & \textbf{0.870$\pm$0.007} \\ \bottomrule
\end{tabular}
 }
\end{table}


\paragraph{\textbf{Discusson on GTSRB and STL10}} 
\label{disgtsrb}
\textcolor{black}{The experiments on the GTSRB dataset evaluated model performance under limited training/validation samples and a relatively low class-imbalance ratio. The dataset contains traffic sign images captured under varied lighting and weather conditions. As shown in Table~\ref{tab:gtsrb_stl10}, our method outperforms all state-of-the-art baselines: it improves balanced accuracy by 1.8\% over DGCMM (the strongest baseline) and by 7.1\% over TailCalibX. In addition, MCC, F1-Score, and Gmean increase by 0.9\%, 0.8\%, and 1\% respectively compared to DGCMM.  
The STL10 dataset presents substantial intra-class variability and inter-class similarity, making classification particularly challenging. Models trained here must effectively capture subtle differences between classes. As shown in Table~\ref{tab:gtsrb_stl10}, our method again achieves the best performance across all metrics. Balanced accuracy improves by 0.1\% compared to TailCalibX, while MCC, F1-Score, and Gmean increase by 0.2\%, 0.7\%, and 0.1\% respectively. Against generative models, the improvements are substantial: relative to DGCMM, we observe gains of 3\%, 3\%, 4.9\%, and 2.4\% in balanced accuracy, MCC, F1-Score, and Gmean; relative to GAMO, the improvements are 2.9\%, 3.1\%, 3.2\%, and 2.3\%.  
These results highlight that our distribution calibration method consistently surpasses both calibration and generation based approaches. Even in scenarios with lower imbalance, it produces new samples that faithfully reflect the minority-class distribution without distorting the original data.}

\begin{table}
\caption{Results for GTSRB and STL10}
\label{tab:gtsrb_stl10}
\resizebox{\textwidth}{!}{%
\begin{tabular}{cccccccccccc}
\toprule
& \multicolumn{4}{c}{\textbf{\hspace{4cm}GTSRB}} &  & \multicolumn{4}{c}{\textbf{\hspace{4cm}STL10}} \\ \cmidrule(l){3-6} \cmidrule(l){8-11} 
\multirow{-2}{*}{\textbf{Methods}}& \multirow{-2}{*}{\textbf{Venue}} & \textbf{BACC} & \textbf{MCC} & \textbf{F1-Score} & \textbf{Gmean} &  & \textbf{BACC} & \textbf{MCC} & \textbf{F1-Score} & \textbf{Gmean} \\ \cmidrule(l){1-2} \cmidrule(l){3-6} \cmidrule(l){8-11}
\textbf{Baseline} & & 0.861$\pm$0.012 & 0.907$\pm$0.010 & 0.866$\pm$0.011 & 0.927$\pm$0.007 &  & 0.347$\pm$0.014 & 0.280$\pm$0.016 & 0.315$\pm$0.016 & 0.567$\pm$0.012 \\
\textbf{ROS} & & 0.905$\pm$0.015 & 0.939$\pm$0.007 & 0.913$\pm$0.013 & 0.95$\pm$0.008 &  & 0.342$\pm$0.014 & 0.273$\pm$0.016 & 0.318$\pm$0.02 & 0.563$\pm$0.012 \\
\textbf{SMOTE} & [JAIR'2002] & 0.883$\pm$0.018 & 0.916$\pm$0.017 & 0.884$\pm$0.016 & 0.939$\pm$0.010 &  & 0.349$\pm$0.000 & 0.281$\pm$0.000 & 0.323$\pm$0.000 & 0.569$\pm$0.000 \\
\textbf{cDCGAN} & [ICLR'2016] & 0.878$\pm$0.013 & 0.924$\pm$0.008 & 0.885$\pm$0.012 & 0.936$\pm$0.007 &  & 0.349$\pm$0.006 & 0.282$\pm$0.008 & 0.324$\pm$0.006 & 0.569$\pm$0.005 \\
\textbf{GAMO} & [ICCV'2019] & 0.832$\pm$0.019 & 0.878$\pm$0.011 & 0.829$\pm$0.015 & 0.911$\pm$0.010 &  & 0.391$\pm$0.011 & 0.329$\pm$0.009 & 0.369$\pm$0.013 & 0.604$\pm$0.009 \\
\textbf{TailCalibX} & [ICLR'2021] & 0.884$\pm$0.028 &	0.919$\pm$0.018	 & 0.883$\pm$0.027  & 0.939$\pm$0.015 & &	0.419$\pm$0.012 &	0.358$\pm$0.014	& 0.394$\pm$0.012	& 0.626$\pm$0.009 \\
\textbf{DeepSMOTE} & [TNNLS'2022] & 0.830$\pm$0.017 & 0.879$\pm$0.011 & 0.830$\pm$0.015 & 0.910$\pm$0.010 &  & 0.369$\pm$0.004 & 0.304$\pm$0.005 & 0.347$\pm$0.006 & 0.586$\pm$0.003 \\
\textbf{DGCMM} & [TPAMI'2023] & 0.937$\pm$0.005 &	0.956$\pm$0.002 &	0.94$\pm$0.004 &	0.967$\pm$0.002 &  & 0.390$\pm$0.001 & 0.33$\pm$0.002 & 0.352$\pm$0.001 & 0.603$\pm$0.001 \\ \cmidrule(l){1-2} \cmidrule(l){3-6} \cmidrule(l){8-11}
\textbf{RCS} & & \textbf{0.955$\pm$0.006} & \textbf{0.965$\pm$0.004} & \textbf{0.948$\pm$0.007} & \textbf{0.977$\pm$0.003} &  & \textbf{0.420$\pm$0.009} & \textbf{0.360$\pm$0.010} & \textbf{0.401$\pm$0.008} & \textbf{0.627$\pm$0.007} \\ \bottomrule
\end{tabular}
}
\end{table}

\paragraph{\textbf{Discussion on SUN397}}
\label{dissun}
SUN397 involves scene recognition that understands the image's context, spatial relationships, and background elements. Table \ref{tab:my-table_sun397} shows that our method outperforms all other competitive methods. When comparing SUN397 with an image size of $32\times32$, we observed that the balanced accuracy metric for our method improved by 3.2\% over the DeepSMOTE method, the best-performing state-of-the-art method. Additionally, the MCC and F-Score increased by 1.7\% and 2.8\%, respectively, compared to the DGCMM method. The F1-Score metric for our method has increased by 2.9\% over the DeepSMOTE method. 
On the other hand, when comparing the results for the SUN397 dataset with an image size of $64\times64$, we observe that the balanced accuracy value for our method has improved by 3\% over the DGCMM method, which is the best among the state-of-the-art methods. Increasing the image size from $32\times32$ to $64\times64$ captures finer details of the images in the dataset, resulting in an 8.8\% increase in balanced accuracy. The MCC, F1-Score, and Gmean metrics for RCS on SUN397 escalate by 1.8\%, 2.8\%, and 2.3\%, respectively, compared to the DGCMM method.

\begin{table}
\centering
\caption{Results for SUN397 with images of sizes 32$\times$32 and 64$\times$64 respectively.}
\label{tab:my-table_sun397}
\resizebox{\textwidth}{!}{%
\begin{tabular}{ccccccccccc}
\toprule
& \multicolumn{4}{c}{\textbf{ \hspace{4cm}       SUN397 with Image Size 32$\times$32}} &  & \multicolumn{4}{c}{\textbf{     \hspace{2cm}   SUN397 with Image Size 64$\times$64}} \\ \cmidrule(l){3-6} \cmidrule(l){8-11} 
\multirow{-2}{*}{\textbf{Methods}}& \multirow{-2}{*}{\textbf{Venue}} & \textbf{BACC} & \textbf{MCC} & \textbf{F1-Score} & \textbf{Gmean} &  & \textbf{BACC} & \textbf{MCC} & \textbf{F1-Score} & \textbf{Gmean} \\ \cmidrule(l){1-2} \cmidrule(l){3-6} \cmidrule(l){8-11}
\textbf{Baseline}        &            & 0.237$\pm$0.004          & 0.299$\pm$0.004        & 0.227$\pm$0.005          & 0.484$\pm$0.004         &  & 0.342$\pm$0.018                              & 0.397$\pm$0.017                             & 0.338$\pm$0.019                                  & 0.581$\pm$0.015                               \\
\textbf{ROS}              &          & 0.232$\pm$0.011          & 0.212$\pm$0.013         & 0.200$\pm$0.013            & 0.478$\pm$0.011         &  & 0.342$\pm$0.006                              & 0.331$\pm$0.008                             & 0.313$\pm$0.004                                  & 0.581$\pm$0.005                               \\
\textbf{SMOTE}           &     [JAIR'2002]      & 0.242$\pm$0.014          & 0.299$\pm$0.012         & 0.232$\pm$0.018          & 0.489$\pm$0.014         &  & 0.347$\pm$0.008                              & 0.401$\pm$0.009                             & 0.345$\pm$0.01                                   & 0.586$\pm$0.006                               \\
\textbf{cDCGAN}         &      [ICLR'2016]      & 0.236$\pm$0.005          & 0.268$\pm$0.007         & 0.229$\pm$0.008          & 0.482$\pm$0.005         &  & 0.294$\pm$0.012                              & 0.334$\pm$0.011                             & 0.284$\pm$0.014                                  & 0.538$\pm$0.011                               \\
\textbf{GAMO}             &     [ICCV'2019]     & 0.261$\pm$0.007          & 0.296$\pm$0.008         & 0.263$\pm$0.008          & 0.507$\pm$0.007         &  & 0.283$\pm$0.006                              & 0.333$\pm$0.005                             & 0.284$\pm$0.005                                  & 0.528$\pm$0.006                               \\
\textbf{TailCalibX} & [ICLR'2021] & 0.271$\pm$0.011	& 0.326$\pm$0.016	& 0.255$\pm$0.014	& 0.517$\pm$0.01 & &	0.349$\pm$0.007	& 0.414$\pm$0.009	& 0.337$\pm$0.004	& 0.587$\pm$0.006 \\

\textbf{DeepSMOTE}        &    [TNNLS'2022]      & 0.286$\pm$0.008          & 0.294$\pm$0.007         & 0.27$\pm$0.003           & 0.531$\pm$0.007         &  & 0.373$\pm$0.013                              & 0.394$\pm$0.023                             & 0.362$\pm$0.016                                  & 0.607$\pm$0.01                                \\
\textbf{DGCMM}         &       [TPAMI'2023]      & 0.276$\pm$0.002          & 0.313$\pm$0.006         & 0.275$\pm$0.002          & 0.521$\pm$0.002         &  & 0.376$\pm$0.004                              & 0.421$\pm$0.003                             & 0.369$\pm$0.004                                  & 0.610$\pm$0.003                                \\ \cmidrule(l){1-2} \cmidrule(l){3-6} \cmidrule(l){8-11}
\textbf{RCS}     &          & \textbf{0.318$\pm$0.009} & \textbf{0.330$\pm$0.014} & \textbf{0.303$\pm$0.013} & \textbf{0.560$\pm$0.008} &  & \textbf{0.406$\pm$0.007}                     & \textbf{0.439$\pm$0.009}  & \textbf{0.397$\pm$0.004}  & \textbf{0.633$\pm$0.006}                      \\ \bottomrule
\end{tabular}%
}
\end{table}

\begin{table}[!ht]
\caption{Results of Our Data Calibration Method on Tabular Datasets}
\label{tab:my-tabular2}
\resizebox{\textwidth}{!}{%
\begin{tabular}{cccccccccccccccc}
\toprule
\textbf{Datasets} & \textbf{Metrics} & \textbf{BL} & \textbf{SM} & \textbf{ADA} & \textbf{blSM} & \textbf{clSM} & \textbf{cuSM} & \textbf{DBSM} & \textbf{kmSM} & \textbf{msSM} & \textbf{mwSM} & \textbf{SOICJ} & \textbf{svmSM} & \textbf{csSM} & \textbf{RCS} \\ \midrule
\multirow{8}{*}{\rotatebox[origin=c]{90}{\textbf{Contraceptive}}} & \multirow{2}{*}{\textbf{BACC}} & 0.511 & 0.514 & 0.501 & 0.52 & 0.519 & 0.486 & 0.523 & 0.489 & 0.493 & 0.533 & 0.471 & 0.507 & 0.517 & \textbf{0.555} \\
 &  & $\pm$0.041 & $\pm$0.036 & $\pm$0.038 & $\pm$0.058 & $\pm$0.059 & $\pm$0.053 & $\pm$0.038 & $\pm$0.014 & $\pm$0.05 & $\pm$0.038 & $\pm$0.051 & $\pm$0.049 & $\pm$0.05 & \textbf{$\pm$0.032} \\
 & \multirow{2}{*}{\textbf{MCC}} & 0.259 & 0.258 & 0.256 & 0.268 & 0.268 & 0.217 & 0.277 & 0.229 & 0.237 & 0.29 & 0.217 & 0.25 & 0.268 & \textbf{0.329} \\
 &  & $\pm$0.058 & $\pm$0.046 & $\pm$0.055 & $\pm$0.085 & $\pm$0.084 & $\pm$0.081 & $\pm$0.053 & $\pm$0.024 & $\pm$0.068 & $\pm$0.055 & $\pm$0.065 & $\pm$0.075 & $\pm$0.076 & \textbf{$\pm$0.039} \\
 & \multirow{2}{*}{\textbf{F1-Score}} & 0.5 & 0.497 & 0.514 & 0.498 & 0.496 & 0.471 & 0.521 & 0.489 & 0.427 & 0.519 & 0.463 & 0.494 & 0.498 & \textbf{0.546} \\
 &  & $\pm$0.039 & $\pm$0.027 & $\pm$0.027 & $\pm$0.056 & $\pm$0.059 & $\pm$0.048 & $\pm$0.031 & $\pm$0.02 & $\pm$0.084 & $\pm$0.038 & $\pm$0.068 & $\pm$0.058 & $\pm$0.042 & \textbf{$\pm$0.026} \\
 & \multirow{2}{*}{\textbf{Gmean}} & 0.62 & 0.622 & 0.613 & 0.626 & 0.626 & 0.599 & \textbf{0.63} & 0.603 & 0.604 & 0.638 & 0.589 & 0.617 & 0.624 & 0.548 \\
 &  & $\pm$0.033 & $\pm$0.028 & $\pm$0.031 & $\pm$0.047 & $\pm$0.048 & $\pm$0.044 & \textbf{$\pm$0.03} & $\pm$0.012 & $\pm$0.041 & $\pm$0.031 & $\pm$0.041 & $\pm$0.04 & $\pm$0.04 & $\pm$0.034 \\ \midrule
\multirow{8}{*}{\rotatebox[origin=c]{90}{\textbf{Isolet}}} & \multirow{2}{*}
{\textbf{BACC}} & 0.5 & 0.98 & 0.981 & 0.978 & 0.98 & 0.975 & 0.978 & 0.975 & 0.98 & 0.979 & 0.597 & 0.982 & 0.981 & \textbf{0.997} \\
 &  & $\pm$0 & $\pm$0.008 & $\pm$0.007 & $\pm$0.008 & $\pm$0.009 & $\pm$0.006 & $\pm$0.006 & $\pm$0.006 & $\pm$0.008 & $\pm$0.008 & $\pm$0.216 & $\pm$0.007 & $\pm$0.008 & \textbf{$\pm$0.005} \\
 & \multirow{2}{*}{\textbf{MCC}} & 0 & 0.89 & 0.872 & 0.845 & 0.864 & 0.858 & 0.868 & 0.867 & 0.878 & 0.885 & 0.178 & 0.865 & 0.874 & \textbf{0.976} \\
 &  & $\pm$0 & $\pm$0.04 & $\pm$0.051 & $\pm$0.038 & $\pm$0.064 & $\pm$0.035 & $\pm$0.059 & $\pm$0.038 & $\pm$0.049 & $\pm$0.035 & $\pm$0.399 & $\pm$0.057 & $\pm$0.063 & \textbf{$\pm$0.041} \\
 & \multirow{2}{*}{\textbf{F1-Score}} & 0.886 & 0.983 & 0.979 & 0.975 & 0.978 & 0.977 & 0.979 & 0.979 & 0.981 & 0.982 & 0.906 & 0.978 & 0.98 & \textbf{0.996} \\
 &  & $\pm$0 & $\pm$0.007 & $\pm$0.01 & $\pm$0.007 & $\pm$0.012 & $\pm$0.006 & $\pm$0.012 & $\pm$0.007 & $\pm$0.009 & $\pm$0.006 & $\pm$0.043 & $\pm$0.011 & $\pm$0.012 & \textbf{$\pm$0.006} \\
 & \multirow{2}{*}{\textbf{Gmean}} & 0.5 & 0.98 & 0.981 & 0.978 & 0.98 & 0.975 & 0.978 & 0.975 & 0.98 & 0.979 & 0.597 & 0.982 & 0.981 & \textbf{0.997} \\
 &  & $\pm$0 & $\pm$0.008 & $\pm$0.007 & $\pm$0.008 & $\pm$0.009 & $\pm$0.006 & $\pm$0.006 & $\pm$0.006 & $\pm$0.008 & $\pm$0.008 & $\pm$0.216 & $\pm$0.007 & $\pm$0.008 & \textbf{$\pm$0.005} \\ \midrule
\multirow{8}{*}{\rotatebox[origin=c]{90}{\textbf{Webpage}}} & \multirow{2}{*}{\textbf{BACC}} & 0.864 & 0.93 & 0.922 & 0.913 & 0.922 & 0.915 & 0.925 & 0.871 & 0.929 & 0.925 & 0.878 & 0.923 & 0.925 & \textbf{0.938} \\
 &  & $\pm$0.013 & $\pm$0.016 & $\pm$0.016 & $\pm$0.018 & $\pm$0.01 & $\pm$0.005 & $\pm$0.014 & $\pm$0.013 & $\pm$0.013 & $\pm$0.015 & $\pm$0.016 & $\pm$0.012 & $\pm$0.022 & \textbf{$\pm$0.022} \\
 & \multirow{2}{*}{\textbf{MCC}} & 0.768 & 0.647 & 0.658 & 0.676 & 0.709 & 0.76 & 0.643 & 0.776 & 0.613 & 0.68 & 0.758 & 0.644 & 0.667 & \textbf{0.879} \\
 &  & $\pm$0.021 & $\pm$0.022 & $\pm$0.049 & $\pm$0.042 & $\pm$0.02 & $\pm$0.019 & $\pm$0.044 & $\pm$0.014 & $\pm$0.01 & $\pm$0.045 & $\pm$0.019 & $\pm$0.033 & $\pm$0.026 & \textbf{$\pm$0.049} \\
 & \multirow{2}{*}{\textbf{F1-Score}} & 0.988 & 0.975 & 0.976 & 0.979 & 0.982 & 0.986 & 0.974 & 0.988 & 0.971 & 0.978 & 0.987 & 0.975 & 0.977 & \textbf{0.993} \\
 &  & $\pm$0.001 & $\pm$0.003 & $\pm$0.005 & $\pm$0.004 & $\pm$0.002 & $\pm$0.001 & $\pm$0.005 & $\pm$0.001 & $\pm$0.001 & $\pm$0.005 & $\pm$0.001 & $\pm$0.004 & $\pm$0.004 & \textbf{$\pm$0.003} \\
 & \multirow{2}{*}{\textbf{Gmean}} & 0.864 & 0.93 & 0.922 & 0.913 & 0.922 & 0.915 & 0.925 & 0.871 & 0.929 & 0.925 & 0.878 & 0.923 & 0.925 & \textbf{0.936} \\
 &  & $\pm$0.013 & $\pm$0.016 & $\pm$0.016 & $\pm$0.018 & $\pm$0.01 & $\pm$0.005 & $\pm$0.014 & $\pm$0.013 & $\pm$0.013 & $\pm$0.015 & $\pm$0.016 & $\pm$0.012 & $\pm$0.022 & \textbf{$\pm$0.024} \\ \midrule
\multirow{8}{*}{\rotatebox[origin=c]{90}{\textbf{Optical\_digits}}} & \multirow{2}{*}{\textbf{BACC}} & 0.5 & 0.971 & 0.975 & 0.969 & 0.977 & 0.967 & 0.977 & 0.928 & 0.974 & 0.977 & 0.86 & 0.983 & 0.974 & \textbf{0.991} \\
 &  & $\pm$0 & $\pm$0.017 & $\pm$0.006 & $\pm$0.009 & $\pm$0.01 & $\pm$0.016 & $\pm$0.008 & $\pm$0.028 & $\pm$0.004 & $\pm$0.009 & $\pm$0.202 & $\pm$0.006 & $\pm$0.008 & \textbf{$\pm$0.004} \\
 & \multirow{2}{*}{\textbf{MCC}} & 0 & 0.895 & 0.922 & 0.931 & 0.935 & 0.934 & 0.925 & 0.852 & 0.929 & 0.937 & 0.727 & 0.914 & 0.922 & \textbf{0.942} \\
 &  & $\pm$0 & $\pm$0.047 & $\pm$0.023 & $\pm$0.015 & $\pm$0.026 & $\pm$0.023 & $\pm$0.028 & $\pm$0.071 & $\pm$0.026 & $\pm$0.025 & $\pm$0.407 & $\pm$0.022 & $\pm$0.034 & \textbf{$\pm$0.025} \\
 & \multirow{2}{*}{\textbf{F1-Score}} & 0.855 & 0.98 & 0.986 & 0.988 & 0.988 & 0.988 & 0.986 & 0.973 & 0.987 & 0.989 & 0.958 & 0.984 & 0.986 & \textbf{0.989} \\
 &  & $\pm$0.001 & $\pm$0.009 & $\pm$0.004 & $\pm$0.003 & $\pm$0.005 & $\pm$0.004 & $\pm$0.005 & $\pm$0.013 & $\pm$0.005 & $\pm$0.005 & $\pm$0.058 & $\pm$0.004 & $\pm$0.006 & \textbf{$\pm$0.005} \\
 & \multirow{2}{*}{\textbf{Gmean}} & 0.5 & 0.971 & 0.975 & 0.969 & 0.977 & 0.967 & 0.977 & 0.928 & 0.974 & 0.977 & 0.86 & 0.983 & 0.974 & \textbf{0.991} \\
 &  & $\pm$0 & $\pm$0.017 & $\pm$0.006 & $\pm$0.009 & $\pm$0.01 & $\pm$0.016 & $\pm$0.008 & $\pm$0.028 & $\pm$0.004 & $\pm$0.009 & $\pm$0.202 & $\pm$0.006 & $\pm$0.008 & \textbf{$\pm$0.004} \\ \midrule
\multirow{8}{*}{\rotatebox[origin=c]{90}{\textbf{Dermatology}}} & \multirow{2}{*}{\textbf{BACC}} & 0.782 & 0.97 & 0.854 & 0.899 & 0.968 & 0.798 & 0.977 & 0.97 & 0.971 & 0.975 & 0.981 & 0.979 & 0.942 & \textbf{0.981} \\
 &  & $\pm$0.083 & $\pm$0.026 & $\pm$0.029 & $\pm$0.071 & $\pm$0.03 & $\pm$0.028 & $\pm$0.015 & $\pm$0.017 & $\pm$0.027 & $\pm$0.024 & $\pm$0.022 & $\pm$0.014 & $\pm$0.067 & \textbf{$\pm$0.025} \\
 & \multirow{2}{*}{\textbf{MCC}} & 0.837 & 0.966 & 0.839 & 0.91 & 0.965 & 0.893 & 0.976 & 0.969 & 0.966 & 0.976 & 0.979 & 0.975 & 0.94 & \textbf{0.979} \\
 &  & $\pm$0.045 & $\pm$0.031 & $\pm$0.033 & $\pm$0.035 & $\pm$0.033 & $\pm$0.032 & $\pm$0.015 & $\pm$0.014 & $\pm$0.034 & $\pm$0.023 & $\pm$0.023 & $\pm$0.02 & $\pm$0.064 & \textbf{$\pm$0.029} \\
 & \multirow{2}{*}{\textbf{F1-Score}} & 0.818 & 0.971 & 0.815 & 0.921 & 0.972 & 0.888 & 0.98 & 0.974 & 0.972 & 0.98 & 0.983 & 0.98 & 0.939 & \textbf{0.983} \\
 &  & $\pm$0.062 & $\pm$0.027 & $\pm$0.055 & $\pm$0.037 & $\pm$0.026 & $\pm$0.025 & $\pm$0.013 & $\pm$0.013 & $\pm$0.028 & $\pm$0.02 & $\pm$0.019 & $\pm$0.016 & $\pm$0.079 & \textbf{$\pm$0.023} \\
 & \multirow{2}{*}{\textbf{Gmean}} & 0.871 & 0.982 & 0.911 & 0.941 & 0.981 & 0.885 & 0.987 & 0.982 & 0.983 & 0.986 & 0.989 & 0.988 & 0.965 & \textbf{0.98} \\
 &  & $\pm$0.05 & $\pm$0.016 & $\pm$0.018 & $\pm$0.04 & $\pm$0.018 & $\pm$0.017 & $\pm$0.009 & $\pm$0.01 & $\pm$0.016 & $\pm$0.014 & $\pm$0.013 & $\pm$0.009 & $\pm$0.04 & \textbf{$\pm$0.027} \\ \midrule
\multirow{8}{*}{\rotatebox[origin=c]{90}{\textbf{Wine}}} & \multirow{2}{*}{\textbf{BACC}} & 0.759 & 0.795 & 0.787 & 0.827 & 0.731 & 0.884 & 0.874 & 0.781 & 0.655 & 0.722 & 0.782 & 0.744 & 0.677 & \textbf{0.95} \\
 &  & $\pm$0.142 & $\pm$0.128 & $\pm$0.071 & $\pm$0.187 & $\pm$0.118 & $\pm$0.087 & $\pm$0.122 & $\pm$0.129 & $\pm$0.077 & $\pm$0.07 & $\pm$0.123 & $\pm$0.217 & $\pm$0.188 & \textbf{$\pm$0.039} \\
 & \multirow{2}{*}{\textbf{MCC}} & 0.695 & 0.711 & 0.683 & 0.763 & 0.623 & 0.822 & 0.807 & 0.699 & 0.548 & 0.609 & 0.693 & 0.651 & 0.562 & \textbf{0.92} \\
 &  & $\pm$0.163 & $\pm$0.165 & $\pm$0.103 & $\pm$0.25 & $\pm$0.172 & $\pm$0.123 & $\pm$0.172 & $\pm$0.185 & $\pm$0.096 & $\pm$0.066 & $\pm$0.156 & $\pm$0.283 & $\pm$0.234 & \textbf{$\pm$0.063} \\
 & \multirow{2}{*}{\textbf{F1-Score}} & 0.738 & 0.716 & 0.708 & 0.794 & 0.61 & 0.851 & 0.815 & 0.683 & 0.553 & 0.621 & 0.692 & 0.669 & 0.541 & \textbf{0.943} \\
 &  & $\pm$0.166 & $\pm$0.195 & $\pm$0.151 & $\pm$0.226 & $\pm$0.205 & $\pm$0.113 & $\pm$0.219 & $\pm$0.227 & $\pm$0.059 & $\pm$0.101 & $\pm$0.205 & $\pm$0.289 & $\pm$0.252 & \textbf{$\pm$0.048} \\
 & \multirow{2}{*}{\textbf{Gmean}} & 0.815 & 0.837 & 0.83 & 0.864 & 0.786 & 0.908 & 0.899 & 0.826 & 0.731 & 0.779 & 0.825 & 0.799 & 0.743 & \textbf{0.946} \\
 &  & $\pm$0.109 & $\pm$0.103 & $\pm$0.058 & $\pm$0.147 & $\pm$0.095 & $\pm$0.069 & $\pm$0.099 & $\pm$0.106 & $\pm$0.053 & $\pm$0.054 & $\pm$0.101 & $\pm$0.171 & $\pm$0.148 & \textbf{$\pm$0.044} \\ \bottomrule
\end{tabular}
}
\end{table}
\subsubsection{Discussion on Tabular Datasets}
\label{distab}
We performed experiments on 7 tabular datasets: Optical\_digits, Webpage, Isolet, Contraceptive, Dermatology, Thyroid, and Wine. The feature vectors were directly used for our calibration technique. We split the feature vectors and their corresponding labels into 5 parts in each of these datasets using the 5-fold Stratified cross-validation technique. Then, in each of these 5 folds, we used 4 parts for training and the remaining 1 for validation. For the Optical$\_$digits dataset, our model handled variations in handwriting styles, and it beat the balanced accuracy metric of the highest competitive method, i.e., svmSM (SVM SMOTE) by 0.8 \%. 
To check the performance of our algorithm, we consider the method with the second-highest accuracy and compare it with our method. 
On the Contraceptive dataset, the metrics balanced accuracy, MCC, and F1-Score increased by $2.2 \%, 3.9 \%$, and $2.7 \%$ than the mwSM method. Moreover, for the Contraceptive dataset, the GMean decreased by $1.5 \%$, falling below the performance of DBSM, the current best-performing method.

For the Isolet dataset, we have two best-performing state-of-the-art methods, as we can observe in Table~\ref{tab:my-tabular2} that the metrics MCC and F1-Score increase by $8.6 \%$ and $1.3 \%$ respectively and the metrics BACC and Gmean gain by $1.3 \%$ and $1.5 \%$ respectively than SVM SMOTE.
For the Dermatology dataset, the balanced accuracy, F1-Score, and MCC metrics are equal to the RCS method for the SOICJ dataset. Also, we can observe that for the RCS method, the Gmean metric value is lower than the SOICJ method by $0.9 \%$.
For the Wine dataset, the balanced accuracy, MCC, F1-Score, and Gmean metric values have increased by $6.6 \%, 9.8\%, 9.2\%$, and $3.8 \%$ than Cure SMOTE, which is the best performing state-of-the-art method. For the Webpage dataset, the balanced accuracy and Gmean increase by $0.8 \%$ and $0.6 \%$ than the SMOTE oversampling method. However,  the MCC and F1-Score metric values improved by $10.3 \%$ and $0.5 \%$ than Kmean SMOTE.
\color{black}

\textcolor{black}{See Supplementary Section~\ref{supp_test} for \textbf{Statistical tests}.}
\color{black}
\begin{table}[!ht]
\centering
\caption{Figures depicting the TSNE-plot of the latent feature vectors from the DeepSMOTE Autoencoder and our Autoencoder for the imbalanced feature vectors and the feature vectors after applying the oversampling algorithm.}
\resizebox{0.9\textwidth}{!}{
\begin{tabular}{c c c c c}
& {DeepSMOTE AE} & {DeepSMOTE Oversampling} & {Our AE} &  {RCS Oversampling}\\ 

\raisebox{1.75\totalheight}{\rotatebox[origin=c]{90}{MNIST}} & {\includegraphics[height=1.5in,width=1.5in]{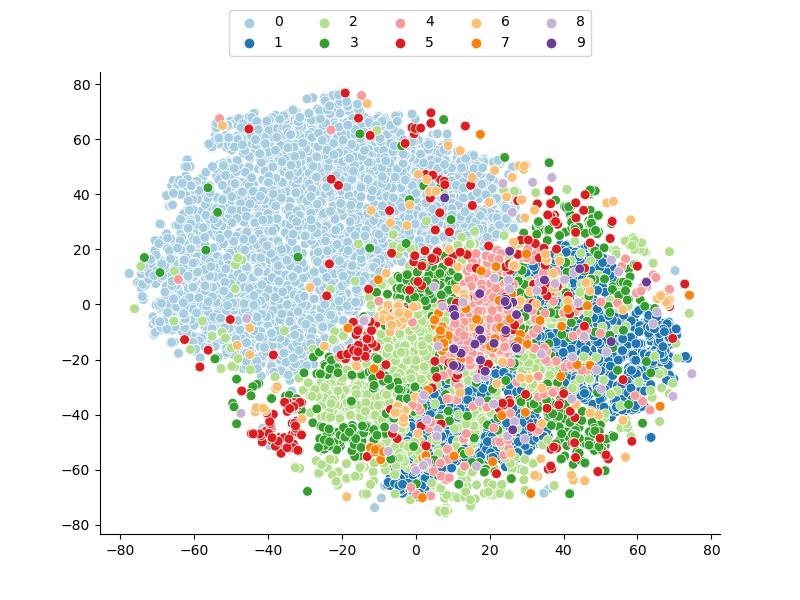}\label{fig:2a}} & {\includegraphics[height=1.5in,width=1.5in]{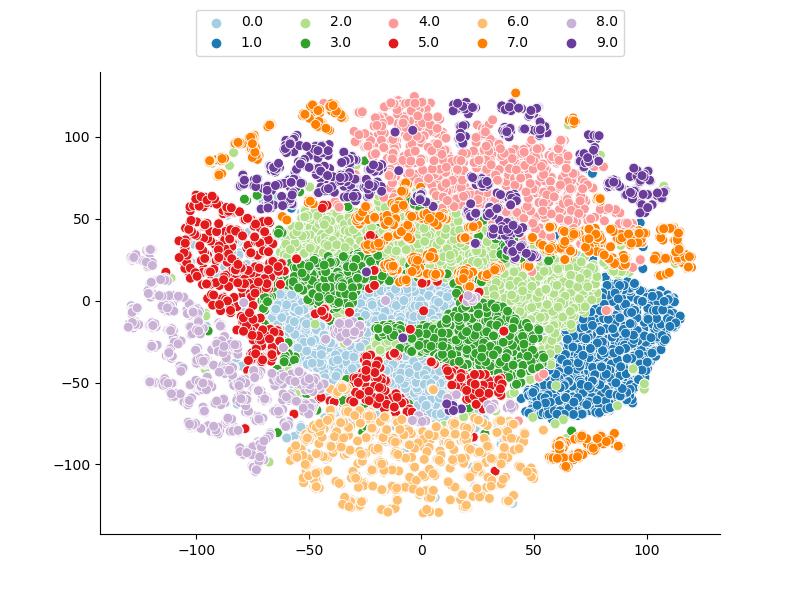}\label{fig:2b}} & {\includegraphics[height=1.5in,width=1.5in]{Images/mnist_ours_ae.jpg}\label{fig:2c}} & {\includegraphics[height=1.5in,width=1.5in]{Images/mnist_ours_over.jpg}\label{fig:2d}} \\ 

\raisebox{1.75\totalheight}{\rotatebox[origin=c]{90}{\textcolor{black}{FMNIST}}} & {\includegraphics[height=1.5in,width=1.5in]{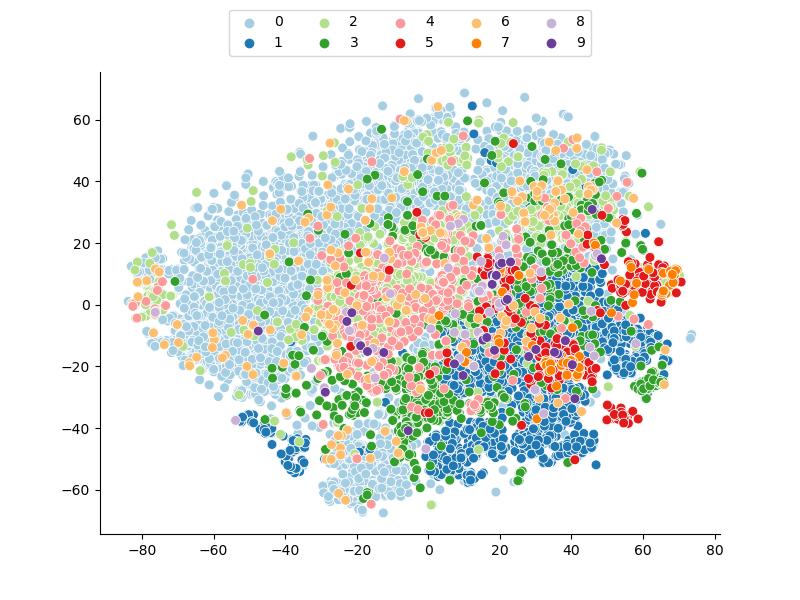}\label{fig:2e}} & {\includegraphics[height=1.5in,width=1.5in]{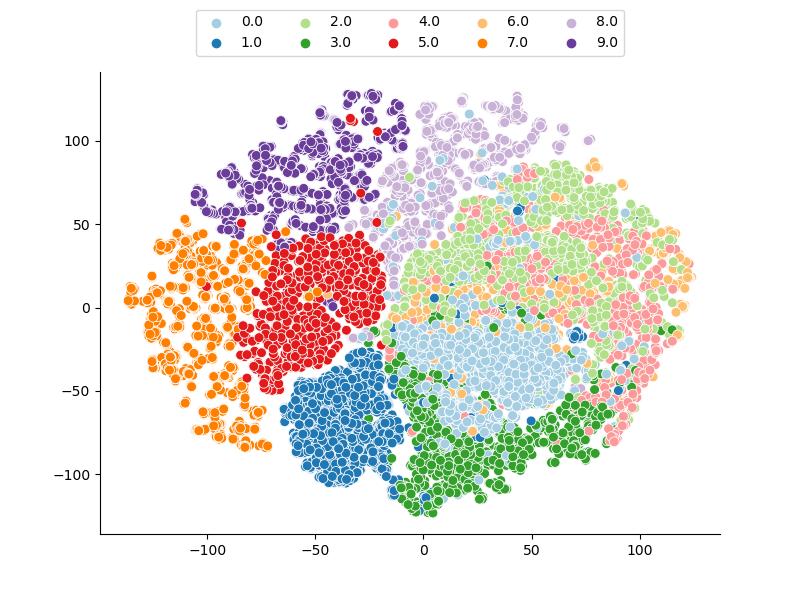}\label{fig:2f}} & {\includegraphics[height=1.5in,width=1.5in]{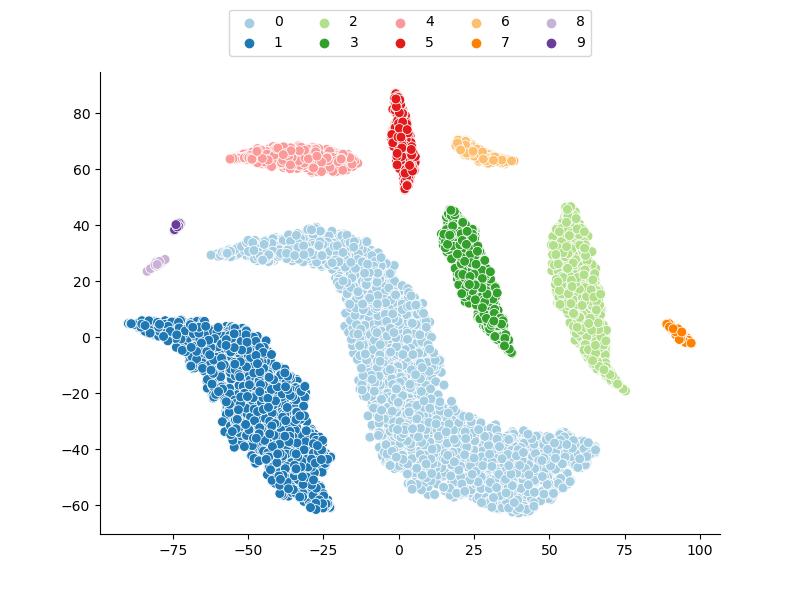}\label{fig:2g}} & {\includegraphics[height=1.5in,width=1.5in]{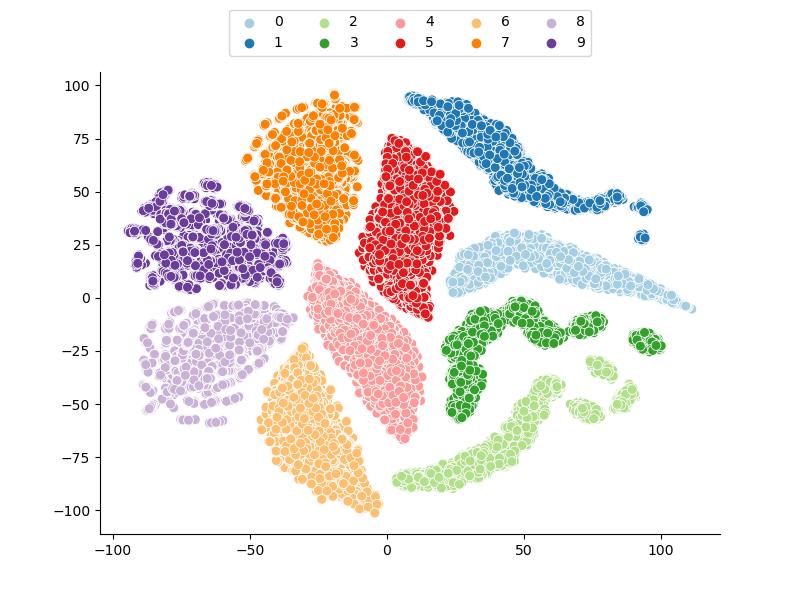}\label{fig:2h}} \\ 

\raisebox{1.75\totalheight}{\rotatebox[origin=c]{90}{CIFAR10}} &{\includegraphics[height=1.5in,width=1.5in]{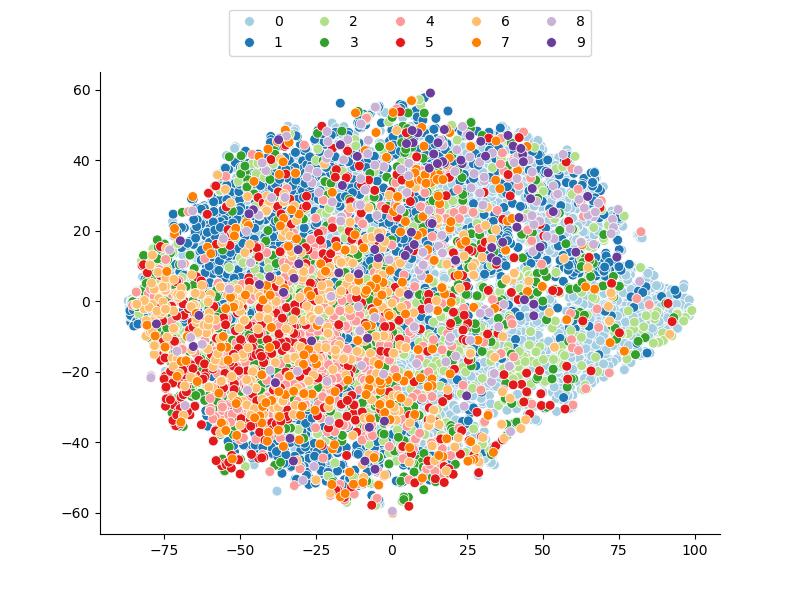}\label{fig:2i}} & {\includegraphics[height=1.5in,width=1.5in]{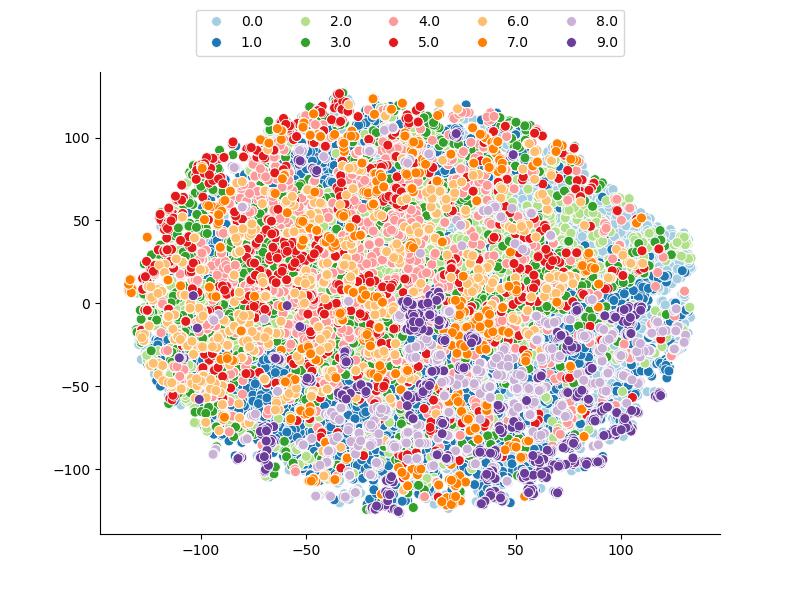}\label{fig:2j}} & {\includegraphics[height=1.5in,width=1.5in]{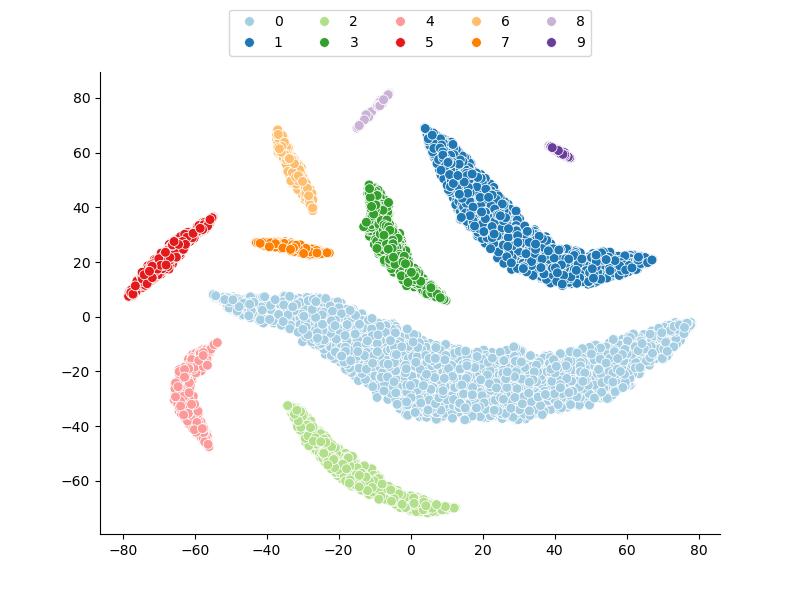}\label{fig:2k}} & {\includegraphics[height=1.5in,width=1.5in]{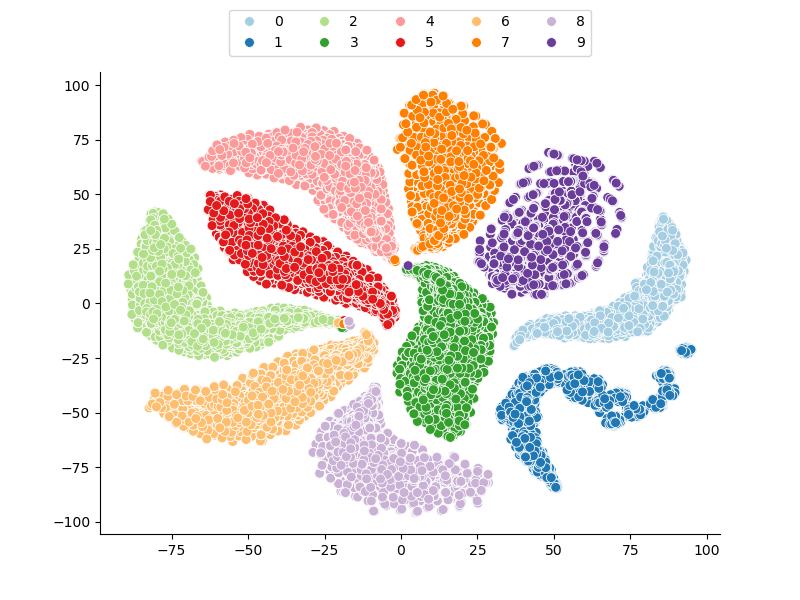}\label{fig:2l}} \\ 

\raisebox{1.85\totalheight}{\rotatebox[origin=c]{90}{SVHN}} & {\includegraphics[height=1.5in,width=1.5in]{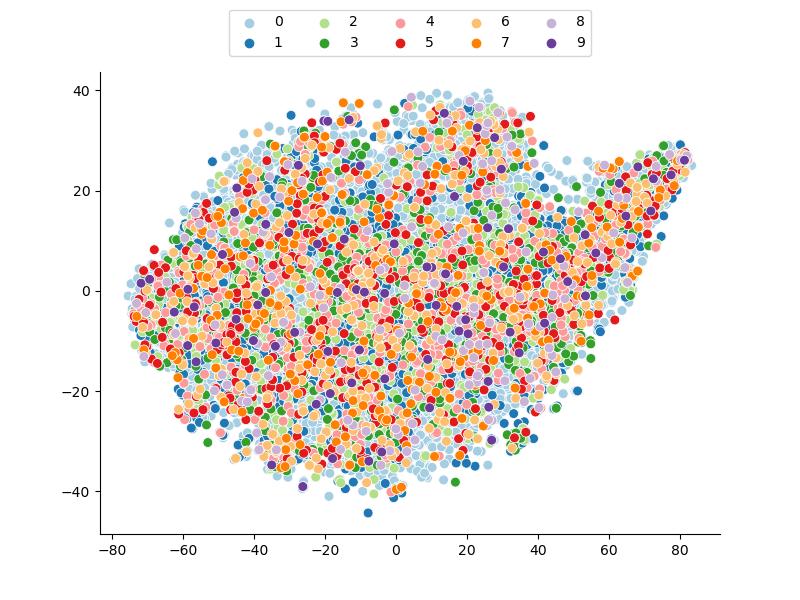}\label{fig:2m}} & {\includegraphics[height=1.5in,width=1.5in]{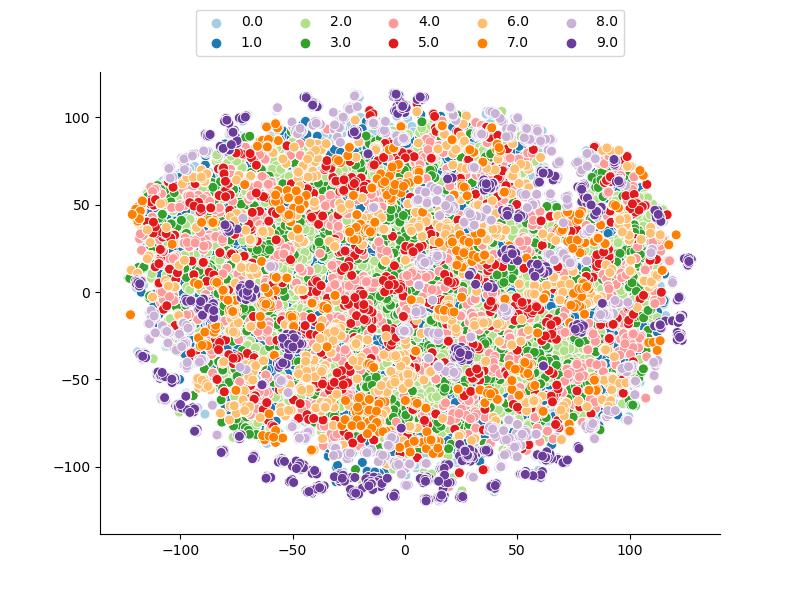}\label{fig:2n}} & {\includegraphics[height=1.5in,width=1.5in]{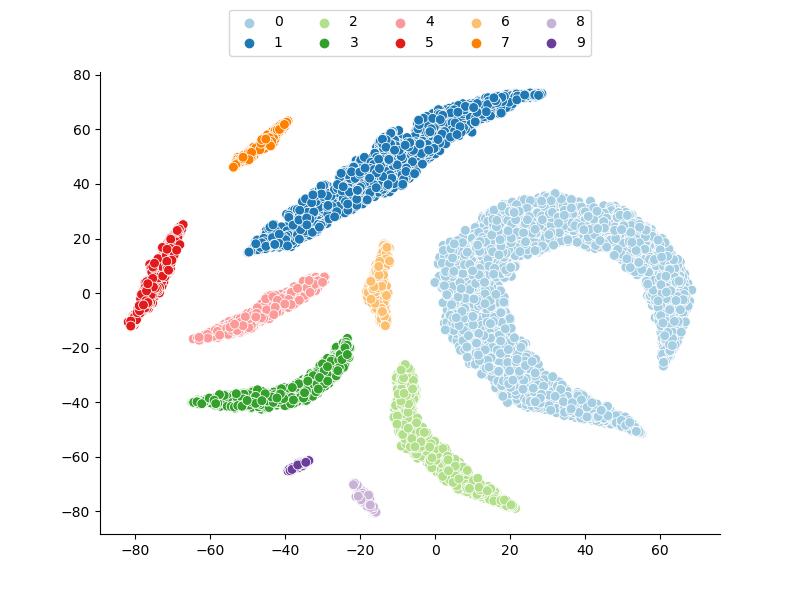}\label{fig:2o}} & {\includegraphics[height=1.5in,width=1.5in]{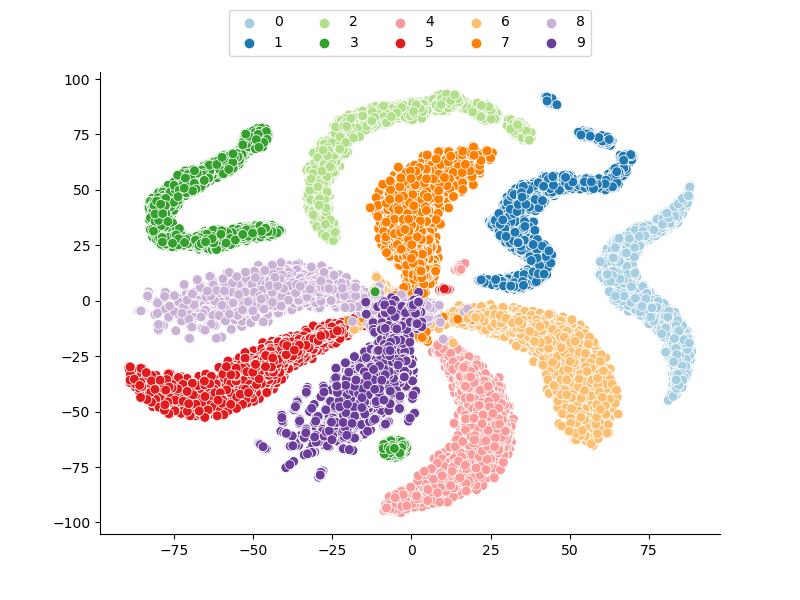}\label{fig:2p}} \\ 

\end{tabular}
}
\label{fig:2}
\end{table}
\begin{table}[!ht]
\centering
\caption{Figures depicting the Synthetic Image Generation of Various Methods and Our Method for MNIST, FMNIST, SVHN, and CIFAR10.}
\resizebox{0.6\textwidth}{!}{
\begin{tabular}{c c c c c}
& \scriptsize{{Original Image}} & \scriptsize{{cDCGAN}} & \scriptsize{{DeepSMOTE}} &  \scriptsize{{RCS}}\\ 

\raisebox{\totalheight}{\rotatebox[origin=c]{90}{\scriptsize{MNIST}}} & {\includegraphics[height=1in,width=0.6in]{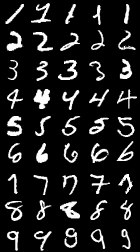}}\label{fig:3a}  & {\includegraphics[height=1.0in,width=0.6in]{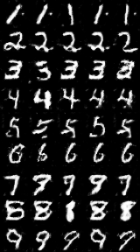}}\label{fig:3b} & {\includegraphics[height=1.0in,width=0.6in]{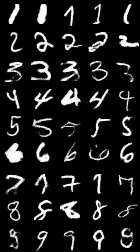}}\label{fig:3c} & {\includegraphics[height=1.0in,width=0.6in]{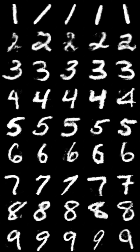}}\label{fig:3d} \\ 

\raisebox{\totalheight}{\rotatebox[origin=c]{90}{\scriptsize{FMNIST}}} & {\includegraphics[height=1.0in,width=0.6in]{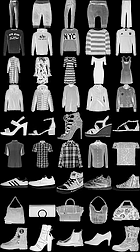}}\label{fig:3e} & {\includegraphics[height=1.0in,width=0.6in]{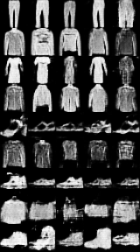}}\label{fig:3f} & {\includegraphics[height=1.0in,width=0.6in]{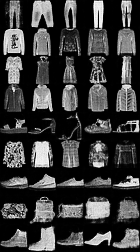}}\label{fig:3g} & {\includegraphics[height=1.0in,width=0.6in]{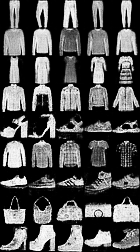}}\label{fig:3h} \\ 


\raisebox{\totalheight}{\rotatebox[origin=c]{90}{\scriptsize{SVHN}}} & {\includegraphics[height=1.0in,width=0.6in]{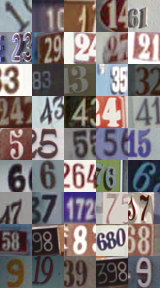}}\label{fig:3i} & {\includegraphics[height=1.0in,width=0.6in]{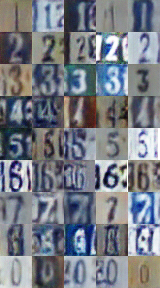}}\label{fig:3j} & {\includegraphics[height=1.0in,width=0.6in]{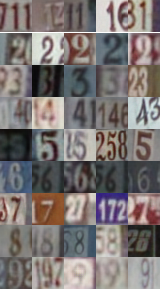}}\label{fig:3k} & {\includegraphics[height=1.0in,width=0.6in]{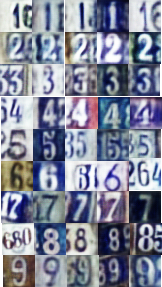}}\label{fig:3l} \\ 

\raisebox{\totalheight}{\rotatebox[origin=c]{90}{\scriptsize{CIFAR10}}} & {\includegraphics[height=1.0in,width=0.6in]{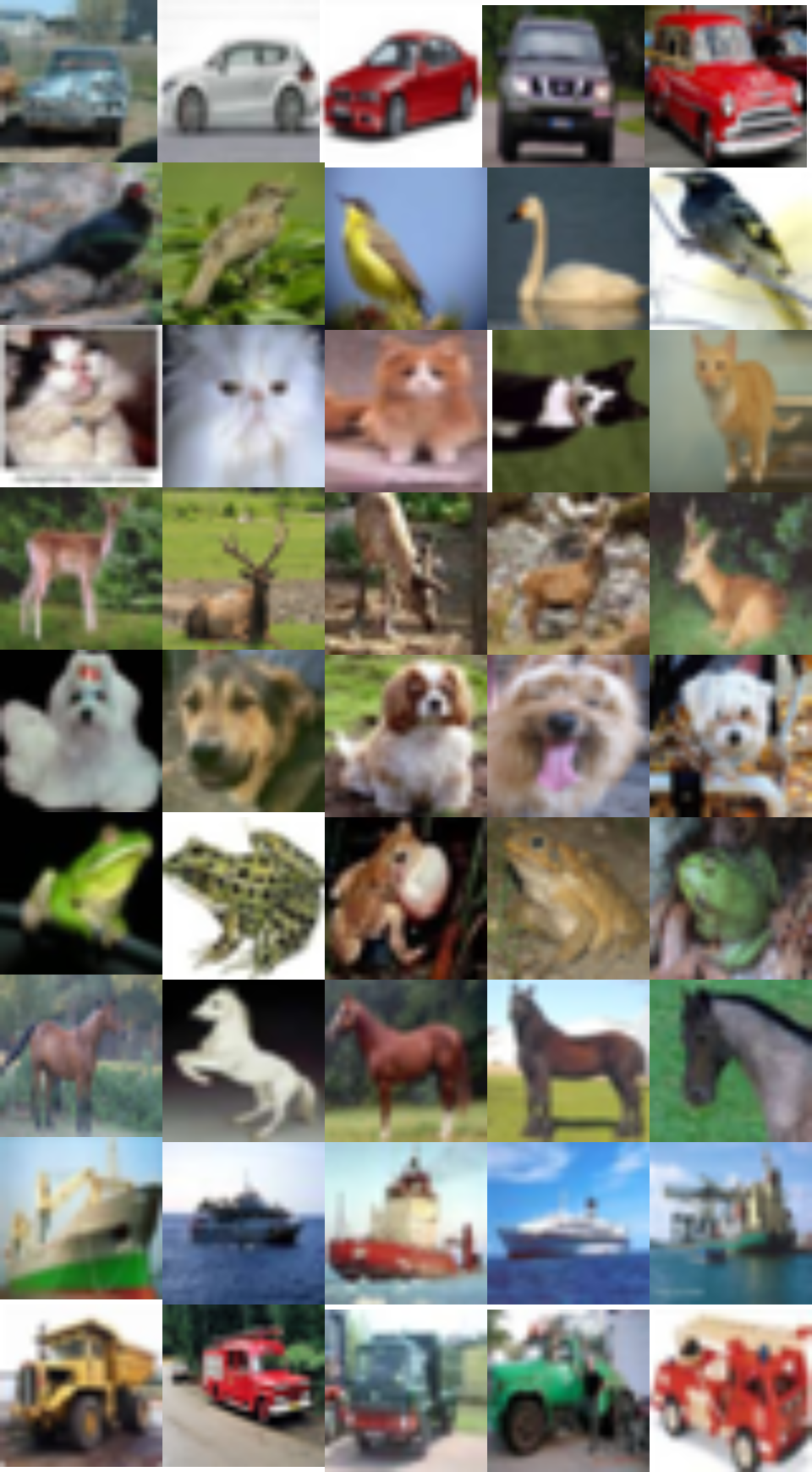}}\label{fig:3m} & {\includegraphics[height=1.0in,width=0.6in]{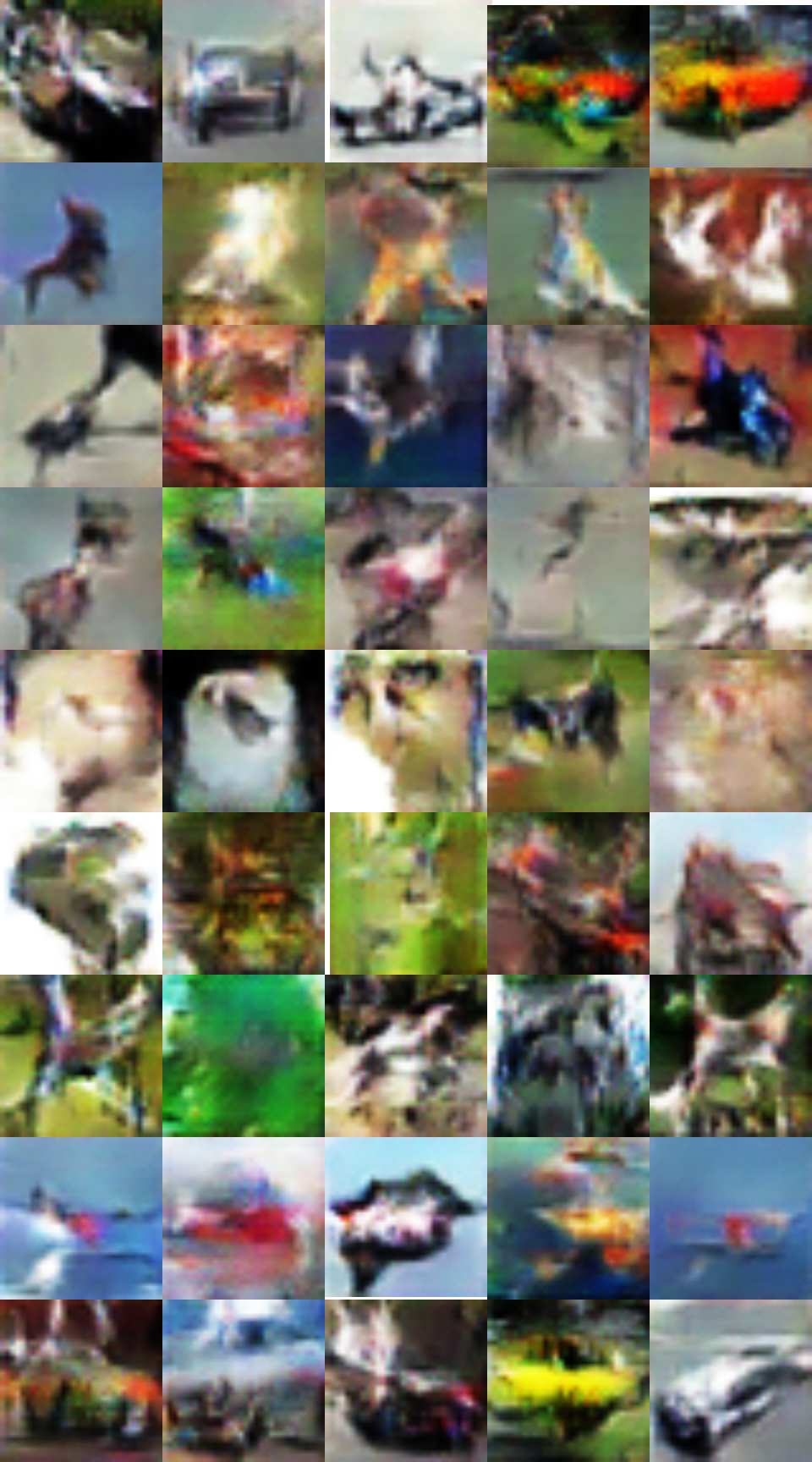}}\label{fig:3n} & {\includegraphics[height=1.0in,width=0.6in]{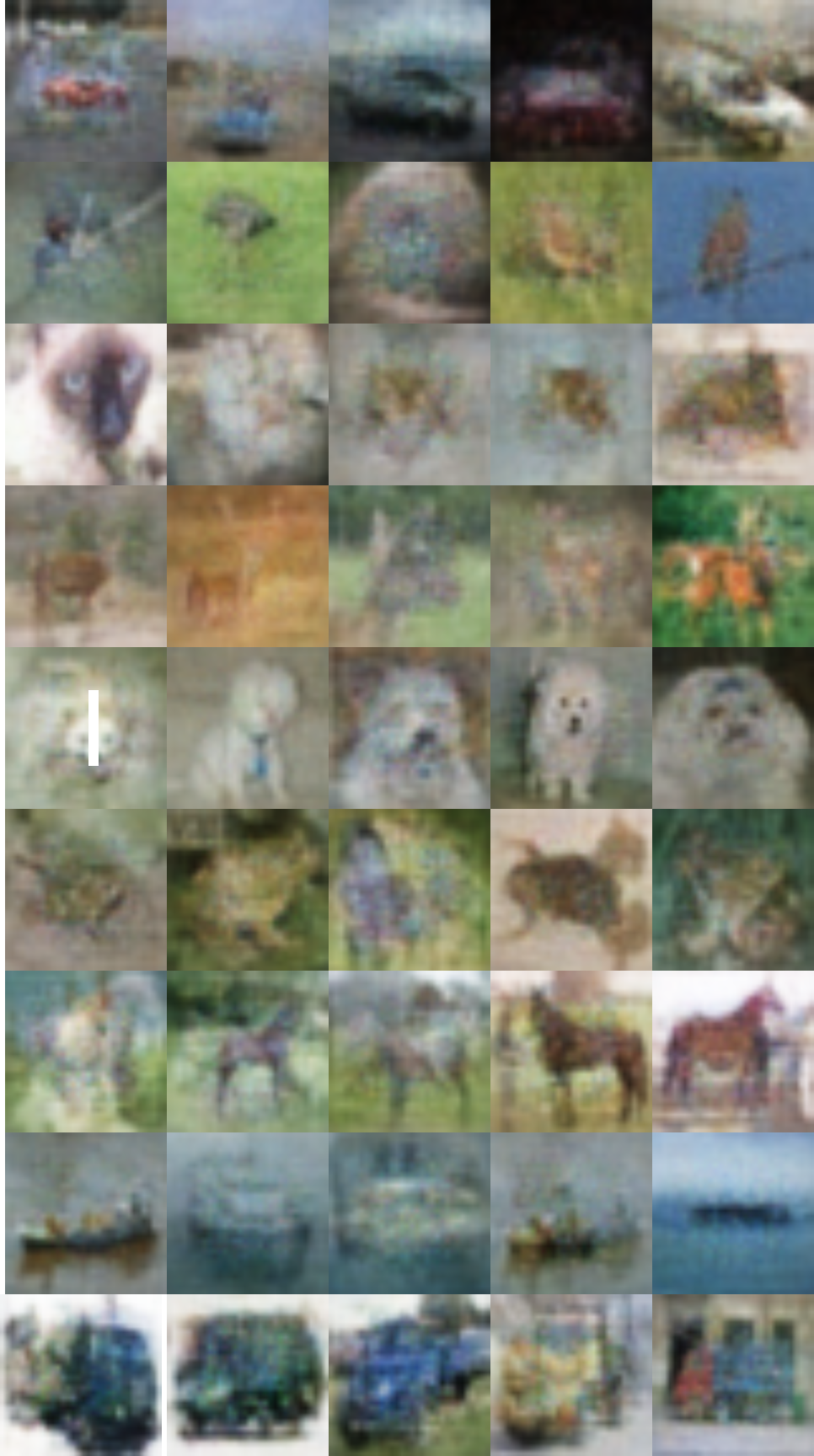}}\label{fig:3o} & {\includegraphics[height=1.0in,width=0.6in]{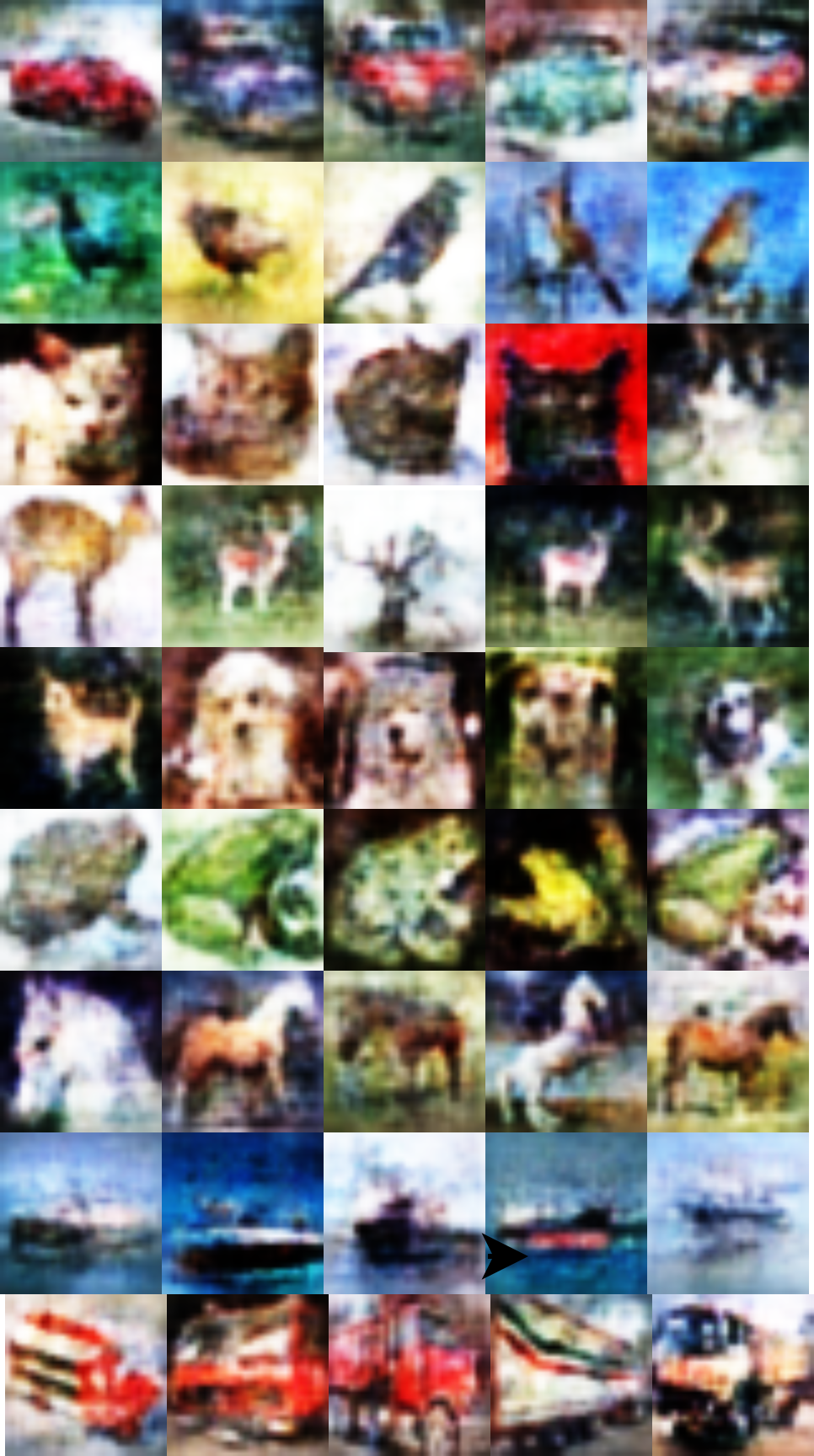}}\label{fig:3p} \\ 
\end{tabular}
}
\label{fig:3}
\end{table}
\section{Ablation Study}
We perform several ablation experiments to conduct a thorough analysis of our calibration method.

 \subsection{Visualization of our method}
\subsubsection{t-SNE plots}
\color{black}Figure~\ref{fig:2} presents the t-distributed Stochastic Neighbor Embedding (t-SNE) plots of latent vectors from MNIST, FMNIST, CIFAR10, and SVHN. Both the DeepSMOTE Autoencoder and our Autoencoder were trained for 200 epochs. After training, the imbalanced datasets were encoded into latent feature vectors, oversampled to mitigate imbalance, and then projected into two dimensions via t-SNE, with distinct colors assigned to each class.

The plots clearly show that our Autoencoder achieves better class separation than the DeepSMOTE Autoencoder under imbalanced conditions. Post-oversampling, our calibration method produced samples that aligned closely with their respective classes while preserving inter-class separation.

For simpler datasets (MNIST, FMNIST), classes remained linearly separable both before and after oversampling, reflecting the robustness of our method. For more complex datasets (SVHN, CIFAR10), classes were well separated prior to oversampling, though some overlap appeared afterward. Even so, our oversampling approach outperformed state-of-the-art methods in accuracy, underscoring its effectiveness.\color{black}
\textcolor{black}{Additionally, in Supplementary Section~\ref{clust_sep}, we present the \textbf{Clustering Separability Index}, using the Davies-Bouldin Index and Silhouette Score to empirically assess the disentanglement of latent feature vectors before and after oversampling.}
\subsubsection{Visualizing the generated images} 
The images from various oversampling methods were illustrated in Figure~\ref{fig:3}. Our RCS method posed a different strategy to generate synthetic samples than the other oversampling methods. Moreover, cDCGAN performed poorly over these datasets. We observed that these methods perform better in MNIST and FMNIST than SVHN and CIFAR10, as the former has a single channel and is easily represented in the feature space. Even though complex datasets such as CIFAR10 and SVHN did not generate images as accurate as ground-truth images, we observe a better accuracy than state-of-the-art methods, as we have discussed earlier. We visualized the generated images for MNIST, FMNIST, SVHN, and CIFAR10 datasets for Original Image, Conditional DCGAN, DeepSMOTE, and RCS Method. We did not include the TailCalibX calibration method since the generated images were not good enough.

To further validate the robustness of our method, we conducted extensive ablation studies analyzing the impact of key hyperparameters, architectural components, and design choices; detailed results are provided in the supplementary material section \ref{supp:add_ablation}.

\color{black}

\color{black}
\section{Conclusion and Future Work}
In real-world applications, data distributions are often inherently skewed, with class distributions typically being independent and well-separated in the feature space. Under such imbalanced conditions, conventional training paradigms tend to favor the majority class, thereby impeding the effective learning of discriminative representations for minority class samples.
To address this, we propose an end-to-end framework specifically designed for imbalanced learning scenarios. At its core, we train an autoencoder to preserve class separability by ensuring a smooth transition of structure from the data space to the latent space. To enhance prior efforts in distribution calibration, we introduce a novel calibration strategy that emphasizes the local statistical properties of feature vectors corresponding to minority class instances, while also incorporating intermediate class samples through targeted oversampling.
Our proposed RCS algorithm consistently outperforms existing SOTA methods across multiple datasets and modalities. Beyond empirical gains, this work shows distribution calibration’s potential for latent-space synthetic sample generation, laying a foundation for future research.
 
\bibliographystyle{elsarticle-num} 
\bibliography{references.bib}
\clearpage
\thispagestyle{plain} 

\begin{center}
  {\LARGE Supplementary Material for Rebalancing with Calibrated Sub-classes (RCS): A Statistical Fusion-based Framework for Robust Imbalanced Classification across Modalities}\\[1.0\baselineskip]

  {\large
    Priyobrata Mondal\textsuperscript{a}, 
    Faizanuddin Ansari\textsuperscript{a},
    Swagatam Das\textsuperscript{a}
  }\\[0.5\baselineskip]

  {\footnotesize \textsuperscript{a}\textit{Electronics and Communication Sciences Unit, Indian Statistical Institute, Kolkata, 700108, West Bengal, India}}
\end{center}

\vspace{1em} 

\phantomsection\label{sec:supplementary}

\setcounter{section}{0}
\setcounter{table}{0}
\setcounter{figure}{0}
\renewcommand{\thesection}{S\arabic{section}}
\renewcommand{\thetable}{S\arabic{table}}
\renewcommand{\thefigure}{S\arabic{figure}}
 \makeatother

\renewcommand{\bibnumfmt}[1]{[S#1]}
\renewcommand{\citenumfont}[1]{\textit{#1}}

\hypersetup{colorlinks=true,allcolors=black}

\WarningFilter*{hyperref}{Suppressing link with}
\WarningFilter*{hyperref}{Token not}
\WarningFilter{latex}{Citation}
\WarningFilter{latex}{Overfull}
\WarningFilter{latex}{Underfull}

\externaldocument{main}

\makeatletter
\long\def\pprintMaketitle{\clearpage
  \iflongmktitle\if@twocolumn\let\columnwidth=\textwidth\fi\fi
  \resetTitleCounters
  \def\baselinestretch{1}%
  \printFirstPageNotes
  \begin{center}%
 \thispagestyle{pprintTitle}%
 \def\baselinestretch{1}%
    \Large\@title\par\vskip18pt
    \normalsize\elsauthors\par\vskip10pt
    \footnotesize\itshape\elsaddress\par\vskip36pt
    \ifvoid\absbox\else\unvbox\absbox\par\vskip10pt\fi
    \ifvoid\keybox\else\unvbox\keybox\par\vskip10pt\fi
    \end{center}%
  \gdef\thefootnote{\arabic{footnote}}%
  }
\makeatother

\section{Detailed Discussion of Related Works}
\label{supp:related_work}
Over the past two decades, conventional machine learning models excluding deep learning have been extensively used to address class imbalance. Despite the growing popularity and recent advances in deep learning, empirical studies on its application to imbalanced scenarios remain limited. Approaches to imbalance mitigation can be broadly categorized into data-level, algorithm-level, and hybrid strategies \cite{das2022supervised}. Data-level techniques focus on resampling, while algorithm-level methods \cite{NEURIPS2019_621461af,wang_et_al_2016,lin2017focal} modify learners via re-weighting or cost-sensitive adjustments to reduce majority bias; hybrid methods integrate both. Since the crux of imbalance lies in sparse minority distributions, oversampling has emerged as a standard remedy by artificially enriching minority samples. Classical oversampling methods such as SMOTE and its variants are effective for low-dimensional data (e.g., tabular), whereas deep generative oversampling leverages neural networks to synthesize realistic minority samples, proving particularly effective for high-dimensional domains like images.

\textbf{Classical oversampling methods for handling Class Imbalance:} These include resampling-based methods such as the Random oversampling which randomly samples datapoints from the data-space, Synthetic Minority Over-Sampling Technique (SMOTE)~\cite{chawla2002smote} generates synthetic samples by binary interpolation in the minority class for each datapoint and a neighboring point among its k-nearest neighbor, and Border-line SMOTE~\cite{han2005borderline} identifies and generates samples near the decision boundary region. ClusterSMOTE~\cite{ClusterSMOTE_2006} integrates a 1-Nearest Neighbor (1NN) outcast handling mechanism to locate and remove noisy or outcast minority instances before oversampling. Distance SMOTE \cite{DistSMOTE_2007} generates synthetic samples by considering the distribution of the minority class. Another algorithm, DBSMOTE (Density-Based Synthetic Minority Over-sampling Technique)~\cite{2012dbsmote} uses DBSCAN~\cite{dbscan1996} clustering algorithm and Dijkstra~\cite{dijkstra2022note} algorithm to generate synthetic samples in the minority class. ADASYN~\cite{Adasyn_2008} addresses the challenge of skewed class distributions, where the method generates synthetic samples by adjusting the class distribution by focusing on the regions of the minority class where the distribution is sparse. SVM-SMOTE~\cite{SVM2012} combines SMOTE and SVM algorithms~\cite{vapnik2013nature} to generate synthetic samples in the boundary regions. CURE-SMOTE~\cite{ma2017cure} uses "CURE" to cluster the minority class instances using the hierarchical clustering algorithm. The representative samples from each cluster are then chosen to generate synthetic samples in the minority class. The CURE-SMOTE has a disadvantage, as it deletes both the boundary and the noisy samples.
Sanchez et al. proposed Synthetic Oversampling of Instances using Clustering and Jittering (SOICJ)~\cite{sanchez2013SOICJ} effectively capturing the underlying data distribution. It finds the standard deviation (\textit{std1}) between each point in the minority class and the standard deviation (\textit{std2}) between the minority instances in a particular cluster. It then finds the minimum of \textit{std1} and \textit{std2} as \textit{std}. The new synthetic samples are generated by the formula $new = current + s \times \textit{std}$, where $s \in (-1,1)$.  The problem with these oversampling techniques is that they do not consider the majority of instances in the feature space, and hence, the generated samples will not align with the original distribution of the given data points. 
Zhao et al. proposed the $K$-means SMOTE~\cite{zhao2021kmeans} algorithm, which first clusters the data space using the Kmeans algorithm~\cite{morissette2013k} and then applies the SMOTE oversampling technique to the specified cluster. 
Another oversampling algorithm, the MeanShiftSMOTE (MSMOTE) algorithm, was proposed~\cite{ashour2022adaptive}, which applies Mean Shift~\cite{meanshift2002} clustering to identify dense regions in the minority class, followed by generating synthetic samples using SMOTE within these clusters.
Cluster Ensembles SMOTE (CE-SMOTE)~\cite{CESMOTE2010} combines SMOTE with a non-linear dimensionality reduction method, Isomap. This algorithm captures the intrinsic geometry of the data and ensures better sample generation. On the other hand, Majority Weighted Minority Oversampling (MWMOTE)~\cite{MWMOTE2014} finds the nearest majority-class neighbors and provides weights to the minority-class samples, followed by generating synthetic samples using the SMOTE algorithm. The drawback of this method is that it removes minority-class samples, which lose valuable information, thereby affecting model generalization.

\textbf{Deep Generative Model-based Oversampling:} However, in dealing with high-dimensional data, the quality of synthesized data points may be compromised due to noise and poor distance measurement. Recent research in deep learning suggests novel approaches for handling complex, imbalanced data. Two primary aspects have gained attention: data perturbation through resampling strategies applied to deep learning models and model perturbation involving uncertainty in model parameters or loss functions. To address the skewed distribution caused by imbalanced data, deep generative models like Variational Autoencoder (VAE)~\cite{kingma2013auto} and Generative Adversarial Network (GAN)\cite{goodfellow2014generative} are introduced to generate synthetic samples in the original feature space for imbalanced classification.
While VAE struggles to create artificial data with fine details, Guo et al. \cite{guo2019discriminative} propose modeling latent representation using two Gaussian distributions with opposite means, specifically for binary classification. Unfortunately, this concept does not apply to imbalanced multiclass data.  
Conditional GAN (cGAN) \cite{gauthier2014conditional} is used to generate class-specific minority samples, considering differences between classes. However, random noise in GAN-based methods may lead to entangled processes and disrupt orientation-related features. To address this, BAGAN \cite{mariani2018bagan} combines Autoencoder (AE), and cGAN in a two-step framework.

The methods described above follow a two-stage strategy, where one stage generates synthetic data, and the other trains a classifier on the augmented data. Despite effectively leveraging data perturbation for imbalance classification performance, there may be a gap between data generation and classifier training. A generative adversarial minority oversampling (GAMO) method \cite{mullick2019generative} proposes a three-player adversarial game involving a convex generator, a multiclass classifier network, and a real/fake discriminator. GAMO generates new samples within the convex hull of actual minority-class samples. Arnab et al. ~\cite{pmlr-v206-mondal23a} use the original data samples to train a regularized autoencoder using a linear classifier and a mixer network. It then uses the latent vectors from the trained autoencoder to generate synthetic samples using the convex combination of all the latent feature vectors in the minority class. The drawback of the GAMO and this method is that for the generation of a single feature vector in the minority class, the samples for all the feature vectors have been taken into consideration.

To address the drawbacks arising from the above methods, \color{black}Dablian et al.\color{black} ~\cite{dablain2021deepsmote} proposed the DeepSMOTE method, which trains the autoencoder using the reconstruction loss and the permutation loss to preserve the classes. The latent feature vectors from this autoencoder are extracted, and oversampling is performed in the minority class using the SMOTE algorithm.

The consideration of sample space distribution remains unaddressed by the methods discussed thus far. Wang et al. proposed "Deep Generative Process for Robust Imbalance Classification"(DGC)~\cite{wang2020deep}, which uses Bayesian inference to generate the image and labels from the latent space. This framework uses a mixture model as a generator network that captures the complex distribution of the majority and minority classes, and the discriminator model has been designed to estimate the class priors from the generative model. Using the same concept, the authors of this paper published an extension of this work, DGCMM~\cite{wang2022deep}, in 2023.

\section{Proof of Theorems}
\subsection{Proof of Theorem~\ref{thm}}
\label{app_thm1}
\begin{proof}
    Let the mean associated with the datapoint $l_j\in\mathscr{m}$ be $\mu_j$.\\
    Then, \\
    \begin{equation}
        \begin{split}
            & \bar{\mu}_j = l_j + \Sigma_{i=1}^{k}(\hat{\mu}_i - l_j)w_i\\
            & \Rightarrow \bar{\mu}_j = \Sigma_{i=1}^k\hat{\mu}_iw_i+l_j-\Sigma_{i=1}^kw_il_j\\
            & \Rightarrow \bar{\mu}_j = w_1\times\hat{\mu}_1+w_2\times\hat{\mu}_2+\cdots+w_k\times\hat{\mu}_k + ( 1-\Sigma_{i=1}^kw_i)l_j\\
        \end{split}
    \end{equation}
    Now, consider the coefficients of $\hat{\mu}_1,\hat{\mu}_2,\cdots,\hat{\mu}_j$ and $l_j$,\\
    $w_1+w_2+\cdots w_k+(1-\Sigma_{i=1}^kw_i) = 1$.\\
    Hence, we can observe that $\bar{\mu}_j$ is a convex combination of $\hat{\mu}_i$'s and $l_j$ and hence $\bar{\mu}_j$ lies in the convex hull region of the $\hat{\mu}_i$'s.\\
    In the second part, we want to show that the estimated mean $\bar{\mu}_i$ lies closer to $l_j$. We know from our definition of the weights ($w_i$) for the $i^{th}$ component is, $w_i = \frac{1}{S_i}$, $S_i$ being the number of elements in the $i^{th}$ component. Now for $k$ many components 
    $\sum_{i=1}^k w_i = \sum_{i=1}^k \frac{1}{S_i}$.  Hence, the value $(1-\sum_{i=1}^k w_i)$ is greater than $\sum_{i=1}^k w_i$ based on the values of $k$ that we have used in our algorithm. Therefore, the coefficient of $l_j$ is a higher quantity and thus the estimated mean lies closer to the estimated mean $\hat{\mu}_j$.
\end{proof}
\subsection{Proof of Theorem~\ref{thm1}}
\label{app_thm2}
\begin{proof}
    Given the sample mean $l^{\prime} = \frac{1}{n}(\Sigma_{i=1}^nl_i)$\\
    Then, 
        \begin{equation}
        \begin{split}
    & E[l^{\prime}] = E[\frac{1}{n}(\Sigma_{i=1}^nl_i)]\\
    & \Rightarrow E[l^{\prime}] = \frac{1}{n}E[(\Sigma_{i=1}^nl_i)]\\
    & \Rightarrow E[l^{\prime}] = \frac{1}{n}\Sigma_{i=1}^nE[(l_i)]\\
    & \Rightarrow E[l^{\prime}] = \frac{1}{n}\Sigma_{i=1}^n\bar{\mu}_j\\
    & \Rightarrow E[l^{\prime}] = \bar{\mu}_j
            \end{split}
    \end{equation}
\end{proof}

\subsection{Proof of Theorem~\ref{thm2}}
\label{app_thm3}
\begin{proof}
\begin{equation}
        \begin{split}
    & \rho(g*x) = \Tilde{f}of (x) \\
    & \Rightarrow \rho(g*x) = \Tilde{f}(g* f(x))\\
    & \Rightarrow \rho(g*x) = \Tilde{f}(g* f(x)) \hspace{0.2cm} \text{from the equivariance of } f\\
    & \Rightarrow \rho(g*x) = g*(\Tilde{f}(f(x))) \hspace{0.2cm} \text{from the equivariance of } \Tilde{f}\\
    & \Rightarrow \rho(g*x) = g*(\Tilde{f}of(x)) \hspace{0.2cm} \\
    & \Rightarrow \rho(g*x) = g*(\rho(x)) \hspace{0.2cm} \\
        \end{split}
    \end{equation}
\end{proof}

\subsection{Hyperparameter used for Image and Tabular Data}
For our method, we have found the optimal values for parameters such as the temperature \( t \) used in the loss function of autoencoder training, the parameter \( k \) representing the number of neighbors required for mean and variance calibration, and the parameter \( \eta \), which determines the minimum number of elements necessary for evaluating minority and intermediate classes. In Table \ref{tab:ablation_tuning}, we have depicted the parameters, i.e., Temperature (\( t \)), Nearest Neighbor (\( k \)), and \( \eta \), for which we have attained the best performance for the image datasets. Table \ref{tab:parameters_tab} presents the hyperparameters for the tabular data. Since we directly apply our data calibration technique to the tabular method, the hyperparameters we use here are \( k \) and \( \eta \).

\begin{table}
\centering
\caption{Optimal values of the Hyperparameters used in our experiments for Image Datasets}
\label{tab:ablation_tuning}
\resizebox{\textwidth}{!}{%
\begin{tabular}{lccccccccc}
\toprule
\textbf{Hyper-} & \textbf{MNIST$_{\textbf{28}\times\textbf{28}}$} & \textbf{FMNIST$_{\textbf{28}\times\textbf{28}}$} & \textbf{CIFAR10$_{\textbf{32}\times\textbf{32}}$} & \textbf{SVHN$_{\textbf{32}\times\textbf{32}}$} & \textbf{GTSRB$_{\textbf{32}\times\textbf{32}}$} & \textbf{STL10$_{\textbf{32}\times\textbf{32}}$} & \textbf{SUN397$_{\textbf{32}\times\textbf{32}}$} & \textbf{SUN397$_{\textbf{64}\times\textbf{64}}$} \\
\textbf{parameter} & & & & & & & & \\ 
\midrule
\textbf{t} & 0.07 & 0.01 & 0.01 & 0.07 & 0.05 & 0.05 & 0.03 & 0.01 \\
\textbf{k} & 5 & 3 & 5 & 6 & 5 & 3 & 6 & 3 \\
\textbf{$\eta$} & 7 & 7 & 4 & 5 & 5 & 7 & 6 & 6 \\
\bottomrule
\end{tabular}
}
\end{table}
\begin{table}[!ht]
\caption{Optimal values of the hyperparameters for the Tabular Datasets used in Our Experiment.}
\label{tab:parameters_tab}
\resizebox{\textwidth}{!}{%
\begin{tabular}{lcccccc}
\hline
\textbf{Hyperparameter} & \textbf{Contraceptive} & \textbf{Isolet} & \textbf{Webpage} & \textbf{Optical\_Digits} & \textbf{Dermatology} & \textbf{Wine} \\ \hline
\textbf{k} & 3 & 3 & 3 & 3 & 3 & 3 \\
\textbf{$\eta$} & 1.7 & 11 & 34 & 9 & 2 & 1.3 \\ \hline
\end{tabular} }
\end{table}

\section{Discussion on Textual Datasets}

\subsection{Dataset Description}
\label{textual_dataset}
For textual dataset, we considered Spam or ham data \footnote{https://www.kaggle.com/code/karnikakapoor/spam-or-ham-sms-classifier}, spam or not spam dataset\footnote{https://www.kaggle.com/datasets/ozlerhakan/spam-or-not-spam-dataset}, Emotion dataset~\cite{lhoest2021datasets}, and the Hate Speech dataset \footnote{https://www.kaggle.com/datasets/mrmorj/hate-speech-and-offensive-language-dataset}. Spam or ham dataset and Spam or not spam were datasets with binary classes having imbalanced ratios of $5$ and $6.46$, respectively. Emotion and hate speech datasets were multi-class datasets having imbalanced ratios of $9.37$ and $13.47$, respectively. In Table~\ref{tab:text_desc}, we provided a detailed description of the textual datasets used in our experiments.
\begin{table*}[]
\caption{\textcolor{black}{Summary of the Textual Datasets}}
\label{tab:text_desc}
\color{black}
\resizebox{\textwidth}{!}{%
\begin{tabular}{|l|c|c|l|l|}
\hline
\textbf{Datasets}      & \multicolumn{1}{l|}{\textbf{Classes}} & \multicolumn{1}{l|}{\textbf{IR}} & \textbf{\begin{tabular}[c]{@{}l@{}}Training Set (For each\\ Class for all Splits)\end{tabular}} & \textbf{\begin{tabular}[c]{@{}l@{}}Testing Set (For each\\ Class for all Splits)\end{tabular}} \\ \hline
\textbf{Emotion}       & 6                                     & 9.37                             & 5362, 4666, 2159, 1937, 1304, 572                                                               & 5362, 4666, 2159, 1937, 1304, 572                                                              \\
\textbf{Spam Not Spam} & 2                                     & 5                                & 2000, 400                                                                                       & 500, 100                                                                                       \\
\textbf{Spam or Ham}   & 2                                     & 6.46                             & 3377, 523                                                                                       & 724, 112                                                                                       \\
\textbf{Hate Speech}   & 3                                     & 13.47                            & 15358, 3328, 1140                                                                               & 3832, 835, 290                                                                                 \\ \hline
\end{tabular}}
\end{table*}

\subsection{Experimental Setup for Textual dataset}
\label{textual_data_discussion}
We performed experiments on 4 text datasets: Spam or ham data, spam or not spam dataset, Emotion dataset, and the Hate Speech dataset described above. We began by carefully pre-processing the textual datasets to ensure they were clean and ready for analysis. This step included tokenizing the text, converting it to lowercase, and removing unnecessary characters or stopwords. Once the data was pre-processed, we used a pre-trained BERT model to generate sentence embeddings. These embeddings provided meaningful, high-dimensional representations of the text, capturing the context and semantics of each sentence in a way that simpler methods could not.
Next, we trained an autoencoder to distil these embeddings into more compact latent feature vectors. The autoencoder worked by compressing the high-dimensional data into a smaller, more informative representation and then reconstructed it to ensure no critical information was lost. These latent vectors were then used for oversampling to address imbalances in the dataset, creating new synthetic examples for underrepresented classes. Finally, we trained a multi-layered perceptron (MLP) for classification. 

\subsection{Hyperparameters Used for Textual Datasets}
\label{ablation_textual}
\textcolor{black}{Table~\ref{tab:hyp_text} presents the optimal hyperparameter values for our method on the textual datasets.}
\begin{table}
\centering
\caption{\color{black}Optimal values of the hyperparameters for the Textual Datasets used in Our Experiment.\color{black}}
\label{tab:hyp_text}
\color{black}
\resizebox{0.7\textwidth}{!}{%
\begin{tabular}{lcccc}

\toprule
\textbf{Hyperparameter}      & \multicolumn{1}{l}{\textbf{Emotion}} & \multicolumn{1}{l}{\textbf{Spam Not Spam}} & \multicolumn{1}{l}{\textbf{Spam or Ham}} & \multicolumn{1}{l}{\textbf{Hate Speech}} \\ \midrule
\textbf{t}                   & 0.09                                 & 0.03                                       & 0.09                                     & 0.07                                     \\
\textbf{k}                   & 3                                    & 3                                          & 3                                        & 3                                        \\
\textbf{$\eta$} & 3                                    & 2                                          & 2                                        & 2                                        \\ \bottomrule
\end{tabular}}
\end{table}
\color{black}

\subsection{Results and Analysis of the Textual Dataset}
\label{textual_result_discussion}
Our proposed RCS method consistently outperformed all baseline and state-of-the-art oversampling techniques across all textual datasets. On the \textit{Emotion} dataset, RCS achieved the highest BACC, MCC, F1-Score, and Gmean, with improvements of over 20\% in BACC and 10--20\% in other metrics compared to the best-performing baselines. For the \textit{Spam Not Spam} dataset, RCS significantly boosted BACC and MCC by over 14\% and 40\%, respectively, and achieved the highest F1-Score and Gmean of 0.943. On the \textit{Spam SMS} dataset, our method surpassed all competitors, showing 5--12\% gains across all metrics, demonstrating strong generalization and robustness. Finally, on the \textit{Hate Speech} dataset, RCS led in every metric, notably attaining a BACC of 0.749 and MCC of 0.639-substantially higher than all other methods, which highlighted its effectiveness in handling complex, imbalanced textual data.
\begin{table}[]
\caption{\textcolor{black}{Results on Textual Data}}
\label{tab:my-table_text}

\resizebox{\textwidth}{!}{%
\color{black}
\begin{tabular}{cccccccccccccccc}
\toprule
\textbf{Datasets} & \textbf{Metrics} & \textbf{Baseline}         & \textbf{SM}               & \textbf{ADA}              & \textbf{blSM}             & \textbf{clSM}             & \textbf{cuSM}             & \textbf{DBSM}             & \textbf{kmSM}             & \textbf{msSM}             & \textbf{mwSM}             & \textbf{SOICJ}            & \textbf{svmSM}            & \textbf{csSM}             & \textbf{Ours}                      \\ \midrule
                                          & \textbf{BACC}                            & {0.266} & {0.247} & {0.250}  & {0.261} & {0.239} & {0.223} & {0.235} & {0.230}  & {0.255} & {0.255} & {0.226} & {0.254} & {0.257} & {\textbf{0.514}} \\
                                          & \textbf{}                                & $\pm$0.001   & $\pm$0.035   & $\pm$0.036   & $\pm$0.028   & $\pm$0.029   & $\pm$0.035   & $\pm$0.014   & $\pm$0.033   & $\pm$0.029   & $\pm$0.028   & $\pm$0.022   & $\pm$0.047   & $\pm$0.033   & \textbf{$\pm$0.005}   \\
                                          & \textbf{MCC}                             & {0.308} & {0.116} & {0.127} & {0.110}  & {0.083} & {0.114} & {0.104} & {0.083} & {0.110}  & {0.116} & {0.128} & {0.100}   & {0.122} & {\textbf{0.417}} \\
\multirow{-2}{*}{\rotatebox[origin=c]{90}{\textbf{Emotion}}}        & \textbf{}                                & $\pm$0.004   & $\pm$0.059   & $\pm$0.067   & $\pm$0.031   & $\pm$0.093   & $\pm$0.073   & $\pm$0.035   & $\pm$0.064   & $\pm$0.061   & $\pm$0.051   & $\pm$0.086   & $\pm$0.039   & $\pm$0.072   & \textbf{$\pm$0.007}   \\
                                          & \textbf{F1-Score}                        & {0.390}  & {0.208} & {0.239} & {0.197} & {0.191} & {0.254} & {0.193} & {0.133} & {0.183} & {0.201} & {0.271} & {0.174} & {0.191} & {\textbf{0.505}} \\
                                          & \textbf{}                                & $\pm$0.002   & $\pm$0.067   & $\pm$0.103   & $\pm$0.043   & $\pm$0.108   & $\pm$0.082   & $\pm$0.097   & $\pm$0.103   & $\pm$0.096   & $\pm$0.089   & $\pm$0.074   & $\pm$0.051   & $\pm$0.118   & \textbf{$\pm$0.006}   \\
                                          & \textbf{Gmean}                           & {0.484} & {0.458} & {0.461} & {0.470}  & {0.449} & {0.435} & {0.446} & {0.440}  & {0.465} & {0.466} & {0.438} & {0.463} & {0.467} & {\textbf{0.493}} \\
\multirow{-4}{*}{\textbf{}}               & \textbf{}                                & $\pm$0.001   & $\pm$0.035   & $\pm$0.036   & $\pm$0.026   & $\pm$0.031   & $\pm$0.036   & $\pm$0.015   & $\pm$0.034   & $\pm$0.029   & $\pm$0.027   & $\pm$0.025   & $\pm$0.044   & $\pm$0.033   & \textbf{$\pm$0.007}   \\ \midrule
                                          & \textbf{BACC}                            & {0.500}   & {0.668} & {0.786} & {0.794} & {0.783} & {0.680}  & {0.615} & {0.764} & {0.650}  & {0.723} & {0.717} & {0.682} & {0.662} & {\textbf{0.943}} \\
                                          & \textbf{}                                & $\pm$0.001   & $\pm$0.035   & $\pm$0.036   & $\pm$0.028   & $\pm$0.029   & $\pm$0.035   & $\pm$0.014   & $\pm$0.033   & $\pm$0.029   & $\pm$0.028   & $\pm$0.022   & $\pm$0.047   & $\pm$0.033   & \textbf{$\pm$0.005}   \\
                                          & \textbf{MCC}                             & {0.000}     & {0.254} & {0.480}  & {0.488} & {0.466} & {0.296} & {0.195} & {0.471} & {0.229} & {0.340}  & {0.375} & {0.277} & {0.251} & {\textbf{0.885}} \\
\multirow{-2}{*}{\rotatebox[origin=c]{90}{\textbf{Spam not Spam}}}  & \textbf{}                                & $\pm$0.004   & $\pm$0.059   & $\pm$0.067   & $\pm$0.031   & $\pm$0.093   & $\pm$0.073   & $\pm$0.035   & $\pm$0.064   & $\pm$0.061   & $\pm$0.051   & $\pm$0.086   & $\pm$0.039   & $\pm$0.072   & \textbf{$\pm$0.007}   \\
\textbf{}                                 & \textbf{F1-Score}                        & {0.758} & {0.600}   & {0.744} & {0.779} & {0.748} & {0.592} & {0.463} & {0.783} & {0.594} & {0.684} & {0.640}  & {0.618} & {0.637} & {\textbf{0.943}} \\
                                          & \textbf{}                                & $\pm$0.002   & $\pm$0.067   & $\pm$0.103   & $\pm$0.043   & $\pm$0.108   & $\pm$0.082   & $\pm$0.097   & $\pm$0.103   & $\pm$0.096   & $\pm$0.089   & $\pm$0.074   & $\pm$0.051   & $\pm$0.118   & \textbf{$\pm$0.006}   \\
                                          & \textbf{Gmean}                           & {0.500}   & {0.668} & {0.786} & {0.794} & {0.783} & {0.680}  & {0.615} & {0.764} & {0.650}  & {0.723} & {0.717} & {0.682} & {0.662} & {\textbf{0.943}} \\
                                          & \textbf{}                                & $\pm$0.001   & $\pm$0.035   & $\pm$0.036   & $\pm$0.026   & $\pm$0.031   & $\pm$0.036   & $\pm$0.015   & $\pm$0.034   & $\pm$0.029   & $\pm$0.027   & $\pm$0.025   & $\pm$0.044   & $\pm$0.033   & \textbf{$\pm$0.007}   \\ \midrule
\multirow{8}{*}{\rotatebox[origin=c]{90}{\textbf{Spam SMS}}}                       & \textbf{BACC}                            & {0.500}   & {0.894} & {0.901} & {0.852} & {0.842} & {0.902} & {0.886} & {0.884} & {0.901} & {0.920}  & {0.845} & {0.901} & {0.885} & {\textbf{0.971}} \\
                                          &                                 & $\pm$0.000       & $\pm$0.013   & $\pm$0.025   & $\pm$0.050    & $\pm$0.163   & $\pm$0.028   & $\pm$0.009   & $\pm$0.032   & $\pm$0.010    & $\pm$0.008   & $\pm$0.051   & $\pm$0.020    & $\pm$0.017   & \textbf{$\pm$0.003}   \\
                                          & \textbf{MCC}                             & {0.000}     & {0.652} & {0.664} & {0.563} & {0.567} & {0.649} & {0.637} & {0.648} & {0.663} & {0.737} & {0.633} & {0.660}  & {0.611} & {\textbf{0.941}} \\
                                          &                                 & $\pm$0.000       & $\pm$0.062   & $\pm$0.042   & $\pm$0.124   & $\pm$0.262   & $\pm$0.073   & $\pm$0.034   & $\pm$0.088   & $\pm$0.025   & $\pm$0.038   & $\pm$0.069   & $\pm$0.077   & $\pm$0.048   & \textbf{$\pm$0.006}   \\
                                          & \textbf{F1-Score}                        & {0.804} & {0.896} & {0.904} & {0.843} & {0.777} & {0.892} & {0.895} & {0.895} & {0.904} & {0.930}  & {0.909} & {0.896} & {0.878} & {\textbf{0.971}} \\
                                          &                                & $\pm$0.000       & $\pm$0.034   & $\pm$0.012   & $\pm$0.088   & $\pm$0.292   & $\pm$0.030    & $\pm$0.016   & $\pm$0.039   & $\pm$0.011   & $\pm$0.013   & $\pm$0.018   & $\pm$0.045   & $\pm$0.027   & \textbf{$\pm$0.003}   \\
                                          & \textbf{Gmean}                           & {0.500}   & {0.652} & {0.762} & {0.710}  & {0.723} & {0.794} & {0.732} & {0.730}  & {0.730}  & {0.740}  & {0.642} & {0.664} & {0.721} & {\textbf{0.971}} \\
                                          &                                 & $\pm$0.000       & $\pm$0.094   & $\pm$0.097   & $\pm$0.155   & $\pm$0.192   & $\pm$0.122   & $\pm$0.103   & $\pm$0.099   & $\pm$0.044   & $\pm$0.119   & $\pm$0.054   & $\pm$0.125   & $\pm$0.050    & \textbf{$\pm$0.003}   \\ \midrule
\multirow{8}{*}{\rotatebox[origin=c]{90}{\textbf{Hate Speech}}}                      & \textbf{BACC}                            & {0.333} & {0.542} & {0.529} & {0.536} & {0.523} & {0.508} & {0.512} & {0.481} & {0.542} & {0.521} & {0.531} & {0.474} & {0.533} & {\textbf{0.749}} \\
                                          &                                 & $\pm$0.000       & $\pm$0.006   & $\pm$0.014   & $\pm$0.012   & $\pm$0.008   & $\pm$0.025   & $\pm$0.037   & $\pm$0.067   & $\pm$0.008   & $\pm$0.023   & $\pm$0.045   & $\pm$0.048   & $\pm$0.010    & \textbf{$\pm$0.006}   \\
                                          & \textbf{MCC}                             & {0.000}     & {0.438} & {0.406} & {0.416} & {0.393} & {0.326} & {0.347} & {0.321} & {0.436} & {0.398} & {0.339} & {0.280}  & {0.429} & {\textbf{0.639}} \\
                                          &                                 & $\pm$0.000       & $\pm$0.027   & $\pm$0.012   & $\pm$0.024   & $\pm$0.010    & $\pm$0.078   & $\pm$0.078   & $\pm$0.131   & $\pm$0.018   & $\pm$0.041   & $\pm$0.104   & $\pm$0.051   & $\pm$0.021   & \textbf{$\pm$0.01}    \\
                                          & \textbf{F1-Score}                        & {0.676} & {0.743} & {0.728} & {0.724} & {0.709} & {0.640}  & {0.661} & {0.579} & {0.740}  & {0.744} & {0.606} & {0.509} & {0.749} & {\textbf{0.762}} \\
                                          &                                 & $\pm$0.000       & $\pm$0.030    & $\pm$0.020    & $\pm$0.033   & $\pm$0.021   & $\pm$0.096   & $\pm$0.078   & $\pm$0.204   & $\pm$0.015   & $\pm$0.027   & $\pm$0.293   & $\pm$0.088   & $\pm$0.017   & \textbf{$\pm$0.007}   \\
                                          & \textbf{Gmean}                           & {0.471} & {0.674} & {0.662} & {0.668} & {0.658} & {0.637} & {0.642} & {0.615} & {0.675} & {0.652} & {0.648} & {0.606} & {0.666} & {\textbf{0.741}} \\
                                          & \textbf{}                                & $\pm$0.000       & $\pm$0.004   & $\pm$0.012   & $\pm$0.010    & $\pm$0.007   & $\pm$0.027   & $\pm$0.034   & $\pm$0.063   & $\pm$0.007   & $\pm$0.021   & $\pm$0.049   & $\pm$0.040    & $\pm$0.009   & \textbf{$\pm$0.006}   \\ \bottomrule
\end{tabular}}
\color{black}
\end{table}

\section{Statistical Analysis}
\label{supp_test}
We conducted the Friedman Rank Test to evaluate the performance differences among various state-of-the-art methods compared to our proposed approach on multiple image datasets. In Table~\ref{tab:friedman} we have obtained the results for the Friedman Rank test of our method with the state of the art method. The analysis was based on accuracy scores computed across five different random seeds for each method. The null hypothesis assumes that there is no significant difference among the Baseline, SMOTE, ROS, cDCGAN, GAMO, DGCMM, DeepSMOTE, TailCalibX, and our method, in terms of performance. In contrast, the alternative hypothesis posits that significant differences do exist among these approaches. For each dataset, we ranked all methods—including ours—based on their accuracy across the five seeds and calculated the corresponding test statistics. This allowed us to determine the $p$-values for each performance metric. The resulting $p$-values were consistently below 0.001, strongly indicating statistical significance and leading us to reject the null hypothesis in favor of the alternative.
\begin{table}[]
\caption{Friedman Rank Test Results for Our method versus the state-of-the-art methods based on the five random seeds}
\label{tab:friedman}
\scriptsize
\resizebox{\textwidth}{!}{%
\begin{tabular}{lcccccccc}
\toprule
 & \multicolumn{2}{c}{\textbf{BACC}} & \multicolumn{2}{c}{\textbf{MCC}} & \multicolumn{2}{c}{\textbf{F1-Score}} & \multicolumn{2}{c}{\textbf{Gmean}} \\ \cmidrule(l){2-3} \cmidrule(l){4-5} \cmidrule(l){6-7} \cmidrule(l){8-9} 
\multirow{-2}{*}{\textbf{Dataset}} & \multicolumn{1}{l}{\textbf{Statistics}} & \multicolumn{1}{l}{\textbf{$p$-Value}} & \multicolumn{1}{l}{\textbf{Statistics}} & \multicolumn{1}{l}{\textbf{$p$-Value}} & \multicolumn{1}{l}{\textbf{Statistics}} & \multicolumn{1}{l}{\textbf{$p$-Value}} & \multicolumn{1}{l}{\textbf{Statistics}} & \multicolumn{1}{l}{\textbf{$p$-Value}} \\ \midrule
\textbf{MNIST} & { 37.55} & { <.001} & { \text{36.80}} & { \text{<.001}} & { \text{37.55}} & { \text{<.001}} & { \text{36.91}} & { \text{<.001}} \\
\textbf{FMNIST} & { 38.56} & { <.001} & { \text{38.93}} & { \text{<.001}} & { \text{39.04}} & { \text{<.001}} & { \text{39.04}} & { \text{<.001}} \\
\textbf{CIFAR10} & { 32.28} & { <.001} & { \text{32.44}} & { \text{<.001}} & { \text{33.64}} & { \text{<.001}} & { \text{32.28}} & { \text{<.001}} \\
\textbf{SVHN} & { \text{38.29}} & { \text{<.001}} & { \text{38.29}} & { \text{<.001}} & { \text{38.08}} & { \text{<.001}} & { \text{38.29}} & { \text{<.001}} \\
\textbf{GTSRB} & { 37.12} & { <.001} & { \text{36.48}} & { \text{<.001}} & { \text{36.43}} & { \text{<.001}} & { \text{37.12}} & { \text{<.001}} \\
\textbf{STL10} & { 26.76} & { <.001} & { \text{26.74}} & { \text{<.001}} & { \text{27.26}} & { \text{<.001}} & { \text{26.76}} & { \text{<.001}} \\
\textbf{SUN$\textbf{367}_{\textbf{32}\times \textbf{32}}$} & { 35.25} & { <.001} & { \text{35.36}} & { \text{<.001}} & { \text{36.48}} & { \text{<.001}} & { \text{35.25}} & { \text{<.001}} \\
\textbf{SUN$\textbf{367}_{\textbf{64}\times \textbf{64}}$} & { 36.32} & { <.001} & { \text{35.52}} & { \text{<.001}} & { \text{35.52}} & { \text{<.001}} & { \text{36.43}} & { \text{<.001}} \\ \bottomrule
\end{tabular}
}
\end{table}

\section{Additional Ablation Study}
\label{supp:add_ablation}

\subsection{Evaluating Hyperparameter Impact Through Ablation Study}
\begin{figure}
    \centering
    \includegraphics[width=1.0\linewidth]{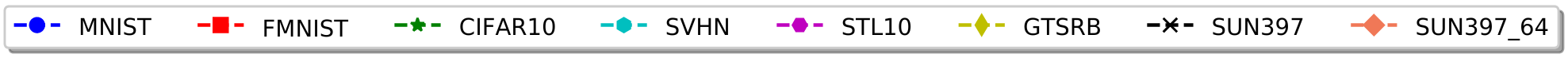}
    \includegraphics[width=1.0\linewidth]{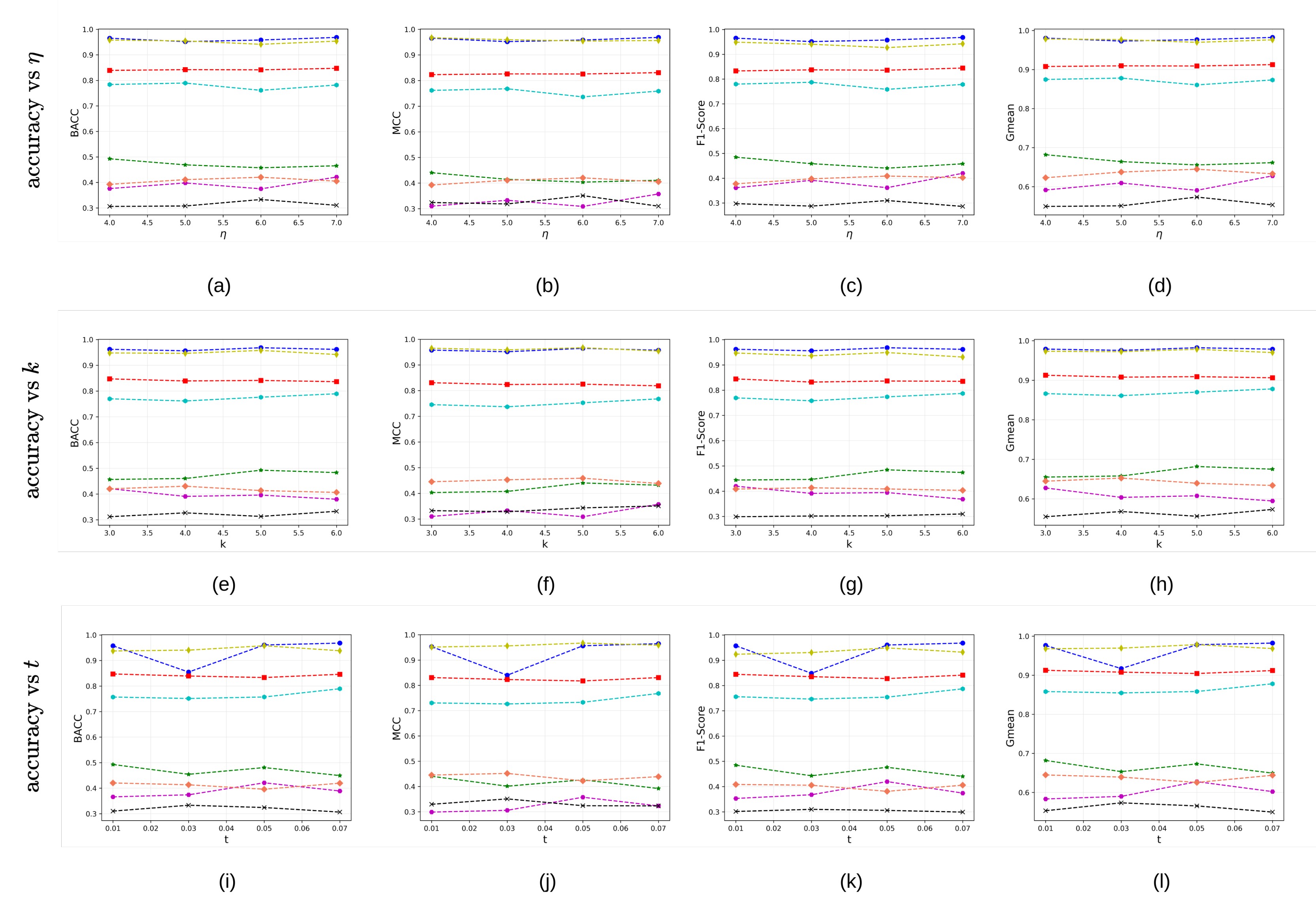}
    \caption{Figures depicting the performance of our method by varying the hyperparameters. a), e), i) depicting the balance accuracy corresponding to the hyperparameter $\eta,k \text{ and }t$ respectively. b), f), j) depicting the Mathews Correlation Coefficient corresponding to the hyperparameter $\eta,k \text{ and }t$ respectively. c), g), k) depicting the F1-Score corresponding to the hyperparameter $\eta,k \text{ and }t$ respectively. d), h), l) depicting the Gmean corresponding to the hyperparameter $\eta,k \text{ and }t$ respectively}
    \label{fig:ablation_all_metrics}
\end{figure}

This section reports a study on the individual impact of each hyperparameter—$k$, $\eta$, and $t$—on the overall performance of the proposed method. To achieve this, two hyperparameters are fixed at a time while varying the third. For each dataset and each performance metric, a scatter plot is generated corresponding to the varying hyperparameter, as shown in Figure~\ref{fig:ablation_all_metrics}.

In the first row, we plot the variation of $\eta$ across all metrics for the datasets MNIST, FMNIST, CIFAR10, SVHN, STL10, GTSRB, SUN397, and SUN397$_{64}$ (SUN397 with image size $64 \times 64$). We observe that for MNIST, FMNIST, and GTSRB, changes in $\eta$ have minimal effect on the accuracy. For STL10 and SVHN, accuracy initially decreases and then increases as $\eta$ increases. For SUN397 (both $32 \times 32$ and $64 \times 64$), accuracy initially increases and then decreases, while for CIFAR10, accuracy decreases monotonically as $\eta$ increases.

In the second row, we analyze the effect of varying $k$ while keeping other hyperparameters fixed. For MNIST, FMNIST, and GTSRB, accuracy remains largely unaffected by changes in $k$. For SVHN, accuracy decreases slightly and then increases to a peak value. For SUN397 ($64 \times 64$), accuracy decreases monotonically. In the case of CIFAR10, accuracy increases monotonically to an optimal point and then starts decreasing. For SUN397 ($32 \times 32$), we observe metric-specific trends: oscillations in BACC and GMean, a monotonically increasing trend in MCC, and stability in F1-Score. For STL10, accuracy decreases monotonically in BACC, F1-Score, and GMean, and shows oscillatory behavior in MCC as $k$ increases.

In the third row, we visualize the change in accuracy with respect to the hyperparameter $t$. For CIFAR10, STL10, and SUN397, accuracy increases and then decreases. For MNIST and SUN397 ($64 \times 64$), accuracy decreases and then increases. For CIFAR10, accuracy exhibits an oscillatory trend. For SVHN, accuracy increases monotonically, whereas for FMNIST, it remains nearly constant as $t$ increases.

\subsection{Analysis of existing calibration method vs RCS}
\label{cal_vs_rcs}
To check whether our distribution calibration performs better than another calibration method, we have compared the RCS method with the TailCalibX Calibration method~\cite{vigneswaran2021feature}. Other calibration techniques~\cite {vigneswaran2021feature} have been proposed earlier. Still, since they are similar to this method, we have compared them with this method, and the accuracy values are presented in Tables~\ref {tab:mnist_fmnist},~\ref{tab:cifar10_svhn_resnet},~\ref{tab:gtsrb_stl10}, and~\ref{tab:my-table_sun397} respectively. Our method surpasses the TailCalibX data Calibration by a large margin for all the datasets. We also observe that for GTSRB, STL10, and SUN397, TailCalibX performs better than some of the competitor methods, and for MNIST, FMNIST, CIFAR10, and SVHN, the performance of TailCalibX is worse. \textcolor{black}{Along with this, we have also compared our oversampling algorithm with the SMOTE algorithm with features extracted from the autoencoders with temperature values as 0.01, 0.03, 0.05, and 0.07. The results and discussions of this scenario are presented in Supplementary Section~\ref{ana_smote_rcs}}

\subsection{\textcolor{black}{Analysis of bias from the Majority Class}}
\color{black}
To visualize whether the bias from the majority class is not transferred to the minority class, we performed a TSNE plot of the majority class along with the samples generated in the minority and intermediate classes. We have also displayed the t-SNE plots of the majority class and the generated samples in the minority class, having the least number of elements. 
\begin{figure}
    \centering
    \hspace{-0.5cm}
\includegraphics[width=0.27\textwidth]{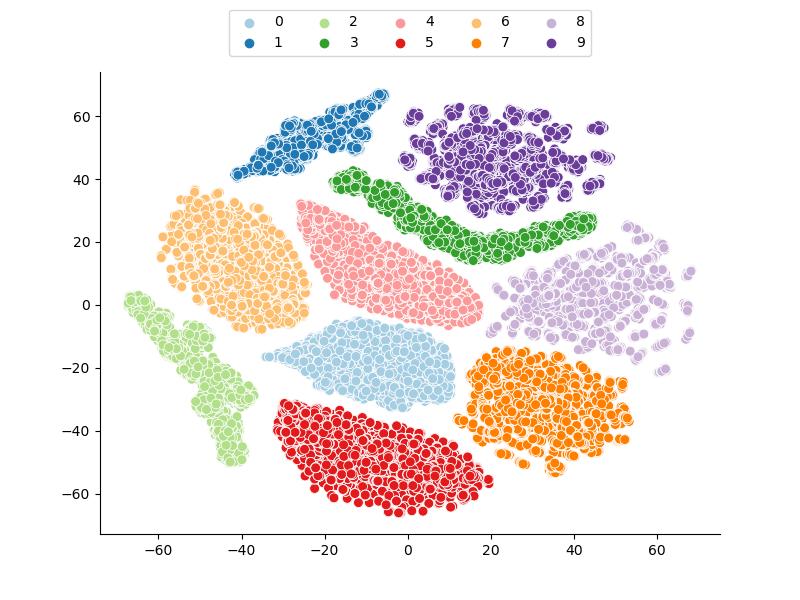} \hspace{-0.7cm}
\includegraphics[width=0.27\textwidth]{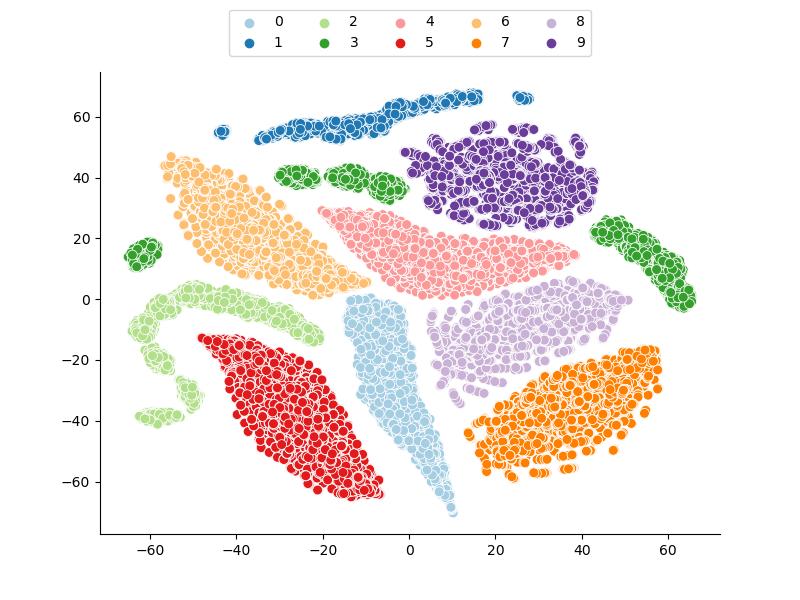} \hspace{-0.7cm}
\includegraphics[width=0.27\textwidth]{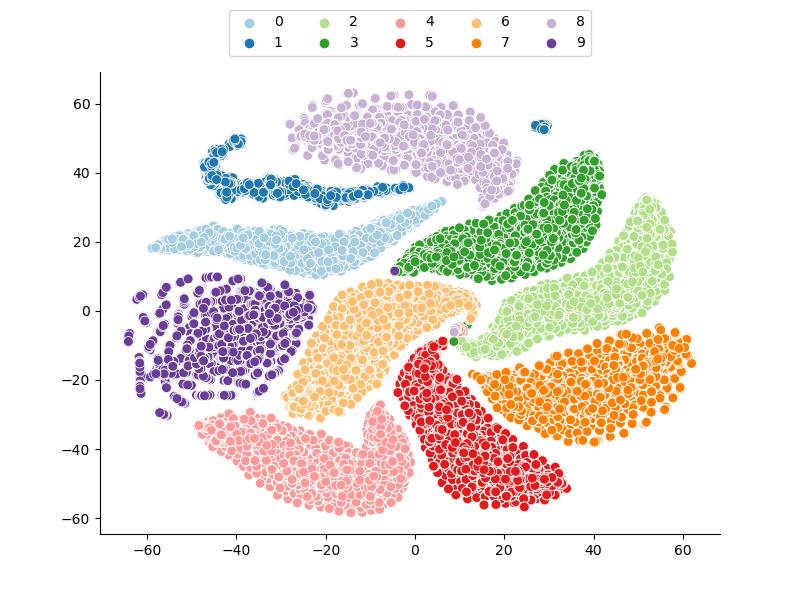} \hspace{-0.7cm}
\includegraphics[width=0.27\textwidth]{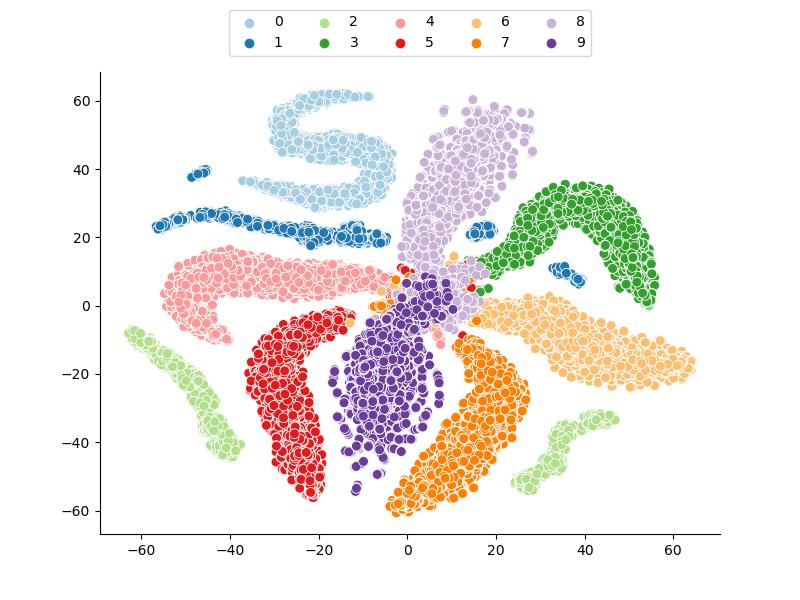}
    \caption{Illustrations of the majority class samples and the generated samples in the other classes for MNIST, FMNIST, CIFAR10, and SVHN (left to right)}
    \label{fig:maj_gen_all}
\end{figure}

We could observe in Figure~\ref{fig:maj_gen_all} for the above datasets, MNIST, FMNIST, CIFAR10, and SVHN, the majority class, i.e., the class 0 is disentangled even after sample generation. We observe closer boundaries between the intermediate and the minority classes, but we still observe a distinct boundary for the majority class and the generated samples in the other classes.
\begin{figure}
    \centering
    \hspace{-0.5cm}
\includegraphics[width=0.27\textwidth]{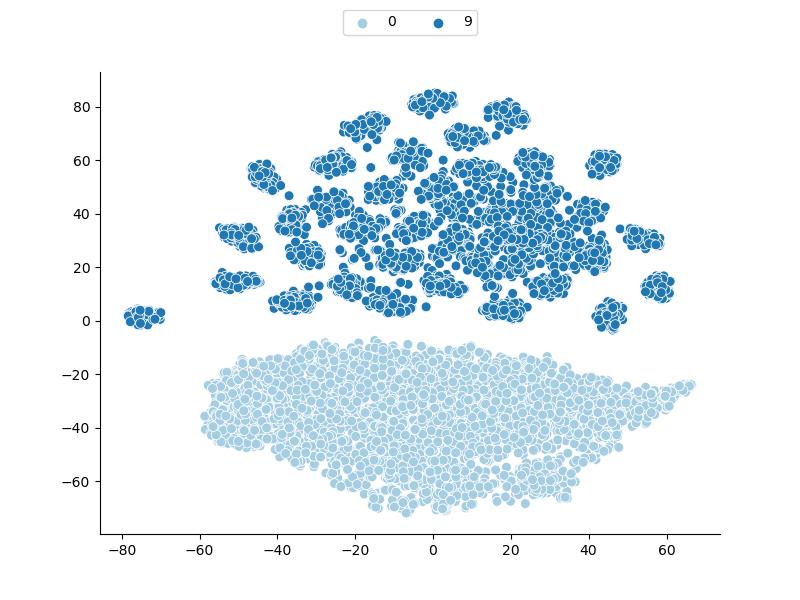} \hspace{-0.7cm}
\includegraphics[width=0.27\textwidth]{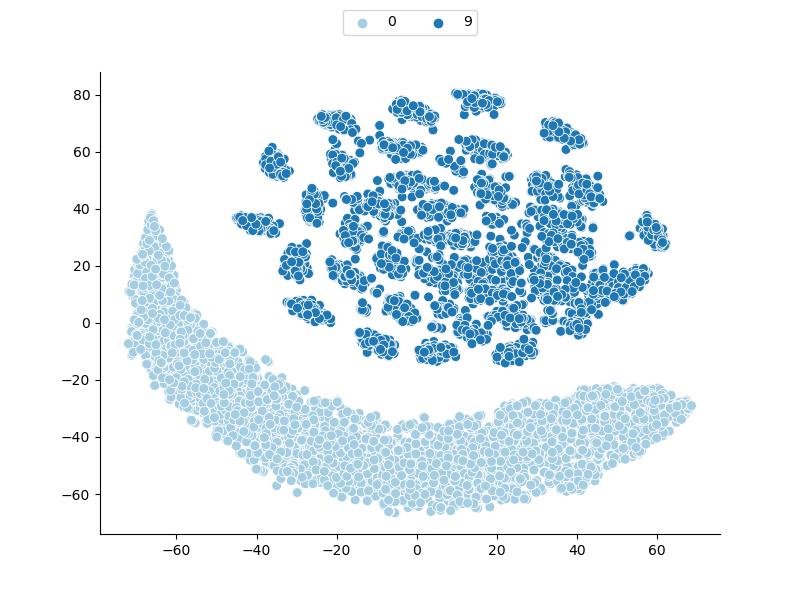} \hspace{-0.7cm}
\includegraphics[width=0.27\textwidth]{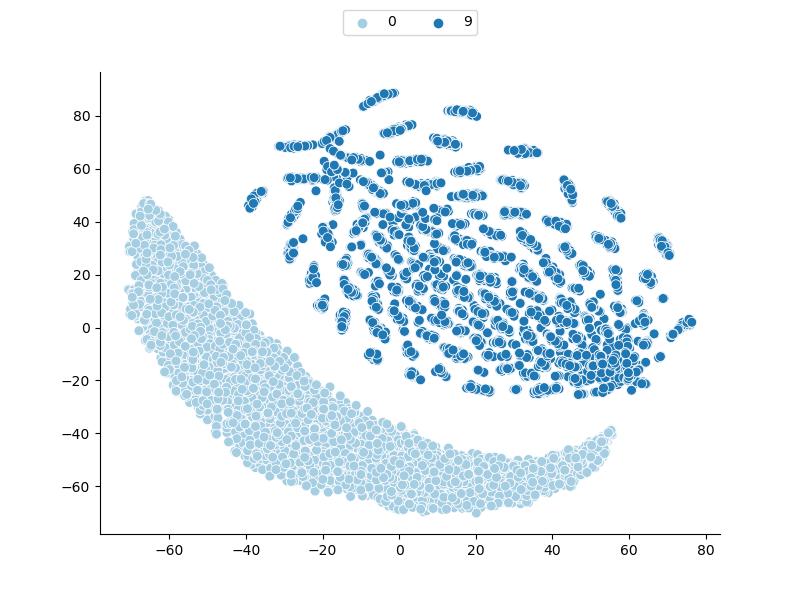} \hspace{-0.7cm}
\includegraphics[width=0.27\textwidth]{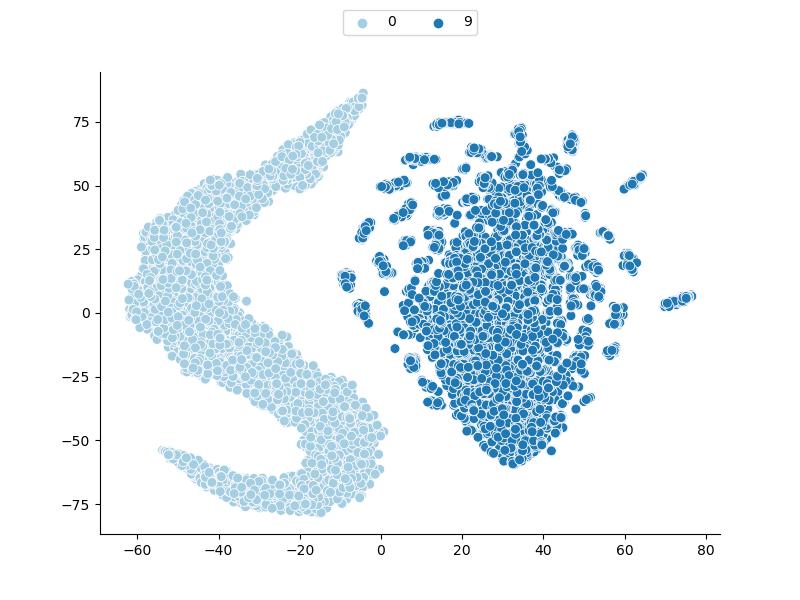}
    \caption{Illustrations of the majority class samples  and the generated samples in the Minority class (having least number of elements)  for MNIST, FMNIST, CIFAR10, and SVHN (left to right)}
    \label{fig:maj_gen_min}
\end{figure}

In Figure~\ref{fig:maj_gen_min}, we have illustrated the majority class and the samples generated in the minority class having the least number of elements. We observe that even the samples generated in the minority class, having the least number of elements, are disentangled from the majority class samples.
\subsection{\textcolor{black}{Analysis of using hard examples in majority class for sample generation}}
\color{black}
We have considered only the hard samples in the majority class to generate samples in the minority class. Hard samples refer to the samples where our classifier fails. In other words, hard samples in the majority class refer to the samples in the boundary regions of the majority class. Firstly, we obtained the hard sample using the Resnet32 classifier that had previously been trained on the imbalanced dataset. After receiving the hard samples, we consider those samples as the entire majority class samples and generate the samples in the minority class in the latent space of the autoencoder corresponding to the optimal temperature values and using the optimal values of $k$ and $\eta$. Using these generated samples and considering the hard samples in the majority class, we classify those samples using Resnet32 and reproduce the results in Table~\ref{tab:hard_samp_gen}. Let Hard samples based on Minority sample Generation be denoted as RCS-HMG. We observe that the samples generated using the hard samples are similar to RCS. This simply means boundary samples in the majority class are more effective than the interior point. This signifies our claim that our method is based on pattern coverage, and since the boundary samples are closer to the samples in the neighbouring class, they inherit the patterns of the boundary samples.
\begin{table}[h]
\centering
\caption{\textcolor{black}{Comparison of using hard examples in the majority class for sample generation in the minority class}}

\label{tab:hard_samp_gen}
\color{black}
\resizebox{0.7\textwidth}{!}{%
\begin{tabular}{cccccc}

\toprule
\textbf{Datasets} & \textbf{Methods}                & \textbf{BACC}                & \textbf{MCC}                 & \textbf{F1-Score}                  & \textbf{Gmean}                  \\ \midrule
\multirow{2}{*}{\textbf{MNIST}}    & \textbf{RCS-HMG} & 0.960$\pm$0.004  & 0.955$\pm$0.005 & 0.959$\pm$0.005 & 0.977$\pm$0.002 \\\vspace{1.5mm}
         & \textbf{RCS}                    & 0.962$\pm$0.002 & 0.958$\pm$0.002 & 0.962$\pm$0.002 & 0.979$\pm$0.001 \\ \hline

\multirow{2}{*}{\textbf{FMNIST}}   & \textbf{$\textbf{RCS-HMG}$} & 0.844$\pm$0.003 & 0.829$\pm$0.004 & 0.839$\pm$0.004 & 0.911$\pm$0.002 \\\vspace{1.5mm}
\textbf{}         & \textbf{RCS}                    & 0.845$\pm$0.004 & 0.829$\pm$0.004 & 0.840$\pm$0.006  & 0.911$\pm$0.002 \\ \midrule

\multirow{2}{*}{\textbf{CIFAR10}}  & \textbf{\textbf{RCS-HMG}} & 0.443$\pm$0.014 & 0.384$\pm$0.014 & 0.438$\pm$0.015 & 0.645$\pm$0.010  \\\vspace{1.5mm}
\textbf{}         & \textbf{RCS}                    & 0.453$\pm$0.058 & 0.397$\pm$0.063 & 0.438$\pm$0.068 & 0.651$\pm$0.046 \\ \hline

\multirow{2}{*}{\textbf{SVHN}}     & \textbf{\textbf{RCS-HMG}} & 0.778$\pm$0.005 & 0.755$\pm$0.005 & 0.775$\pm$0.005 & 0.871$\pm$0.003 \\
\textbf{}         & \textbf{RCS}                    & 0.777$\pm$0.008 & 0.755$\pm$0.009 & 0.774$\pm$0.010  & 0.871$\pm$0.005 \\ \bottomrule
\end{tabular}}
\end{table}
\color{black}

\subsection{Analysis of SMOTE vs RCS}
\label{ana_smote_rcs}
In order to show the effectiveness of RCS over SMOTE using our Autoencoder framework, we have varied the hyperparameter $t$ to obtain the latent feature representation and perform the oversampling with the SMOTE algorithm, and compared the RCS method with the optimal hyperparameter that we have discussed above.

In Table~\ref{tab:smote_supcon_ablation}, we denote ATNSM01 as our autoencoder trained with the temperature parameter 0.01, followed by the oversampling with the SMOTE algorithm, ATNSM03, ATNSM05, and ATNSM07 as the autoencoder trained with temperature parameter values 0.03, 0.05, and 0.07, respectively. We use the ResNet32 classifier on these oversampled images and observed in Table~\ref{tab:smote_supcon_ablation} that our data calibration method outperforms ATNSM01, ATNSM03, ATNSM05, and ATNSM07. These experiments observe that the highest improvement is for the STL10 dataset, with an improvement of 4.6 \%.

\begin{table}[!ht]
\caption{Results comparing our calibration method with oversampling with smote by considering our autoencoder network.}
\label{tab:smote_supcon_ablation}
\scriptsize
\resizebox{\textwidth}{!}{%
\begin{tabular}{ccccccc}
\toprule
\textbf{Datasets} & \textbf{Metrics} & \textbf{ATNSM01} & \textbf{ATNSM03} & \textbf{ATNSM05} & \textbf{ATNSM07} & \textbf{RCS} \\ \hline
\multirow{4}{*}{\textbf{MNIST$_{\textbf{28}\times\textbf{28}}$}} & \textbf{BACC} & 0.952$\pm$0.003 & 0.870$\pm$0.022 & 0.949$\pm$0.005 & 0.946$\pm$0.002 & \textbf{0.962$\pm$0.002} \\
 & \textbf{MCC} & 0.946$\pm$0.003 & 0.856$\pm$0.023 & 0.943$\pm$0.006 & 0.940$\pm$0.003 & \textbf{0.958$\pm$0.002} \\
 & \textbf{F1-Score} & 0.951$\pm$0.003 & 0.864$\pm$0.024 & 0.948$\pm$0.006 & 0.945$\pm$0.002 & \textbf{0.962$\pm$0.002} \\
 & \textbf{Gmean} & 0.973$\pm$0.002 & 0.926$\pm$0.013 & 0.971$\pm$0.003 & 0.970$\pm$0.001 & \textbf{0.979$\pm$0.001} \vspace{1.5mm} \\ 
 
\multirow{4}{*}{\textbf{FMNIST$_{\textbf{28}\times\textbf{28}}$}} & \textbf{BACC} & 0.811$\pm$0.005 & 0.812$\pm$0.012 & 0.811$\pm$0.007 & 0.811$\pm$0.005 & \textbf{0.845$\pm$0.004} \\
 & \textbf{MCC} & 0.794$\pm$0.005 & 0.795$\pm$0.012 & 0.793$\pm$0.007 & 0.794$\pm$0.005 & \textbf{0.829$\pm$0.004} \\
 & \textbf{F1-Score} & 0.801$\pm$0.005 & 0.804$\pm$0.013 & 0.803$\pm$0.009 & 0.804$\pm$0.007 & \textbf{0.840$\pm$0.006} \\
 & \textbf{Gmean} & 0.891$\pm$0.003 & 0.892$\pm$0.007 & 0.891$\pm$0.004 & 0.891$\pm$0.003 & \textbf{0.911$\pm$0.002} \vspace{1.5mm} \\ 
\multirow{4}{*}{\textbf{CIFAR10$_{\textbf{32}\times\textbf{32}}$}} & \textbf{BACC}  & 0.420$\pm$0.012                          & 0.419$\pm$0.009                         & 0.410$\pm$0.006                          & 0.393$\pm$0.006                         & \textbf{0.472$\pm$0.011}                         \\
 & \textbf{MCC} & 0.364$\pm$0.013 & 0.360$\pm$0.01   & 0.353$\pm$0.005                         & 0.334$\pm$0.007                         & \textbf{0.418$\pm$0.012}                         \\
 & \textbf{F1-Score} & 0.390$\pm$0.015  & 0.403$\pm$0.012 & 0.384$\pm$0.013                         & 0.364$\pm$0.009                         & \textbf{0.467$\pm$0.011}                         \\
 & \textbf{Gmean} & 0.627$\pm$0.009 & 0.626$\pm$0.007 & 0.619$\pm$0.005                         & 0.605$\pm$0.005                         & \textbf{0.667$\pm$0.008}                   \vspace{1.5mm}      \\ 
\multirow{4}{*}{\textbf{SVHN$_{\textbf{32}\times\textbf{32}}$}} & \textbf{BACC} & 0.703$\pm$0.010  & 0.705$\pm$0.008 & 0.703$\pm$0.018                         & 0.704$\pm$0.011                         & \textbf{0.776$\pm$0.011}                         \\
 & \textbf{MCC} & 0.677$\pm$0.010  & 0.678$\pm$0.009 & 0.676$\pm$0.020                          & 0.677$\pm$0.012                         & \textbf{0.753$\pm$0.012}                         \\
 & \textbf{F1-Score} & 0.689$\pm$0.014 & 0.694$\pm$0.009 & 0.692$\pm$0.021                         & 0.693$\pm$0.014                         & \textbf{0.773$\pm$0.012}                         \\
 & \textbf{Gmean}  & 0.825$\pm$0.006                         & 0.826$\pm$0.005                         & 0.825$\pm$0.011                         & 0.825$\pm$0.007                         & \textbf{0.870$\pm$0.007}  \vspace{1.5mm} \\ 

\multirow{4}{*}{\textbf{GTSRB$_{\textbf{32}\times\textbf{32}}$}} & \textbf{BACC} & 0.931$\pm$0.012 & 0.934$\pm$0.008 & 0.939$\pm$0.008 & 0.936$\pm$0.011 & \textbf{0.955$\pm$0.006} \\
 & \textbf{MCC} & 0.957$\pm$0.008 & 0.961$\pm$0.004 & 0.961$\pm$0.005 & 0.962$\pm$0.006 & \textbf{0.965$\pm$0.004} \\
 & \textbf{F1-Score} & 0.934$\pm$0.013 & 0.939$\pm$0.007 & 0.941$\pm$0.008 & 0.940$\pm$0.009 & \textbf{0.948$\pm$0.007} \\
 & \textbf{Gmean} & 0.965$\pm$0.006 & 0.966$\pm$0.004 & 0.969$\pm$0.004 & 0.967$\pm$0.006 & \textbf{0.977$\pm$0.003} \vspace{1.5mm} \\ 
\multirow{4}{*}{\textbf{STL10$_{\textbf{32}\times\textbf{32}}$}} & \textbf{BACC} & 0.367$\pm$0.014 & 0.374$\pm$0.013 & 0.371$\pm$0.014 & 0.368$\pm$0.013 & \textbf{0.420$\pm$0.009} \\
 & \textbf{MCC} & 0.301$\pm$0.016 & 0.308$\pm$0.015 & 0.305$\pm$0.016 & 0.302$\pm$0.014 & \textbf{0.360$\pm$0.010} \\
 & \textbf{F1-Score} & 0.345$\pm$0.008 & 0.352$\pm$0.014 & 0.354$\pm$0.014 & 0.348$\pm$0.015 & \textbf{0.401$\pm$0.008} \\
 & \textbf{Gmean} & 0.584$\pm$0.011 & 0.590$\pm$0.011 & 0.588$\pm$0.012 & 0.585$\pm$0.010 & \textbf{0.627$\pm$0.007} \vspace{1.5mm} \\ 
\multirow{4}{*}{\textbf{SUN397$_{\textbf{32}\times\textbf{32}}$}} & \textbf{BACC} & 0.28$\pm$0.006 & 0.281$\pm$0.007 & 0.285$\pm$0.011 & 0.278$\pm$0.004 & \textbf{0.318$\pm$0.009} \\
 & \textbf{MCC} & 0.289$\pm$0.008 & 0.293$\pm$0.006 & 0.295$\pm$0.015 & 0.293$\pm$0.007 & \textbf{0.330$\pm$0.014} \\
 & \textbf{F1-Score} & 0.264$\pm$0.008 & 0.267$\pm$0.006 & 0.277$\pm$0.012 & 0.264$\pm$0.007 & \textbf{0.303$\pm$0.013} \\
 & \textbf{Gmean} & 0.525$\pm$0.006 & 0.527$\pm$0.006 & 0.530$\pm$0.010 & 0.523$\pm$0.003 & \textbf{0.560$\pm$0.008} \vspace{1.5mm} \\ 
\multirow{4}{*}{\textbf{SUN397$_{\textbf{64}\times\textbf{64}}$}} & \textbf{BACC} & 0.378$\pm$0.011 & 0.372$\pm$0.009 & 0.377$\pm$0.003 & 0.371$\pm$0.009 & \textbf{0.406$\pm$0.007} \\
 & \textbf{MCC} & 0.404$\pm$0.008 & 0.398$\pm$0.017 & 0.402$\pm$0.007 & 0.396$\pm$0.014 & \textbf{0.439$\pm$0.009} \\
 & \textbf{F1-Score} & 0.369$\pm$0.008 & 0.367$\pm$0.010 & 0.370$\pm$0.005 & 0.361$\pm$0.009 & \textbf{0.397$\pm$0.004} \\
 & \textbf{Gmean} & 0.611$\pm$0.009 & 0.606$\pm$0.008 & 0.610$\pm$0.003 & 0.605$\pm$0.007 & \textbf{0.633$\pm$0.006} \\ \bottomrule
\end{tabular}
}
\end{table}

\subsection{\textcolor{black}{Analysis of using Center loss while training Autoencoder}}
\color{black}
We have used a regularized loss consisting of the supervised contrastive loss, the classifier loss, and the mean square error loss. The supervised contrastive loss not only handles the intra-class compactness but also handles inter-class separation. On the other hand, centre loss only handles intra-class compactness but does not consider the inter-class separability. In contrast, the classifier guidance loss guides latent features to the correct classes and prevents misclassification. 
Along with the Reconstruction loss $(\mathcal{L}_{AC})$ and the Classifier Guidance
loss $(\mathcal{L}_{CG})$, we use both the Center Loss $(\mathcal{L}_C)$ and the Supervised Contrastive loss $(\mathcal{L}_{CS})$ as provided in Table~\ref{tab:supcon_center}.
After that, we generated samples in the latent space of our trained autoencoder and the optimal values of the hyperparameters $k$ and $\eta$.
We observe distinctive improvements in our method than using the center loss. In terms of balanced accuracy for SVHN, we observe around 27$\%$, for CIFAR10, we observe 10$\%$ improvement, for FMNIST and MNIST, we have improvements of $4\%$ and $0.6\%$ values respectively.
\color{black}
\begin{table}
\centering
\caption{\textcolor{black}{Comparison of performance while training Autoencoder using Combination of Loses with $\mathcal{L}_{CS}$ and $\mathcal{L}_{C}$.}}
\label{tab:supcon_center}
\color{black}
\resizebox{0.7\textwidth}{!}{%
\begin{tabular}{cccccccc}
\toprule
\textbf{Datasets}      & \textbf{$\mathcal{L}_{CS}$}     & \textbf{$\mathcal{L}_{C}$}     & \textbf{BACC}                & \textbf{MCC}                 & \textbf{F1-Score}                  & \textbf{Gmean}                  \\ \midrule
\multirow{2}{*}{\textbf{MNIST}}    & $\times$ & $\checkmark$ & 0.956$\pm$0.002 & 0.951$\pm$0.002 & 0.956$\pm$0.002 & 0.975$\pm$0.001 \\\vspace{1.5mm}
\textbf{}          & $\checkmark$ & $\times$ & 0.962$\pm$0.002 & 0.958$\pm$0.002 & 0.962$\pm$0.002 & 0.979$\pm$0.001 \\
\multirow{2}{*}{\textbf{FMNIST}}  & $\times$ & $\checkmark$ &  0.805$\pm$0.003 & 0.788$\pm$0.003 & 0.794$\pm$0.004 & 0.888$\pm$0.002 \\\vspace{1.5mm}
\textbf{}          & $\checkmark$ & $\times$ & 0.845$\pm$0.004 & 0.829$\pm$0.004 & 0.840$\pm$0.006  & 0.911$\pm$0.002 \\
\multirow{2}{*}{\textbf{CIFAR10}}   & $\times$ & $\checkmark$ & 0.351$\pm$0.012 & 0.286$\pm$0.014 & 0.324$\pm$0.023 & 0.571$\pm$0.011 \\\vspace{1.5mm}
\textbf{}          & $\checkmark$ & $\times$ & 0.453$\pm$0.058 & 0.397$\pm$0.063 & 0.438$\pm$0.068 & 0.651$\pm$0.046 \\
\multirow{2}{*}{\textbf{SVHN}}      & $\times$ & $\checkmark$ & 0.504$\pm$0.039 & 0.456$\pm$0.043 & 0.478$\pm$0.043 & 0.690$\pm$0.028  \\
\textbf{}          & $\checkmark$ & $\times$ & 0.777$\pm$0.008 & 0.755$\pm$0.009 & 0.774$\pm$0.010  & 0.871$\pm$0.005 \\ \bottomrule
\end{tabular}}
\end{table}

\subsection{Comparative Analysis of Autoencoder Variants in Our Framework}
To evaluate the role of the encoder-decoder architecture in our framework, we conducted an ablation study by replacing the backbone Autoencoder with alternative designs while keeping the rest of the oversampling pipeline fixed. Specifically, we compare three variants: (i) \textbf{Variational Autoencoder (VAE)}, (ii) \textbf{Contrastive Autoencoder}, and (iii) our proposed \textbf{Autoencoder with Regularized Loss} as described in Equation~\ref{combined_loss}. In all cases, the encoder was used to extract latent representations of the input data, on which our RCS oversampling algorithm was subsequently applied to generate synthetic data. The augmented dataset was then used to train a downstream classifier, and the performance was evaluated across multiple metrics.

For the \emph{VAE}~\cite{kingma2019introduction}, we trained the model using the standard variational loss comprising a reconstruction term and a Kullback–Leibler (KL) divergence, along with our latent space regularization terms $\mathcal{L}_{CG}$ (Equation \ref{CG}) and $\mathcal{L}_{CS}$ (Equation \ref{CS}), to ensure consistency with our framework. After training, latent vectors were extracted and oversampled using RCS.

The \emph{Contrastive Autoencoder} was trained using a contrastive loss~\cite{chen2020simple} to encourage the learning of robust and invariant representations. The model received five augmented views of each input image during training, promoting alignment of similar samples and separation of dissimilar ones in the latent space. Once trained, we extracted latent vectors from the encoder and applied RCS-based oversampling.

Finally, we evaluated our complete framework using the \emph{Autoencoder with Regularized Loss (Equation~\ref{combined_loss})}. This architecture forms the final version used throughout our experiments and results.

Table~\ref{tab:autoencoder_train} presents a comprehensive comparison of classification performance after oversampling using latent vectors extracted from each autoencoder variant. The results in Table~\ref{tab:autoencoder_train} clearly demonstrate the superiority of our proposed Autoencoder with Regularized Loss across a diverse set of datasets and evaluation metrics. On simpler datasets such as MNIST and FMNIST, all methods yield reasonably high scores; however, our model consistently outperforms both the Variational and Contrastive Autoencoders, indicating more robust latent representations and more effective oversampling in even moderately challenging scenarios. The benefits of our method are particularly pronounced in complex, high-variance datasets such as CIFAR10, STL10, and SUN397. For instance, on CIFAR10, our method improves Balanced Accuracy by $2.1$\% over the Variational Autoencoder and around $7.0$\% over the Contrastive Autoencoder. Similarly, for SUN397 (32$\times$32), our method achieves a G-mean of $0.560$, compared to $0.551$ and $0.484$ with the Variational and Contrastive Autoencoders respectively. Interestingly, while the Variational Autoencoder performs well on certain datasets like GTSRB, it underperforms on more heterogeneous datasets, likely due to the entanglement introduced by KL regularization and lack of contrastive or discriminative supervision. Contrastive Autoencoders, although promoting invariance through augmented views, fall short due to their unsupervised nature and inability to directly preserve class semantics in the latent space. Our proposed autoencoder, by incorporating supervised constraints and optimizing for class preservation in the latent, bridges this gap effectively. In summary, these empirical findings reinforce the importance of latent space regularization and supervised disentanglement in facilitating meaningful synthetic data generation. When combined with our RCS oversampling algorithm, this leads to substantial and consistent gains in classification performance, particularly in the presence of severe class imbalance and distributional complexity.
\color{black}
\begin{table}[b]
\caption{\textcolor{black}{Comparison of classification performance after RCS oversampling using latent features extracted from different autoencoder architectures}}
\label{tab:autoencoder_train}
\resizebox{\columnwidth}{!}{%
\color{black}
\begin{tabular}{ccccccccccccc}
\toprule
 \textbf{Backbone} $\rightarrow$    & \multicolumn{4}{c}{\textbf{Contrastive Autoencoder~\cite{chen2020simple}}}                                                                                                              & \multicolumn{4}{c}{\textbf{Variational Autoencoder~\cite{kingma2019introduction}}}                                                                                                               & \multicolumn{4}{c}{\textbf{Vanilla Autoencoder}}                                                                                                                           \\ \cmidrule(l){1-1} \cmidrule(l){2-5} \cmidrule(l){6-9} \cmidrule(l){10-13}
\textbf{Data} $\downarrow$ & \textbf{BACC} & \textbf{MCC} & \textbf{F1-Score} & \textbf{Gmean} & \textbf{BACC} & \textbf{MCC}  & \textbf{F1-Score} & \textbf{Gmean} & \textbf{BACC} & \textbf{MCC}  & \textbf{F1-Score} & \textbf{Gmean} \\ \cmidrule(l){1-1} \cmidrule(l){2-5} \cmidrule(l){6-9} \cmidrule(l){10-13}
\textbf{MNIST}                  & 0.886$\pm$0.014          & 0.874$\pm$0.015         & 0.882$\pm$0.016              & 0.935$\pm$0.008           & 0.958$\pm$0.004          & 0.953$\pm$0.005          & 0.957$\pm$0.004              & 0.976$\pm$0.002           & \textbf{0.962$\pm$0.002} & \textbf{0.958$\pm$0.002} & \textbf{0.962$\pm$0.002}     & \textbf{0.979$\pm$0.001}  \\
\textbf{FMNIST}                 & 0.757$\pm$0.018          & 0.735$\pm$0.020         & 0.739$\pm$0.015              & 0.858$\pm$0.011           & 0.838$\pm$0.004          & 0.821$\pm$0.004          & 0.833$\pm$0.004              & 0.907$\pm$0.002           & \textbf{0.845$\pm$0.004} & \textbf{0.829$\pm$0.004} & \textbf{0.840$\pm$0.006}     & \textbf{0.911$\pm$0.002}  \\
\textbf{CIFAR10}                & 0.383$\pm$0.016          & 0.323$\pm$0.017         & 0.348$\pm$0.018              & 0.597$\pm$0.013           & 0.432$\pm$0.015          & 0.374$\pm$0.016          & 0.415$\pm$0.019              & 0.636$\pm$0.012           & \textbf{0.453$\pm$0.058} & \textbf{0.397$\pm$0.063} & \textbf{0.438$\pm$0.068}     & \textbf{0.651$\pm$0.046}  \\
\textbf{SVHN}                   & 0.686$\pm$0.015          & 0.656$\pm$0.015         & 0.674$\pm$0.019              & 0.814$\pm$0.009           & 0.758$\pm$0.007          & 0.734$\pm$0.007          & 0.752$\pm$0.008              & 0.859$\pm$0.004           & \textbf{0.776$\pm$0.011} & \textbf{0.753$\pm$0.012} & \textbf{0.773$\pm$0.012}     & \textbf{0.870$\pm$0.007}  \\
\textbf{GTSRB}                  & 0.822$\pm$0.011          & 0.871$\pm$0.008         & 0.810$\pm$0.017              & 0.905$\pm$0.006           & \textbf{0.961$\pm$0.007} & \textbf{0.970$\pm$0.004} & \textbf{0.952$\pm$0.008}     & \textbf{0.980$\pm$0.003}  & 0.955$\pm$0.006          & 0.965$\pm$0.004          & 0.948$\pm$0.007              & 0.977$\pm$0.003           \\
\textbf{STL10}                  & 0.350$\pm$0.019          & 0.284$\pm$0.021         & 0.318$\pm$0.022              & 0.570$\pm$0.016           & 0.362$\pm$0.009          & 0.295$\pm$0.011          & 0.351$\pm$0.011              & 0.580$\pm$0.008           & \textbf{0.420$\pm$0.009} & \textbf{0.360$\pm$0.010} & \textbf{0.401$\pm$0.008}     & \textbf{0.627$\pm$0.007}  \\
\textbf{SUN397}$_{\textbf{32}\times \textbf{32}}$                 & 0.238$\pm$0.010          & 0.292$\pm$0.014         & 0.226$\pm$0.007              & 0.484$\pm$0.010           & 0.308$\pm$0.007          & 0.330$\pm$0.007          & 0.295$\pm$0.005              & 0.551$\pm$0.006           & \textbf{0.318$\pm$0.009} & \textbf{0.330$\pm$0.014} & \textbf{0.303$\pm$0.013}     & \textbf{0.560$\pm$0.008}  \\
\textbf{SUN397}$_{\textbf{64}\times \textbf{64}}$             & 0.484$\pm$0.010          & 0.344$\pm$0.007         & 0.401$\pm$0.016              & 0.338$\pm$0.011           & 0.583$\pm$0.006          & 0.394$\pm$0.012          & 0.423$\pm$0.014              & 0.386$\pm$0.010           & 0.624$\pm$0.010          & \textbf{0.406$\pm$0.007} & \textbf{0.439$\pm$0.009}     & \textbf{0.397$\pm$0.004}  \\ \bottomrule
\end{tabular}}
\end{table}

\subsection{\textcolor{black}{Analysis on the scalability of our method}}
\color{black}
The datasets in our evaluation are arranged in ascending order of training set size, from STL10 (2,370 samples) to GTSRB (39,209 samples). Our method consistently outperforms other state-of-the-art approaches across datasets, as detailed below.

For small datasets such as STL10 (2,370 samples), RCS achieves a performance improvement of 5.04$\%$, which is significantly higher than most methods (e.g., CDCGAN: 7.06$\%$, GAMO: 23.76$\%$). This demonstrates the ability of RCS to handle small datasets effectively without overfitting. For medium datasets such as CIFAR10 and SVHN (~10,280 samples), RCS maintains competitive performance with improvements of $4.76\%$ (CIFAR10) and $8.55\%$ (SVHN). For larger datasets such as SUN$397_{64\times 64}$ ($14,258$ samples), RCS demonstrates robustness with an improvement of 3.23$\%$, outperforming GAMO (12.28$\%$) and maintaining competitive efficiency compared to DGCMM (2.94$\%$).
For even large-sized GTSRB datasets (39,209 samples), RCS achieves a performance improvement of 8.32$\%$, showing its robustness in handling datasets with many classes and substantial training data. This outperforms methods like DeepSMOTE (7.21$\%$) and CDCGAN (8.01$\%$), further highlighting the efficiency of RCS in such scenarios.


Apart from these, we have considered the CINIC10 dataset having 90000 training images and validation images with 9000 and 1000 samples in each class, respectively. We have created an imbalanced dataset with samples in each class as 9000, 5395, 3234, 1938, 1162, 696, 417, 250, 150, and 90 and an imbalanced ratio of 100. The total number of samples is 22182 which is greater than SUN397 and less than GTSRB. Table~\ref{tab:cinic10_results} below illustrates the results of our method on the CINIC10 dataset; we observe that even this dataset with a higher imbalanced ratio outperforms CDCGAN, DeepSMOTE, and DGCMM. We have also performed an execution time analysis on datasets with varying dataset sizes and the execution time of various methods on our datasets as provided in Table~\ref{tab:my-table_scalability}. RCS shows competitive retraining times for datasets with smaller samples, such as STL10 (2,370 samples) and MNIST (9,000 samples). For instance, on STL10, RCS takes 7.13 mins in total, slightly more than DeepSMOTE (6.96 mins) and cDCGAN (5.93 mins), but significantly less than DGCMM, which takes an enormous 392 mins. This pattern is consistent for MNIST and FMNIST, where RCS takes marginally longer than DeepSMOTE and cDCGAN but remains far more efficient than DGCMM. The slightly higher training time of RCS compared to cDCGAN and DeepSMOTE is due to its sampling strategy, which performs calibration-based data augmentation in the minority class rather than simply applying predefined oversampling techniques. However, this additional cost is minimal for smaller datasets and does not pose scalability concerns at this level.
\color{black}
\begin{table}[!ht]
\caption{\textcolor{black}{Computational time analysis of various methods on different datasets}}
\label{tab:my-table_scalability}

\color{black}
\resizebox{\textwidth}{!}{%
\begin{tabular}{lccccccccccccccc}
\toprule
\textbf{Datasets}   & \multicolumn{1}{l}{\textbf{\begin{tabular}[c]{@{}l@{}}No. of\\ Classes\end{tabular}}} & \multicolumn{1}{l}{\textbf{\begin{tabular}[c]{@{}l@{}}No. of\\ datapoints\end{tabular}}} & \multicolumn{1}{l}{\textbf{\begin{tabular}[c]{@{}l@{}}Training\\ cDCGAN\end{tabular}}} & \multicolumn{1}{l}{\textbf{\begin{tabular}[c]{@{}l@{}l@{}}Sample\\ Generation\\ cDCGAN\end{tabular}}} & \multicolumn{1}{l}{\textbf{\begin{tabular}[c]{@{}l@{}}cDCGAN\\ Classify\end{tabular}}} & \multicolumn{1}{l}{\textbf{cDCGAN}} & \multicolumn{1}{l}{\textbf{\begin{tabular}[c]{@{}l@{}}Training\\ DeepSMOTE\end{tabular}}} & \multicolumn{1}{l}{\textbf{\begin{tabular}[c]{@{}l@{}l@{}}Sample\\ Generation\\ DeepSMOTE\end{tabular}}} & \multicolumn{1}{l}{\textbf{\begin{tabular}[c]{@{}l@{}}DeepSMOTE\\ Classify\end{tabular}}} & \multicolumn{1}{l}{\textbf{DeepSMOTE}} & \multicolumn{1}{l}{\textbf{DGCMM}} & \multicolumn{1}{l}{\textbf{\begin{tabular}[c]{@{}l@{}}Training\\ RCS\end{tabular}}} & \multicolumn{1}{l}{\textbf{\begin{tabular}[c]{@{}l@{}l@{}}Sample\\ Generation\\ RCS\end{tabular}}} & \multicolumn{1}{l}{\textbf{\begin{tabular}[c]{@{}l@{}}RCS\\ Classify\end{tabular}}} & \multicolumn{1}{l}{\textbf{RCS}} \\ \midrule
\textbf{STL10}      & 10                                          & 2370                                           & 1.8                                          & 0.03                                                  & 4.1                                          & 5.93                                & 2.8                                             & 0.06                                                     & 4.1                                             & 6.96                                   & 392                                & 2.8                                       & 0.23                                               & 4.1                                       & \textbf{7.13}                    \\
\textbf{MNIST}      & 10                                          & 9000                                           & 2.4                                          & 0.19                                                  & 12                                           & 14.59                               & 4                                               & 1.21                                                     & 12                                              & 17.21                                  & 138.37                             & 4.6                                       & 1.85                                               & 12                                        & \textbf{18.45}                   \\
\textbf{FMNIST}     & 10                                          & 9000                                           & 2.4                                          & 0.22                                                  & 12.1                                         & 14.72                               & 4                                               & 1.34                                                     & 12.1                                            & 17.44                                  & 157.6                              & 4.6                                       & 1.89                                               & 12.1                                      & \textbf{18.59}                   \\
\textbf{CIFAR10}    & 10                                          & 10280                                          & 3.2                                          & 0.24                                                  & 13.7                                         & 17.14                               & 8.6                                             & 0.56                                                     & 13.7                                            & 22.86                                  & 217.4                              & 9.2                                       & 4.26                                               & 13.7                                      & \textbf{27.16}                   \\
\textbf{SVHN}       & 10                                          & 10280                                          & 3.2                                          & 0.23                                                  & 12.8                                         & 16.23                               & 8.8                                             & 0.38                                                     & 12.8                                            & 21.98                                  & 180                                & 9.6                                       & 3.37                                               & 12.8                                      & \textbf{25.77}                   \\
\textbf{CINIC10}    & 10                                          & 22332                                          & 5.6                                          & 0.42                                                  & 20.6                                         & 26.62                               & 10.2                                            & 1.35                                                     & 20.6                                            & 32.15                                  & 162                                & 33.4                                      & 14.12                                              & 20.6                                      & \textbf{68.12}                   \\
\textbf{SUN397\_32} & 50                                          & 14258                                          & 110.4                                        & 0.25                                                  & 19.7                                         & 130.35                              & 322                                             & 0.36                                                     & 19.7                                            & 342.06                                 & 368.2                              & 15                                        & 2.94                                               & 19.7                                      & \textbf{37.64}                   \\
\textbf{SUN397\_64} & 50                                          & 14258                                          & 91.8                                         & 2.68                                                  & 55.7                                         & 150.18                              & 374.4                                           & 0.41                                                     & 55.7                                            & 430.51                                 & 404.8                              & 15                                        & 2.75                                               & 55.7                                      & \textbf{73.45}                   \\
\textbf{GTSRB32}    & 43                                          & 39209                                          & 9.6                                          & 0.36                                                  & 86.8                                         & 96.76                               & 46.8                                            & 0.72                                                     & 86.8                                            & 134.32                                 & 1758                               & 36.2                                      & 7.66                                               & 86.8                                      & \textbf{130.66}                  \\ \bottomrule
\end{tabular}
}
\end{table}

\begin{figure}
\centering
\includegraphics[width=0.6\textwidth]{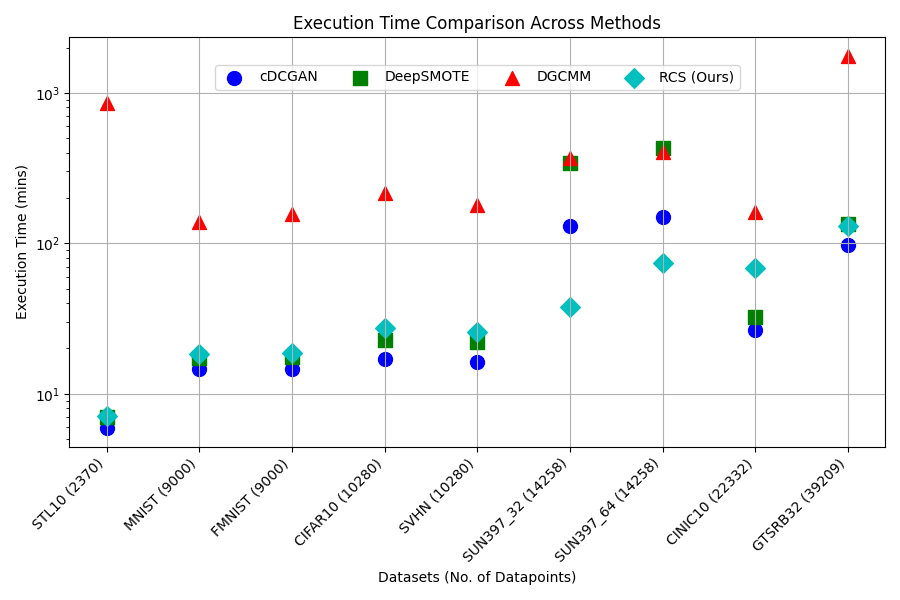}

\caption{\textcolor{black}{Execution time illustration for various methods}}
\label{fig:scalibility_time}
\end{figure}

\begin{table}
\caption{\textcolor{black}{Results on CINIC10 dataset}}
\label{tab:cinic10_results}
\centering
\color{black}
\resizebox{0.7\textwidth}{!}{%
\begin{tabular}{ccccc}
\toprule
\textbf{Metrics} & \textbf{DeepSMOTE}                           & \textbf{CDCGAN}                              & \textbf{DGCMM}                               & \textbf{RCS}                                         \\ \midrule
\textbf{BACC}    & 0.253$\pm$0.007       & 0.339$\pm$0.009       & 0.215$\pm$0.005      & \textbf{0.348$\pm$0.014}       \\
\textbf{MCC}     & 0.181$\pm$0.008 & 0.276$\pm$0.009 & 0.145$\pm$0.008 & \textbf{0.284$\pm$0.016} \\
\textbf{F1}      & 0.189$\pm$0.0145  & 0.283$\pm$0.0128  & 0.145$\pm$0.004 & \textbf{0.315$\pm$0.017} \\
\textbf{GMean}   & 0.482$\pm$0.006 & 0.560$\pm$0.008 & 0.443$\pm$0.005 & \textbf{0.568$\pm$0.012} \\ \bottomrule
\end{tabular}}
\end{table}

\color{black}
As dataset sizes increase, for the medium-sized datasets, the computational burden of RCS grows, but in a controlled manner compared to other methods. Take CIFAR10 (10,280 samples), for example—while cDCGAN and DeepSMOTE complete retraining in 17.14 mins and 22.86 mins respectively, RCS takes 27.16 mins. Although slightly higher, it remains significantly more efficient than DGCMM, which requires a 217.4 mins. A similar trend is seen in SVHN (10,280 samples), where RCS takes 25.77 mins, slightly more than DeepSMOTE (21.98 mins) and cDCGAN (16.23 mins). While RCS may take longer to train initially, its sample generation and classification efficiency make it more sustainable in the long run.

For even larger datasets, such as CINIC10 (22,332 samples) and SUN397 (14,258 samples with 50 classes), RCS begins to show its true scalability advantage. In CINIC10, for instance, RCS takes 68.12 mins, while DeepSMOTE and cDCGAN take 32.15 mins and 26.62 mins, respectively. However, DGCMM takes longer time 162 mins, showing that traditional clustering-based methods struggle as dataset sizes grow. The biggest advantage of RCS becomes clear in complex datasets like SUN397. In the SUN$397_{32\times 32}$ dataset, RCS completes retraining in just $37.64$ mins—far less than DeepSMOTE ($342.06$ mins) and DGCMM ($368.20$ mins). Similarly, in SUN$397_{64\times 64}$, RCS takes $73.45$ mins, whereas DeepSMOTE and DGCMM require time of about $430.51$ mins and $404.80$ mins respectively. This demonstrates that RCS can handle large datasets with many classes much more efficiently than the other methods.

The benefits of RCS are also evident in GTSRB32 (39,209 samples with 43 classes), where it completes retraining in 130.66 mins—similar to DeepSMOTE (134.32 mins) but vastly more efficient than DGCMM, which requires an overwhelming 1,758 mins. This efficiency comes from RCS’s reinforcement learning approach, which intelligently balances class distribution during training, reducing unnecessary computations and ensuring better sample efficiency.

In Figure~\ref{fig:scalibility_time} we have provided a scatter plot and the scalability of RCS and other state-of-the-art method. We observe that as dataset size increases the overall execution time for our method also increases consistently. So, RCS proves to be more scalable and computationally efficient than DeepSMOTE, cDCGAN, and DGCMM as dataset sizes increase. While it may take slightly longer to train on smaller datasets, its intelligent sample generation and classification strategies allow it to handle larger datasets much more efficiently. Unlike DeepSMOTE and cDCGAN, which experience exponential growth in sample generation time, RCS maintains a steady and manageable increase. This scalability stems from the adaptive nature of our method, which leverages autoencoder-based feature learning to efficiently handle datasets of increasing size and complexity. 
\color{black}
\begin{figure}[!ht]
    \centering
    \includegraphics[width=0.42\textwidth]{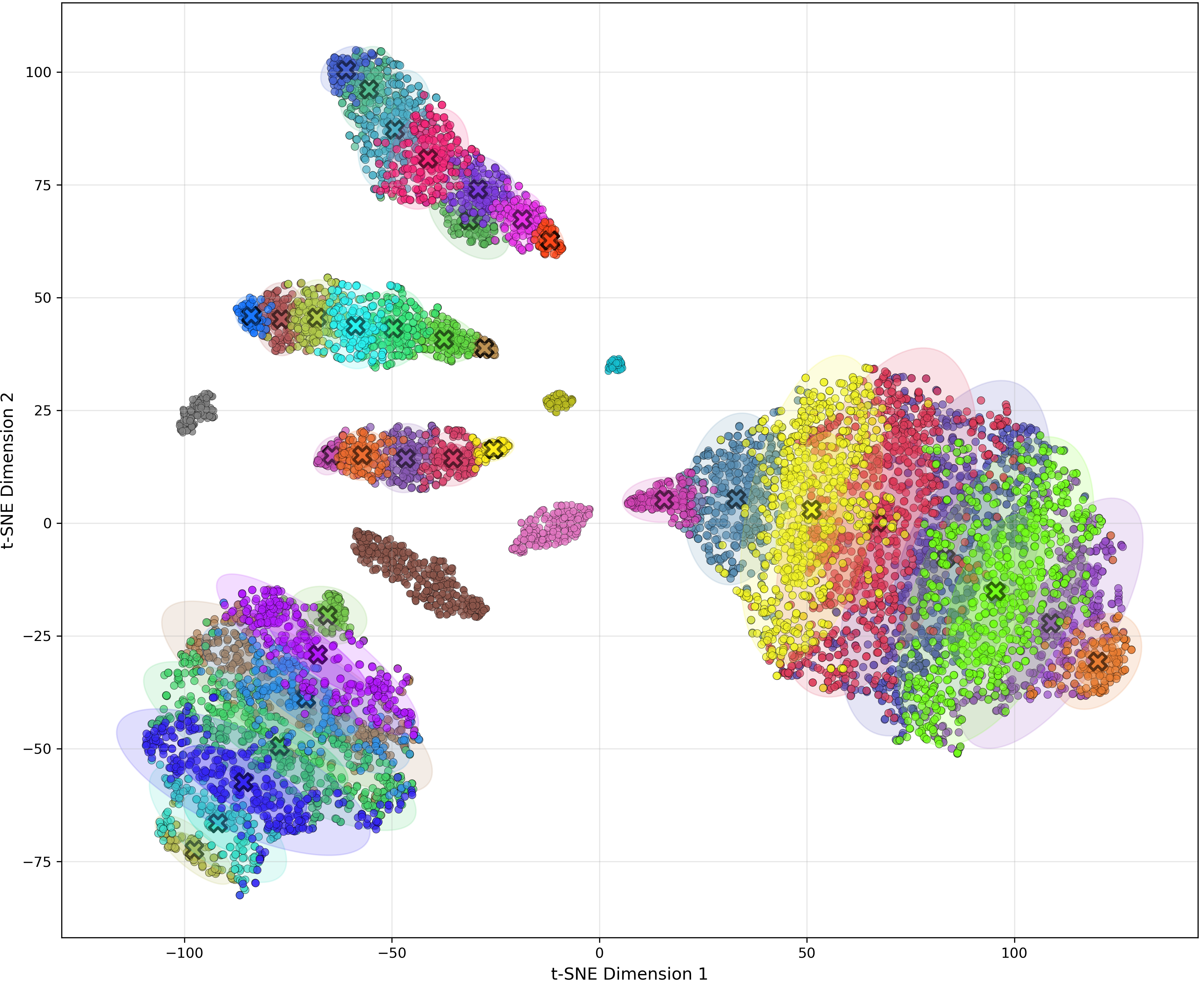} 
\includegraphics[width=0.42\textwidth]{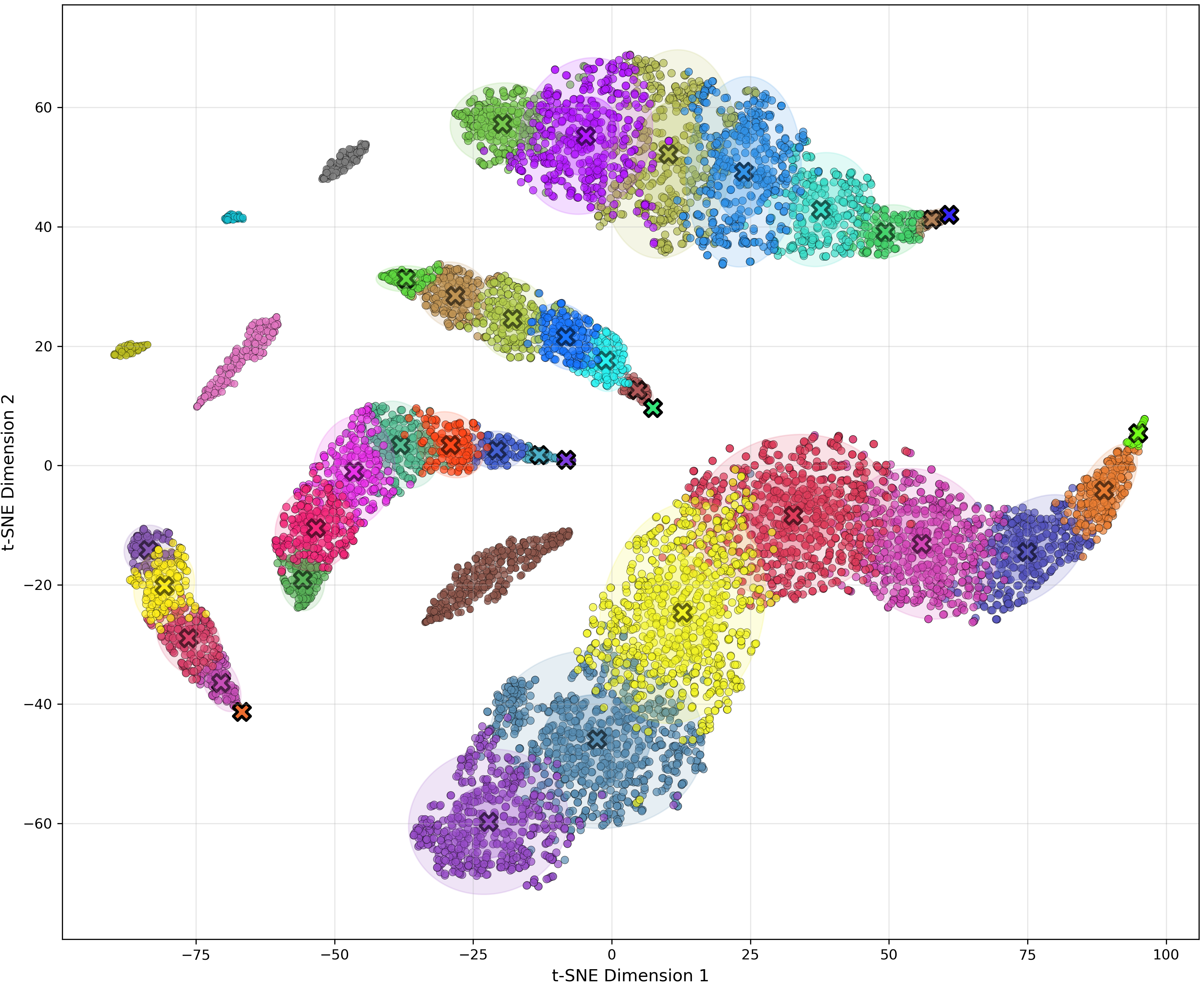} 
\raisebox{0.2cm}{\includegraphics[width=0.1\textwidth, height=5.0cm]{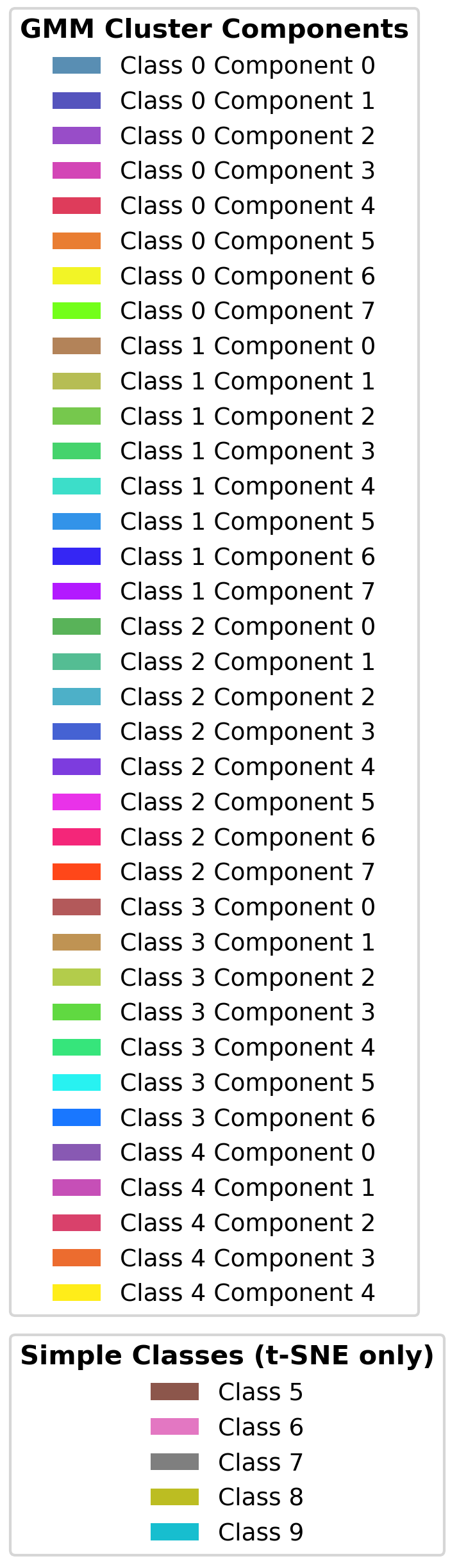}}
    \caption{\textcolor{black}{Visualizations on the Gaussian Mixture components (at most 8) on MNIST and FMNIST dataset, the classes 0,1,2,3,4 as the majority and intermediate classes, and the remaining classes are considered as minority}}
    \label{fig:gmm_centroids}
\end{figure}
\subsection{\textcolor{black}{Analysis of Gaussian Mixture Model Components in Latent Space}}
\color{black}
To further support the design decisions in our method, particularly the choice of using GMMs on latent representations, we include a visual analysis of the GMM clustering applied to the majority and intermediate classes. Analysis is done on the latent vector representations for the MNIST and FMNIST datasets. We then apply GMM clustering (with up to 8 components per class) to the latent vectors of the majority and intermediate classes. These components serve as the basis for local distribution calibration during synthetic generation.

\begin{figure}[h]
    \centering
\includegraphics[width=0.9\textwidth]{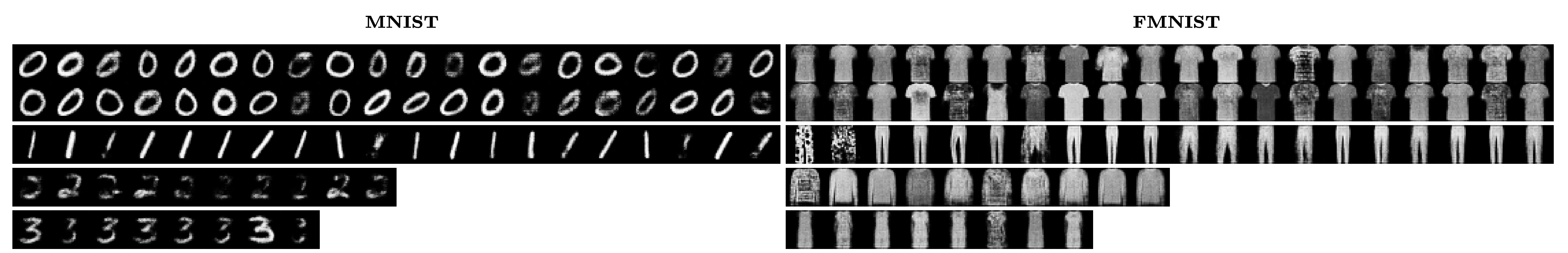} \hspace{-0.7cm}
    \caption{\textcolor{black}{Visualizations on the centroids for the components of the Gaussian Mixture Models corresponding to the majority and the intermediate classes on the MNIST and FMNIST datasets}}
    \label{fig:centroids_mnist}
\end{figure}

Figure~\ref{fig:gmm_centroids} presents t-SNE visualizations of these latent vectors, where each color represents a different GMM component (subclass) within a class. As seen in the plots, the learned latent space is well-structured and class-discriminative, with GMM components capturing localized intra-class modes. Notably, components from different classes remain well-separated in the t-SNE space, suggesting that our encoder's regularization has yielded a highly disentangled latent representation. We further decode the centroids (means) of these components using the trained decoder to obtain the visual prototypes of each subclass. As shown in Figure~\ref{fig:centroids_mnist}, the centroid images exhibit rich intra-class variation. For example, different stroke styles for digits in MNIST or apparel types in FMNIST. These centroid visualizations confirm that the GMM components are not arbitrary but correspond to meaningful subconcepts within each class.

Building on this structure, our synthetic generation strategy leverages these centroids in a k-nearest neighbor (k-NN) search. For each minority class sample requiring augmentation, we perform a neighbor search against the GMM centroids of the majority and intermediate classes. This neighbor selection enables us to identify statistically similar components in latent space, which then guides the calibration of Gaussian parameters used for generating synthetic data. More concretely, for a given minority sample, we select the top-k nearest centroids and compute a weighted average of their means and covariances. This neighbor-weighted estimation helps maintain distributional fidelity. The estimated distributional parameters derived from this centroid-based neighbor selection are subsequently used to sample synthetic instances from the corresponding Gaussian distribution, ensuring generated samples maintain statistical fidelity to the original data manifold while addressing class imbalance. This neighbor-weighted estimation strategy operates by computing weighted statistical moments based on the proximity of neighboring centroids to the target component, where closer centroids contribute more significantly to the final parameter estimates. 
Our ablation results (in section \ref{cal_vs_rcs}) demonstrate that this centroid-based approach significantly outperforms random neighbor selection and distance-based methods, achieving superior performance in downstream classification tasks while maintaining distributional consistency across all experimental benchmarks, thereby validating the critical role of GMM centroids in guiding effective synthetic data generation for imbalanced learning scenarios. 
Also, in Figure ~\ref{fig:centroids_mnist}, we observe how the subconcepts in a particular class are captured by the centroids of the Gaussian Mixture Components of the Majority and the Intermediate classes. These subconcepts enhance the pattern coverage of the minority class in generating synthetic samples.

\color{black}

\subsection{Visualization of our method based on Cluster Separability Index}
\label{clust_sep}
\textcolor{black}{Moreover, in Table~\ref{tab:my-table_clust_score}, we presented the Clustering Separability Index to measure the degree of disentanglement of the latent feature vectors before and after oversampling. We obtained the Silhouette Score and Davis Bouldin Index on the latent feature vectors of MNIST, FMNIST, CIFAR10, and SVHN datasets to achieve that. Silhouette Score measures the goodness of clustering, having values ranging from -1 to 1. Where the value 1 signifies clusters are well apart from each other and clearly distinguished, value 0 implies the distance between clusters is not significant, and -1 implies clusters are wrongly assigned. Davis Bouldin Index is calculated as the average similarity measure of each cluster with the most similar cluster, where the similarity measure is defined as the ratio between inter-cluster and intra-cluster distances. A higher DBI value indicates that the clusters are not well separated and not compact, whereas the lower values indicate that the clusters are well separated and compact. In Table~\ref{tab:my-table_clust_score}, we observe that for the MNIST dataset, the Silhouette Score values of our method improve by $0.873$ and $0.738$ for imbalanced latent features and oversampled latent feature vectors, respectively, which is better than the DeepSMOTE Autoencoder. In terms of the Davis Bouldin Index of our method on the MNIST dataset, we observed $0.309$ and $0.417$ values, which are closer to zero and much lower than the values of DeepSMOTE Autoencoder as $29.921$ and $22.341$ for our imbalanced latent features and oversampled latent feature vectors, respectively. For the FMNIST dataset, the Silhouette Score values for our method improved by 0.590 and 0.429 for imbalanced latent features and oversampled latent feature vectors, respectively, which is better than the DeepSMOTE Autoencoder. In terms of the Davis Bouldin Index of our method on the FMNIST dataset, we observed $0.762$ and $1.041$ values, which are much lower than the values of DeepSMOTE Autoencoder as $29.360$ and $15.217$ for our imbalanced latent features and oversampled latent feature vectors, respectively. Similarly, for CIFAR10 and SVHN, we observed that the Silhouette score with respect to the Imbalanced Latent Features and Oversampled latent features has a higher value compared to the values for DeepSMOTE
Autoencoder. Also, in terms of the Davis Bouldin Index for CIFAR10 and SVHN datasets, we observed much lower values for our method with respect to Imbalanced Latent Features and Oversampled Latent Vectors, respectively, than the DeepSMOTE Autoencoder.
}
\begin{table}[!ht]
\scriptsize
\centering
\color{black}
\caption{\textcolor{black}{Relative Measure of Cluster Separability between DeepSMOTE and our Proposal}}
\label{tab:my-table_clust_score}
\resizebox{\textwidth}{!}{%
\begin{tabular}{cccccc}
\toprule
\multirow{2}{*}{\textbf{Dataset}} & \multirow{2}{*}{\textbf{Method}} & \multicolumn{2}{c}{\textbf{Imbalanced Latent Vectors}}                                          & \multicolumn{2}{c}{\textbf{Oversampled Latent Vectors}}                                         \\ \cmidrule{3-4}\cmidrule{5-6}
                                  &                                  & \multicolumn{1}{l}{\textbf{Silhouette Score}} & \multicolumn{1}{l}{\textbf{Davis Bouldin Index}} & \multicolumn{1}{l}{\textbf{Silhouette Score}} & \multicolumn{1}{l}{\textbf{Davis Bouldin Index}} \\ \midrule
\multirow{2}{*}{\textbf{MNIST}}   & \textbf{DeepSMOTE}               & -0.051                                  & 29.921                                      & -0.047                               & 22.341                                      \\
                                  & \textbf{Ours}                    & \textbf{0.822}                          & \textbf{0.309}                            & \textbf{0.691}                        & \textbf{0.417}                            \\\\
\multirow{2}{*}{\textbf{FMNIST}}  & \textbf{DeepSMOTE}               & -0.035                                 & 22.360                                      & -0.023                               & 15.217                                      \\
                                  & \textbf{Ours}                    & \textbf{0.555}                           & \textbf{0.762}                            & \textbf{0.406}                        & \textbf{1.041}                             \\\\
\multirow{2}{*}{\textbf{CIFAR10}} & \textbf{DeepSMOTE}               & -0.155                                  & 9.022                                      & -0.112                                & 8.072                                      \\
                                  & \textbf{Ours}                    & \textbf{0.374}                          & \textbf{1.049}                             & \textbf{0.282}                        & \textbf{1.338}                             \\\\
\multirow{2}{*}{\textbf{SVHN}}    & \textbf{DeepSMOTE}               & -0.026                                 & 36.508                                       & -0.050                               & 28.445                                      \\
                                  & \textbf{Ours}                    & \textbf{0.490}                           & \textbf{0.678}                            & \textbf{0.350}                        & \textbf{0.920}                            \\ \bottomrule
\end{tabular}}
\end{table}


\end{document}